\documentclass[11pt]{article}

\RequirePackage[OT1]{fontenc}
\RequirePackage{amsthm,amsmath}
\RequirePackage[numbers]{natbib}
\RequirePackage[colorlinks,citecolor=blue,urlcolor=blue]{hyperref}

\usepackage{hyperref}       
\usepackage{url}            
\usepackage{booktabs}       
\usepackage{amsfonts}       
\usepackage{nicefrac}       
\usepackage{microtype}      

\usepackage[titletoc,title]{appendix}

\usepackage{amssymb}

\usepackage{subfloat}

\usepackage{enumitem}
\usepackage{caption}
\usepackage{subcaption}
\usepackage{graphicx}

\usepackage{xcolor,colortbl}

\usepackage{xspace}

\usepackage[margin=1in]{geometry}

\newtheorem{theorem}{Theorem} 
\newtheorem{lemma}{Lemma}
\newtheorem{proposition}{Proposition}
\newtheorem{corollary}{Corollary}

\newtheorem{remark}{Remark}

\newcommand{\delcrit}{\ensuremath{\delta_n}}

\newcommand{\DelHat}{\ensuremath{\widehat{\Delta}}}

\newcommand{\tracer}[2]{\ensuremath{\langle \!\langle {#1}, \; {#2}
\rangle \!\rangle}}

\newcommand{\finalver}[1]{#1}
\newcommand{\draftver}[1]{}

\newcommand{\green}[1]{\textcolor[rgb]{.1,.6,.1}{#1}}
\newcommand{\blue}[1]{\textcolor{blue}{#1}}
\draftver{
\newcommand{\mjwcomment}[1]{{\bf{{\blue{{MJW --- #1}}}}}}
\newcommand{\sbcomment}[1]{{\bf{{\brown{{SB --- #1}}}}}}
\newcommand{\ns}[1]{{\bf{{\green{{NS --- #1}}}}}}
\newcommand{\rev}[1]{{\color{red}{#1}}}
}
\finalver{
\newcommand{\mjwcomment}[1]{}
\newcommand{\sbcomment}[1]{}
\newcommand{\ns}[1]{}
\newcommand{\rev}[1]{#1}
}

\newcommand{\defn}{\ensuremath{:\,=}}
\newcommand{\Lnorm}[2]{\ensuremath{\|#1\|_{#2}}}
\newcommand{\inprod}[2]{\ensuremath{\langle #1 , \, #2 \rangle}}
\newcommand{\Exs}{\ensuremath{\mathbb{E}}}
\newcommand{\argmax}{\operatornamewithlimits{arg~max}}
\newcommand{\argmin}{\operatornamewithlimits{arg~min}}
\newcommand{\polylog}[1]{\ensuremath{\mbox{polylog}(#1)}}

\newcommand{\kl}[2]{\ensuremath{D_{\mathrm{KL}}(#1\|#2)}}
\newcommand{\reals}{\ensuremath{\mathbb{R}}}
\newcommand{\mprob}{\ensuremath{\mathbb{P}}}
\newcommand{\hamming}{\ensuremath{\dham}}
\newcommand{\half}{\ensuremath{{\frac{1}{2}}}}

\newcommand{\noise}{\ensuremath{\Wmat}}
\newcommand{\covnum}{\ensuremath{N}}
\newcommand{\metent}{\ensuremath{\log \covnum}}

\newcommand{\obs}{\ensuremath{Y}}

\newcommand{\permset}{\ensuremath{{\cal P}}}
\newcommand{\perm}{\ensuremath{\pi}}

 \newcommand{\plaincon}{c}

\newcommand{\packnum}{\ensuremath{\beta}}

 \newcommand{\ones}{\ensuremath{1}}

\newcommand{\cardinality}[1]{\ensuremath{| #1 |}}

\newcommand{\AuxEvent}{\ensuremath{\mathcal{A}_t}}

\newcommand{\chattDiffmx}{\ensuremath{D}}

\newcommand{\pp}{\ensuremath{p_{\mathrm{obs}}}}

\newcommand{\matsnorm}[2]{|\!|\!| #1 | \! | \!|_{{#2}}}

\newcommand{\order}{\ensuremath{\mathcal{O}}}

\long\def\comment#1{}

\newcommand{\indicator}[1]{\ensuremath{\mathbf{1}\{#1\}}}
\newcommand{\diag}[1]{\ensuremath{\operatorname{diag}(#1)}}
\newcommand{\opnorm}[1]{\ensuremath{\matsnorm{#1}{\mbox{\tiny{op}}}}}
\newcommand{\frobnorm}[1]{\ensuremath{\matsnorm{#1}{\mbox{\tiny{F}}}}}

\newcommand{\Wmat}{\ensuremath{W}}

\newcommand{\real}{\ensuremath{\mathbb{R}}}

\newcommand{\eigenvalue}[2]{\lambda_{#1}(#2)}

\newcommand{\ULOW}{\ensuremath{\plaincon_{\scalebox{.5}{\mbox{\!L}}}}}
\newcommand{\UUP}{\ensuremath{\plaincon_{\scalebox{.5}{\mbox{\!U}}}}}
\newcommand{\UHP}{\ensuremath{\plaincon_{\scalebox{.5}{\mbox{\!H}}}}}
\newcommand{\UNUM}{\ensuremath{\plaincon_{\scalebox{.5}{\mbox{\!0}}}}}

\newcommand{\dham}{\ensuremath{d_{\mbox{\tiny{H}}}}}

\newcommand{\numwork}{n}

\newcommand{\probvector}{q}
\newcommand{\probDS}{\probvector^{\mbox{\tiny DS}}}
\newcommand{\probDSplus}{\probvector^{\mbox{\tiny DS$+$}}}
\newcommand{\probDSminus}{\probvector^{\mbox{\tiny DS$-$}}}
\newcommand{\probInt}{\widetilde{\probvector}}

\newcommand{\numques}{d}
\newcommand{\permwork}{\pi}
\newcommand{\permworkinv}{\permwork^{-1}}
\newcommand{\permworkstar}{\permwork^*}
\newcommand{\permques}{\sigma}
\newcommand{\permquesstar}{\permques^*}
\newcommand{\probmx}{Q}
\newcommand{\probmxstar}{{\probmx}^*}
\newcommand{\probmxhat}{\widehat{\probmx}}
\newcommand{\probmxtilde}{\widetilde{\probmx}}
\newcommand{\probmxlse}{\widetilde{\probmx}_{\mbox{\tiny{LS}}}}
\newcommand{\classPerm}{\ensuremath{\mathbb{C}_{\mbox{\scalebox{.8}{{Perm}}}}}}
\newcommand{\classDS}{\ensuremath{\mathbb{C}_{\mbox{\scalebox{.8}{{DS}}}}}}
\newcommand{\genericclass}{\ensuremath{\mathbb{C}}}

\newcommand{\ans}{\ensuremath{x}}
\newcommand{\ansstar}{\ensuremath{\ans^*}}
\newcommand{\anshat}{\ensuremath{\widehat{\ans}}}

\newcommand{\anslse}{\ensuremath{\widetilde{\ans}_{\mbox{\tiny{LS}}}}}

\newcommand{\permqueslse}{\ensuremath{\widetilde{\permques}_{\mbox{\tiny{LS}}}}}
\newcommand{\permworklse}{\ensuremath{\widetilde{\permwork}_{\mbox{\tiny{LS}}}}}

\newcommand{\ansmv}{\ensuremath{\widetilde{\ans}_{\mbox{\tiny{MV}}}}}
\newcommand{\Qloss}[3]{\mathcal{L}_{#1}(#2,#3)}
\newcommand{\QlossPlain}[1]{\mathcal{L}_{#1}}

\newcommand{\lseinter}{V} 
\newcommand{\lseinterstar}{\lseinter^*}

\newcommand{\setexpertworkers}{S} 

\newcommand{\genericvec}{v}

\newcommand{\winsizeWAN}{k_{\mbox{\scalebox{.5}{WAN}}}}

\newcommand{\winsize}{k}

\newcommand{\ansWAN}[1][\permwork]{\anshat_{\mbox{\scalebox{.5}{WAN}}}(#1)}

\newcommand{\probmxshift}{R^*}
\newcommand{\probshift}{r^*}

\newcommand{\onesvec}[1]{g_{#1}}

\newcommand{\ansOBIWAN}{\anshat_{\mbox{\scalebox{.5}{OBI-WAN}}}}

\newcommand{\hard}{h}
\newcommand{\hardstar}{{\hard}^*}
\newcommand{\classInter}{\ensuremath{\mathbb{C}_{\mbox{\scalebox{.8}{Int}}}}}

\newcommand{\setques}{T}
\newcommand{\setquesval}{\widetilde{T}}
\newcommand{\errevent}{\mathcal{E}}

\newcommand{\probshiftapprox}{\widetilde{r}}

\newcommand{\permallwork}{\ensuremath{\Pi_\numwork}}
\newcommand{\permallques}{\ensuremath{\Sigma_\numques}}
\newcommand{\lseintertilde}{\widetilde{\lseinter}}
\newcommand{\lseinterdiff}{\ensuremath{\mathbb{\lseinter}_{\mbox{\scalebox{.5}{{DIFF}}}}}}

\renewcommand{\delcrit}{\delta_0}

\newcommand{\mxnew}{\widetilde{M}}
\newcommand{\mxorig}{M}
\newcommand{\mxdelta}{\Delta M}
\newcommand{\vecnewtoprightSV}{\widetilde{v}}
\newcommand{\vectoprightSV}{v}

\newcommand{\eigvecobs}{u}
\newcommand{\eigvecobsplus}{u^{+}}
\newcommand{\eigvecobsminus}{u^{-}}

\newcommand{\setsignalques}{J}

\newcommand{\numtopwork}{\alpha}

\newcommand{\probshiftmag}{\rho} 

\newcommand{\probshiftplus}{r^{*+}}
\newcommand{\probshiftminus}{r^{*-}}

\newcommand{\obiwan}{{OBI-WAN}\xspace}

\newcommand{\actualbias}{\gamma}

\newcommand{\Cost}{\ensuremath{\mathcal{C}}}
\newcommand{\setquesnew}[1]{J(#1)}

\newcommand{\capconst}{\widetilde{c}}

\newcommand\blfootnote[1]{%
  \begingroup
  \renewcommand\thefootnote{}\footnote{#1}%
  \addtocounter{footnote}{-1}%
  \endgroup
}

 \long\def\comment#1{}

\begin{document}


\begin{center}
{\bf{\LARGE{A Permutation-based Model for Crowd Labeling:\\ Optimal
			Estimation and Robustness}}}


\vspace*{.2in}

\begin{tabular}{c}
	Nihar B. Shah$^{\ast}$, Sivaraman Balakrishnan$^{\sharp}$ and Martin J. Wainwright$^{\dagger}$
\end{tabular}

\vspace*{.2in}

\begin{tabular}{c}
	$^{\ast}$ Machine Learning Department and Computer Science Department\\
	$^{\sharp}$Department of Statistics and Data Science\\
Carnegie Mellon University\\~\\
	$^{\dagger}$Department of EECS and Department of Statistics\\
	University of California, Berkeley\\
	~\\
\end{tabular}

\end{center}
\blfootnote{Author email addresses: nihars@cs.cmu.edu,
	siva@stat.cmu.edu, wainwrig@berkeley.edu.}


\begin{abstract}
The task of aggregating and denoising crowd-labeled data has gained
increased significance with the advent of crowdsourcing platforms and
massive datasets.  We propose a permutation-based model for crowd
labeled data that is a significant generalization of the classical
Dawid-Skene model, and introduce a new error metric by which to
compare different estimators. \rev{We derive global minimax rates for
  the permutation-based model that are sharp up to logarithmic
  factors, and match the minimax lower bounds derived under the
  simpler Dawid-Skene model.  We then design two
  computationally-efficient estimators: the {\sc WAN} estimator for
  the setting where the ordering of workers in terms of their
  abilities is approximately known, and the {\sc \obiwan} estimator
  where that is not known.  For each of these estimators, we provide
  non-asymptotic bounds on their performance.  We  conduct synthetic simulations and
  experiments on real-world crowdsourcing data, and the experimental results  
   corroborate our
  theoretical findings.}
\end{abstract}


\section{Introduction}
\label{sec:intro}

Recent years have witnessed a surge of interest in the use of
crowdsourcing for labeling massive datasets.  Expert labels are often
difficult or expensive to obtain at scale, and crowdsourcing platforms
allow for the collection of labels from a large number of low-cost
workers.  This paradigm, while enabling several new applications of
machine learning, also introduces some key challenges: first, low-cost
workers are often non-experts and the labels they produce can be quite
noisy, and second, data collected in this fashion has a high amount of
heterogeneity with significant differences in the quality of labels
across workers and tasks.  Thus, it is important to develop realistic
models and scalable algorithms for aggregating and drawing meaningful
inferences from the noisy labels obtained via crowdsourcing.


This paper focuses on objective labeling tasks involving binary
choices, meaning that each question or task is associated with a
single correct binary answer or label.\footnote{In this paper, we use
  the terms $\{$question, task$\}$, and $\{$answer, label$\}$ in an
  interchangeable manner.} There is a vast literature on the problem
of estimation from noisy crowdsourced labels
(e.g.,~\cite{sheng2008get, raykar2010learning, karger2011iterative,
  karger2011budget, ghosh2011moderates, liu2012variational,
  gao2013minimax, dalvi2013aggregating, zhang2014spectral,
  gao2016exact}).  The bulk of this past work is based on the
classical Dawid-Skene model~\cite{dawid1979maximum}, in which each
worker $i$ is associated with a single scalar parameter $\probDS_i \in
[0,1]$, and it is assumed that the probability that worker $i$ answers
any question $j$ correctly is given by the same scalar $\probDS_i$.
Thus, the Dawid-Skene model imposes a homogeneity condition on the
questions, one which is often not satisfied in practical applications
where some questions may be more difficult than others. {We note that
  the original model by Dawid and Skene~\cite{dawid1979maximum} also
  allows for asymmetric errors across different classes. In this
  paper, we focus on the setting with symmetric error probabilities,
  that has popularly come to be known as the ``one-coin Dawid-Skene
  model'', and has been the focus of much of past
  literature~\cite{karger2011iterative, karger2011budget,
    ghosh2011moderates, dalvi2013aggregating}. Both the asymmetric and
  symmetric models, however, are governed by restrictive
  parameter-based assumptions and assume homogeneity of questions.}

Accordingly, in this paper, we propose and analyze a more general
permutation-based model that allows the noise in the answer to depend
on the particular question-worker pair.  Within the context of such
models, we propose and analyze a variety of estimation algorithms.
One possible metric for analysis is the Hamming error, and there is a
large body of past work~\cite{karger2011iterative, karger2011budget,
  ghosh2011moderates, gao2013minimax, dalvi2013aggregating,
  zhang2014spectral, gao2016exact} that provide sufficient conditions that guarantee
zero Hamming error---meaning that every question is answered
correctly---with high probability.  Although the Hamming error can be
suitable for the analysis of Dawid-Skene style models, we argue in the
sequel that it is less appropriate for the heterogenous settings
studied in this paper.  Instead, when tasks have heterogenous
difficulties, it is more natural to use a weighted metric that also
accounts for the underlying difficulty of the tasks. Concretely, an
estimator should be penalized less for making an error on a question
that is intrinsically more difficult.  In this paper, we introduce and
provide analysis under such a difficulty-weighted error metric.

\noindent From a high-level perspective, the contributions of this
paper can be summarized as follows:
\begin{itemize}[leftmargin=*]
\setlength\itemsep{.1em}
\item We introduce a new ``permutation-based'' model for crowd-labeled
  data, which is considerably richer than the popular Dawid-Skene class of models. 
\item In order to incorporate the richness in the model, we introduce a new \mbox{difficulty-weighted} loss that extends the
popular Hamming loss. {We prove non-asymptotic upper and lower bounds on the global
  minimax error under the difficulty-weighted loss, sharp up to
  logarithmic factors, for estimation under the permutation-based
  model. These bounds match those under the Dawid-Skene model up to
  logarithmic factors.} 

\item  
\rev{We propose a computationally-efficient estimator, termed the WAN
  estimator, for the setting where an approximate ordering of the
  workers in terms of their abilities is known. We show that under the
  permutation-based model, this estimator has strong guarantees for
  the 0-1 loss and also achieves the global minimax limits (up to logarithmic factors) for the
  difficulty-reweighted loss.}
\item \rev{We provide a computationally-efficient estimator, termed the
  \obiwan estimator, when no prior information about the workers is known. This estimator achieves strong guarantees for the 0-1 loss under the Dawid-Skene model and an intermediate model, and simultaneously also has guarantees over the much richer permutation-based model thereby establishing its robustness to model specification.}

 \item We conduct synthetic simulations as well as real-world experiments using data from the Amazon Mechanical Turk crowdsourcing platform. These experiments reveal a strong performance of the \obiwan estimator in practice.
\end{itemize}

The remainder of this paper is organized as follows.  In
Section~\ref{SecModel}, we provide some background, setup the problems
we address in this paper, and provide an overview of related
literature. Section~\ref{SecMainResults} is devoted to our main
results.  We present numerical simulations and real-world experiments in
Section~\ref{SecExperiments}. We present proofs of the claimed theoretical results in Section~\ref{SecProofs}. We conclude the paper with a discussion of
future research directions in Section~\ref{SecConclusions}.


\section{Background and model formulation}
\label{SecModel}

We begin with some background on existing crowd-labeling models,
followed by an introduction to our proposed models; we conclude with a
discussion of related work.


\subsection{Observation model}
\label{SecObsModel}

Consider a crowdsourcing system that consists of $\numwork$ workers
and $\numques$ questions.  We assume every question has two possible
answers, denoted by $\{-1,+1\}$, of which exactly one is correct.  We
let $\ansstar \in \{-1,1\}^{\numques}$ denote the binary vector of
correct answers to all $\numques$ questions.  We model the
question-answering via an unknown matrix $\probmxstar \in
[0,1]^{\numwork \times \numques}$ whose $(i,j)^{th}$ entry,
$\probmxstar_{ij}$, represents the probability that worker $i$ answers
question $j$ correctly. Otherwise, with probability $1 -
\probmxstar_{ij}$, worker $i$ gives the incorrect answer to question
$j$.  For future reference, note that the (one-coin) Dawid-Skene model
involves a special case of such a matrix, namely one of the form
$\probmxstar = \probDS 1^T$, where the vector $\probDS \in
[0,1]^{\numwork}$ corresponds to the vector of correctness
probabilities, with a single scalar associated with each worker.

We denote the response of worker $i$ to question $j$ by a variable
$\obs_{ij} \in \{-1,0,1\}$, where we set $\obs_{ij} = 0$ if worker $i$
is not asked question $j$, and set $\obs_{ij}$ to the answer (either
$-1$ or $1$) provided by the worker otherwise. We also assume that
worker $i$ is asked question $j$ with probability $\pp \in [0,1]$,
independently for every pair $(i,j) \in [\numwork] \times [\numques]$,
and that a worker is never asked the same question twice.  We also
make the standard assumption that given the values of $\ansstar$ and
$\probmxstar$, the entries of $\obs$ are all mutually independent. In
summary, we observe a matrix $\obs$ which has independent entries
distributed as
\begin{align*}
\obs_{ij} = 
\begin{cases}
\ansstar_j & \qquad \mbox{with probability $\pp \; \probmxstar_{ij}$}\\
-\ansstar_j & \qquad \mbox{with probability $\pp \; (1 -
  \probmxstar_{ij})$}\\
0 & \qquad \mbox{with probability $(1 - \pp)$.}
\end{cases}
\end{align*}
Given this random matrix $\obs$, our goal is to estimate the binary
vector $\ansstar \in \{-1,1\}^\numques$ of true labels.

Obtaining non-trivial guarantees for this problem requires that some
structure be imposed on the probability matrix $\probmxstar$.  The
Dawid-Skene model is one form of such structure: it requires that the
probability matrix $\probmxstar$ be rank one, with identical columns
all equal to $\probDS \in \real^\numwork$.  As noted previously, this
structural assumption on $\probmxstar$ is very strong. It assumes that
each worker has a fixed probability of answering a question correctly,
and is likely to be violated in settings where some questions are more
difficult than others.

Accordingly, in this paper, we study a more general permutation-based
model of the following form.  We assume that there are two underlying
orderings, both of which are unknown to us: first, a permutation
$\permworkstar:[\numwork] \rightarrow [\numwork]$ that orders the
$\numwork$ workers in terms of their (latent) abilities, and second, a
permutation $\permquesstar: [\numques] \rightarrow [\numques]$ that
orders the $\numques$ questions with respect to their (latent)
difficulties.  In terms of these permutations, we assume that the
probability matrix $\probmxstar$ obeys the following conditions:
\begin{itemize}[leftmargin=*]
\setlength\itemsep{0em}
 \setlength{\parskip}{0pt}
 \item {\bf{Worker monotonicity:}} For every pair of workers $i$ and
   $i'$ such that $\permworkstar(i) < \permworkstar(i')$ and every question
   $j$, we have \mbox{$\probmxstar_{ij} \geq \probmxstar_{i' j}$.}
\item {\bf{Question monotonicity:}} For every pair of questions $j$ and $j'$ such that
  $\permquesstar(j) < \permquesstar(j')$ and every worker $i$, we have $\probmxstar_{ij} \geq \probmxstar_{i j'}$.
\end{itemize}
In other words, the permutation-based model assumes the existence of a
permutation of the rows and columns such that each row and each column
of the permuted matrix $\probmxstar$
has non-increasing entries. 
The rank of the resulting matrix is
allowed to be as large as $\min\{\numwork, \numques\}$. It is straightforward to
verify that the Dawid-Skene model corresponds to a particular type of such
probability matrices, restricted to have identical columns.  

In summary, we let $\classPerm$ denote the set of all possible values
of matrix $\probmxstar$ under the proposed permutation-based model,
that is,
\begin{align*}
\classPerm \defn \big\{  \probmx  \in  [0,1]^{\numwork
	\times \numques}  \mid & \mbox{there exist permutations $(\permwork,
	\permques)$ such that}\\
& \mbox{ question \& worker monotonicity hold} \big\}.
\end{align*}
For future reference, we also use
\begin{align*}
\classDS \defn \big \{ \probmx \in \classPerm \mid \probmx = \probDS
\ones^T \mbox{ for some $\probDS \in [0, 1]^\numwork$} \big \},
\end{align*}
to denote the subset of such matrices that are realizable under the
Dawid-Skene assumption.

It should be noted that none of these models are identifiable without
further constraints.  For instance, changing $\ansstar$ to $-
\ansstar$ and $\probmxstar$ to $(\ones \ones^T - \probmxstar)$ does
not change the distribution of the observation \mbox{matrix $\obs$.}
In the context of the Dawid-Skene model, several
papers~\cite{karger2011iterative, karger2011budget, gao2013minimax,
	zhang2014spectral} have resolved this issue by requiring that
$\frac{1}{\numwork} \sum_{i=1}^{\numwork} \probDS_i \geq \half+ \mu$
for some constant value $\mu > 0$. Although this condition resolves
the lack of identifiability, the underlying assumption---namely that
every question is answerable by a subset of the workers---can be
violated in practice.  In particular, one frequently encounters
questions that are too difficult to answer by any of the hired
workers, and for which the worker's answers are near uniformly random
(e.g., see the papers~\cite{eickhoff2011crowdsourcable,
	shah2016doubleJMLR}).  On the other hand, empirical observations also
show that workers in crowdsourcing platforms, as opposed to being
adversarial in nature, at worst provide random answers to labeling
tasks~\cite{yuen2011survey, eickhoff2011crowdsourcable,
	gadiraju2015understanding, gadiraju2015training}.  {On this basis, for certain results in the paper, we will consider the regime:}
\begin{align}
\tag{R1}
\probmxstar_{ij} \geq \half \qquad \forall i \in [\numwork], j \in [\numques].
\label{EqnMoreHalf}
\end{align}
{Note that neither the condition~\eqref{EqnMoreHalf} nor the condition $\frac{1}{\numwork} \sum_{i=1}^{\numwork} \probDS_i \geq \half+ \mu$ from past literature dominate one another.}



\subsection{Evaluating estimators}
\label{SecEvalModel}

In this section, we introduce the criteria used to evaluate estimators
in this paper.  In formal terms, an estimator $\anshat$ is a
measurable function that maps any observation matrix $\obs$ to a
vector in the Boolean hypercube $\{-1,1\}^{\numques}$. The most
popular way of assessing the performance of such an estimator is in
terms of its (normalized) \emph{Hamming error}
  \vspace{-.1cm}
\begin{align}
\label{EqnDefnHamming}
\dham({\anshat},{\ansstar}) \defn \frac{1}{\numques} \sum_{j=1}^{\numques}
\indicator{\anshat_j \neq \ansstar_j},
  \vspace{-.1cm}
\end{align}
where $\indicator{\anshat_j \neq \ansstar_j}$ denotes a binary
indicator which takes the value $1$ if $\anshat_j \neq \ansstar_j$,
and $0$ otherwise.  A potential
deficiency of the Hamming error is that it places a uniform weight on
each question.  As mentioned earlier, there are applications of
crowdsourcing in which some subset of the questions are very
difficult, and no hired worker can answer reliably.  In such settings,
any estimator will have an inflated Hamming error, not due to any
particular deficiencies of the estimator, but rather due to the intrinsic
hardness of the assigned collection of questions.  This error
inflation will obscure possible differences between estimators. 

 {Our goal in choosing an appropriate loss function is to allow for evaluation and comparison of various estimators.} Thus, with the aforementioned issue in mind, we an alternative error measure that
weights the Hamming error with the difficulty of each task.  A more
general class of error measures takes the form
\begin{align}
\label{EqnGeneral}
\Qloss{\probmxstar}{\anshat}{\ansstar} = \frac{1}{\numques} \sum_{j=1}^{\numques}
\indicator{ \anshat_j \neq \ansstar_j } \Psi( \probmxstar_{1j},\ldots,
\probmxstar_{\numwork j}),
\end{align}
for some function $\Psi\!:\! [0,1]^{\numwork} \rightarrow \reals_+$ 
which captures the difficulty of estimating the answer to a
question.

\paragraph*{The $\probmxstar$-loss:}
In order to choose a suitable function $\Psi$, we note that past work
on the Dawid-Skene model~\cite{karger2011iterative, karger2011budget,
  ghosh2011moderates, gao2013minimax, dalvi2013aggregating} has shown
that the quantity 
\begin{align}
\label{eqn:colint}
\frac{1}{\numwork} \sum_{i=1}^{\numwork} (2\probDS_i
- 1)^2, 
\end{align}
popularly known as the \emph{collective intelligence} of
the crowd, is central to characterizing the overall difficulty of the
crowd-sourcing problem under the Dawid-Skene assumption.  A natural
generalization, then, is to consider the weights
\begin{subequations}
\begin{align}
\label{EqnWeight}
\Psi( \probmxstar_{1j},\ldots, \probmxstar_{\numwork j}) & =
\frac{1}{\numwork} \sum_{i=1}^\numwork \big (2\probmxstar_{ij} - 1
\big)^2 \qquad \mbox{for each task $j \in [\numques]$,}
\end{align}
which characterizes the difficulty of task $j$ for a given collection
of workers.  This choice gives rise to the \emph{$\probmxstar$-loss
  function}
\begin{align}
\label{EqnDefnLoss}
\Qloss{\probmxstar}{\anshat}{\ansstar} & \defn \frac{1}{\numques} \sum_{j =1}^{\numques}
 \Big( \indicator{\anshat_j \neq \ansstar_j} \; \frac{1}{\numwork}
\sum_{i = 1}^{\numwork} (2\probmxstar_{ij} - 1)^2 \Big) \\
& = \frac{1}{\numques \numwork} \frobnorm{ (\probmxstar - \frac{1}{2} \ones
  \ones^T) \; \diag{\anshat - \ansstar} }^2,
\end{align}
\end{subequations}
where $\diag{\anshat - \ansstar}$ denotes the matrix in
$\mathbb{R}^{\numques \times \numques}$ whose diagonal entries are
given by the vector $\anshat - \ansstar$.  Note that under the
Dawid-Skene model (in which $\probmxstar = \probDS \ones^T$), this
loss function reduces to 
\begin{align}
\label{EqnQlossToHamming}
\Qloss{\probmxstar}{\anshat}{\ansstar} = \Big( \frac{1}{\numwork}
 \sum_{i=1}^{\numwork} (2\probDS_{i} - 1)^2 \Big) \; \underbrace{\Big( \frac{1}{\numques}
  \sum_{j =1}^{\numques} \indicator{\anshat_j \neq \ansstar_j} \Big
  )}_{ \dham({\anshat},{\ansstar})},
\end{align}
corresponding to the normalized Hamming error rescaled by the collective intelligence. 

For future reference, let us summarize some properties of the function
$\QlossPlain{\probmxstar}$: (a) it is symmetric in its arguments
$(\ansstar, \anshat)$, and satisfies the triangle inequality; (b) it
takes values in the interval $[0, 1]$; and (c) if for every question $j
\in [\numques]$, there exists a worker $\ell \in [\numwork]$ such that
$\probmxstar_{\ell j} \neq \half$, then $\QlossPlain{\probmxstar}$ defines
a metric; if not, it defines a pseudo-metric.


%

%


\paragraph*{Regime of interest:} 

In this paper, we focus on understanding the minimax risk as well as
the risk of various computationally efficient estimators.  We work in
a non-asymptotic framework where we are interested in evaluating the
risk in terms of the triplet $(\numwork, \numques, \pp)$.  We assume
that $\pp \geq \frac{1}{\numwork}$, which ensures that on average, at
least one worker answers any question. We also operate in the regime
$\numques \geq \numwork$, which is commonplace in practical
applications.  Indeed, as also noted in earlier
works~\cite{zhang2014spectral}, typical medium or large-scale
crowdsourcing tasks employ tens to hundreds of workers, while the
number of questions is on the order of hundreds to many thousands. We
assume that the value of $\pp$ is known. This is a mild assumption
since it is straightforward to estimate $\pp$ very accurately using
its empirical expectation. We encompass the aforementioned conditions as the regime: 
\begin{align}
\tag{R2}
\pp \geq \frac{1}{\numwork} \quad \mbox{and} \quad \numques \geq \numwork.
\label{EqnRegime}
\end{align} 


\subsection{Related work} 

Having set up our model and notation, let us now relate it to past
work in the area. For the problem of crowd labeling, the Dawid-Skene
model~\cite{dawid1979maximum} is the dominant paradigm, and has been
widely studied~\cite{karger2011iterative, karger2011budget,
  ghosh2011moderates,liu2012variational, gao2013minimax,
  dalvi2013aggregating, zhang2014spectral}. Some papers have studied
models that generalize the Dawid-Skene model. In a recent work, Khetan and
Oh~\cite{khetan2016reliable} analyze an extension of the Dawid-Skene
model where a vector $\probInt \in \reals^\numwork$, capturing the
abilities of the workers, is supplemented with a second vector
\mbox{$\hardstar \in [0,1]^\numques$,} and the likelihood of worker
$i$ correctly answering question $j$ is set as $\probInt_i (1 -
\hardstar_j + (1 - \probInt_i) \hardstar_j)$. Although this model now
has $(\numwork + \numques)$ parameters instead of just $\numwork$ as
in the Dawid-Skene model, it retains parametric-type assumptions. Each
worker and each question is described by a single parameter, and in
this model the probability of correctness takes a specific form
governed by these parameters. In contrast, in the permutation-based
model each worker-question pair is described by a single parameter.
Our permutation-based model forms a strict superset of this class.
{Zhou et al.~\cite{zhou2012learning,zhou2015regularized} propose a  model based on a certain minimax entropy principle, whereas
  Whitehill et al.~\cite{whitehill2009whose} propose a parameter-based
  model that also incorporates question difficulties.  However, the
  algorithms proposed in these papers~\cite{zhou2012learning,zhou2015regularized,whitehill2009whose} have yet to be rigorously
  analyzed. }  

\rev{In this paper, we introduce a class of models that are
  considerably more flexible than the Dawid-Skene model, as well as a
  novel algorithm for estimation in such models, which we equip with
  some theoretical guarantees.  The present paper also introduces
  another new algorithm for the setting in which an ordering of the
  workers in terms of their abilities is approximately known, for
  instance, based on some initial test.  To be clear, the results of
  this paper have some limitations as compared to past work on the
  Dawid-Skene model, and we hope that these limitations will be
  removed in future work on the permutation-based model. Concretely,
  while the present paper addresses the setting of binary labels with
  symmetric error probabilities, several of these prior works also
  address settings with more than two classes, and where the
  probability of error of a worker may be asymmetric across the
  classes. The results presented in this paper have logarithmic factor
  gaps, that is, the `optimal' results are optimal up to logarithmic factors, as
  stated throughout the paper. For the Dawid-Skene model,
  particularly under sparse observations, the past
  works~\cite{karger2011iterative, karger2011budget, gao2013minimax,
    zhang2014spectral,khetan2016reliable} have results with sharper
  logarithmic factors. Finally, the guarantees provided in past
  results have error exponents that adapt to the underlying signal,
  whereas ours do not.}

A related problem in the context of crowdsourcing is to estimate
pairwise outcome probabilities from pairwise comparison data. In our
past work~\cite{shah2015stochastically, shah2016feeling}, we have
considered this problem under an assumption of ``strong stochastic
transitivity (SST)'', which is a regularity condition related to the
permutation-based model of this paper.  Accordingly, parts of our
proofs make use of metric entropy calculations from this past work.
Unlike our previous work, the current paper involves an unknown set of
labels, as well as a significantly different observation model: in
particular, the observed data couples the unknown matrix $\probmxstar$
with the unknown labels. Moreover, rather than estimating the unknown
probabilities $\probmxstar$, our primary goal in this paper is to
estimate these underlying labels, for which significantly different
algorithmic ideas and proof techniques are required.

Finally, the problem of aggregating labels of crowdsourcing workers is
conceptually similar to that of combining classifiers in an
unsupervised context, each solving multiple classification
problems~\cite{parisi2014ranking, jaffe2015estimating}.  Our work has
implications for this line of research as well.


\section{Main results}
\label{SecMainResults}

We now turn to the statement of our main results.  We use {\it
  $\plaincon$, $\UUP$, $\ULOW$, $\UNUM$, $\UHP$} to denote positive
universal constants that are independent of all other problem
parameters. Recall that the $\probmxstar$-loss takes values in the
interval $[0,1]$.


\subsection{Minimax risk for estimation under the permutation-based model}

We begin by proving sharp upper and lower bounds on the minimax risk
for the permutation-based model $\classPerm$.  The upper bound is
obtained via an analysis of the following least squares estimator
\begin{align}
\label{EqnDefnLSE}
(\anslse, \probmxlse) \in \argmin \limits_{\ans \in
  \{-1,1\}^{\numques},\ \probmx \in \classPerm} \frobnorm{ \pp^{-1} \;
  \obs - (2\probmx - \ones \ones^T) \; \diag{\ans}}^2.
\end{align}
{In order to provide some intuition for this estimator, one can show
  (see the proof of Theorem~\ref{ThmMinimax}(a) for details) that the
  unknowns $\ansstar$ and $\probmxstar$ are related to the mean of the
  observed matrix $\obs$ via the equality \mbox{$\Exs[\obs] = \pp
    (2\probmxstar - \ones \ones^T) \; \diag{\ansstar}$}. Consequently,
  the estimate $(\anslse, \probmxlse)$ computed via the
  program~\eqref{EqnDefnLSE} equals the true solution $(\ansstar,
  \probmxstar)$ when $\obs$ is replaced by its population version.}

We do not know of a computationally efficient way to solve the
optimization problem~\eqref{EqnDefnLSE}.  Despite this computational issue, our statistical analysis
of this estimator serves to provide a benchmark for comparing other
computationally-efficient estimators, to be discussed in the sequel.
The following theoretical guarantees hold in the
regime~\eqref{EqnMoreHalf}$\cap$\eqref{EqnRegime}:
\begin{theorem}
\label{ThmMinimax}
(a) For any binary vector $\ansstar \in \{-1,1\}^\numques$ and any
matrix $\probmxstar \in \classPerm$, the least squares estimator
$\anslse$ has error at most
\begin{subequations}
\begin{align}
\label{EqnMinimaxUpper}
 \Qloss{\probmxstar}{\anslse}{\ansstar} \leq \UUP
 \frac{1}{\numwork \pp} (\log \numques)^2,
\end{align}
with probability at least $1 - e^{- \UHP \numques \log (\numques
  \numwork)}$. \\ (b) There exists a matrix $\probmxtilde \in
\classDS$ such that any estimator $\anshat$ (which may even know the
value of $\probmxtilde$) has error at least
\begin{align}
\label{EqnMinimaxLower}
\sup \limits_{\ansstar
	\in \{-1,1\}^\numques} \Exs [ \Qloss{\probmxtilde}{\anshat}{\ansstar}
] \geq \ULOW \frac{1}{\numwork \pp}.
\end{align} 
\end{subequations}
\end{theorem}
We provide the this theorem  in
 Sections~\ref{SecProofThmMinimaxa}
 and~\ref{SecProofThmMinimaxb}. 
 \rev{As a consequence of this result, we see that in terms of the
   (global) minimax risk under the $\probmxstar$-loss, there is only a
   polylogarithmic factor difference between the Dawid-Skene and the
   permutation-based models, despite the permutation-based model being considerably richer. }

\rev{We note that while the upper bound of Theorem~\ref{ThmMinimax}(a) is quite involved, the lower bound of Theorem~\ref{ThmMinimax}(b) is a straightforward result of a simple ``worst case'' construction. This suggests that the global minimax error be augmented with an investigation of local minimax errors under various subclasses of $\classPerm$ and various notions and values of the signal to noise ratio, which we leave as important future work.}

The least squares estimator analyzed above also
yields an accurate estimate of the probability matrix $\probmxstar$ in
the Frobenius norm, useful in settings where the calibration of
workers or questions might be of interest.  Again, this result holds
in the regime~\eqref{EqnMoreHalf}$\cap$\eqref{EqnRegime}:
\begin{corollary}
\label{CorEstimateQstar}
(a) For any $\ansstar \in \{-1,1\}^\numques$ and any $\probmxstar \in
\classPerm$,, the least squares estimate $\probmxlse$ has error at
most
\begin{subequations}
\begin{align}
\frac{1}{\numques \numwork} \frobnorm{\probmxlse - \probmxstar}^2 \leq
\UUP \frac{1}{\numwork \pp} \log^2 \numques,
\end{align}
with probability at least $1 - e^{- \UHP \numques \log (\numques \numwork)}$.\\
(b) Conversely, for any answer vector $\ansstar \in \{-1,
1\}^\numques$, any estimator $\probmxhat$ (which is allowed to know the value of $\ansstar$) has error at least
\begin{align}
\sup \limits_{\probmxstar \in \classPerm} \Exs [ \frac{1}{\numques
    \numwork} \frobnorm{\probmxhat - \probmxstar}^2 ] \geq \ULOW
\frac{1}{\numwork \pp}.
\end{align}
\end{subequations}
\end{corollary}
\noindent Please see Sections~\ref{SecProofCorEstimateQstara}
and~\ref{SecProofCorEstimateQstarb} for the proof of this corollary.

We do not know if there exist computationally-efficient estimators
that can achieve the upper bound on the sample complexity established
in Theorem~\ref{ThmMinimax} over the entire
permutation-based model class. In the following sections, we design
and analyze polynomial-time estimators that have interesting (but suboptimal) guarantees over the permutation-based model and also useful guarantees over popular 
subclasses of the permutation-based model.


\subsection{The WAN estimator: When workers' ordering is (approximately) known}
\label{SecCalibrated}
  
Several organizations employ crowdsourcing workers only after a
thorough testing and calibration process. Motivated by this fact, we
now turn to the study of the setting in which the workers are
calibrated, in the sense that it is known how they are ordered in
terms of their respective abilities. More formally, recall from
Section~\ref{SecObsModel} that any matrix $\probmxstar \in \classPerm$
is associated with two permutations: a permutation of the workers in
terms of their abilities, and a permutation of the questions in terms
of their difficulty. In this section, we assume that the permutation
of the workers is (approximately) known to the estimation
algorithm. Note that the estimator does \emph{not} know the
permutation of the questions, nor does it know the values of the
entries of $\probmxstar$.

Given a permutation $\permwork$ of the workers, our estimator consists
of two steps, which we refer to as Windowing and Aggregating
Na\"{\i}vely, respectively, and accordingly term the procedure as the
WAN estimator:
\begin{itemize}[leftmargin=*]
\vspace{-\topsep}
\setlength{\itemsep}{0em}
\item
Step $1$ (Windowing): Compute the integer
\begin{subequations}
\vspace{-\topsep}
\begin{align}
\label{EqnWindow}
\winsizeWAN \in \argmax_{\winsize \in \{ \pp^{-1} \log^{1.5} (\numques
  \numwork), \ldots,\numwork\}} ~~~ \sum_{j \in [\numques]} \mathbf{1}
\Big\{ \big| \sum_{i \in [\winsize]} \obs_{\permworkinv(i)j} \big| \geq
\sqrt{\winsize \pp \log^{1.5} (\numques \numwork) } \Big\},
\end{align}
\rev{where ties in the argmax are broken arbitrarily.}
\item Step 2 (Aggregating Na\"{\i}vely): Set $\ansWAN$ as a majority
  vote of the best $\winsizeWAN$ workers---that is
\begin{align}
\label{EqnNaiveAgg}
[\ansWAN]_j \in \argmax_{b \in \{-1,1\}} ~~~ \sum_{i=1}^{\winsizeWAN}
\indicator{ \obs_{\permworkinv(i)j} = b } \qquad \mbox{for every $j\in
  [\numques]$}.
\end{align}
\end{subequations}
\end{itemize}
The windowing step finds a value $\winsizeWAN$ such that the answers
of the best $\winsizeWAN$ workers to most questions are significantly
biased towards one of the options, thereby indicating that these
workers are knowledgeable---or at least, are in agreement with each
other. The second step then simply takes a majority vote of this set
of the best $\winsizeWAN$ workers. We remark that it is important to
choose an appropriate value of $\winsizeWAN$ (as done in Step 1),
since an overly large value could include many random workers, thereby
increasing the noise in the input to the second step; on the flip side,
choosing too small a value could eliminate too much of the ``signal''.
Both steps can be carried out in time $\order(\numwork \numques)$.

For the case when $\permwork$ is an approximate ordering, we now establish
a bound on the error of the WAN estimator.  For every $j \in [\numques]$, let
$\probmxstar_j$ denote the $j^{th}$ column of $\probmxstar$; for any
ordering $\permwork$ of the workers, $\probmx^\permwork_j$ denote the
vector obtained by permuting the entries of $\probmxstar_j$ in the
order given by $\permwork$, that is, with the first entry of
$\probmx^\permwork_j$ corresponding to the best worker according to
$\permwork$, and so on. Also recall the notation $\permworkstar$
representing the true permutation of the workers in terms of their
actual abilities.  As with all of our theoretical, results, the
following claim holds in the regime~\eqref{EqnMoreHalf}$\cap$\eqref{EqnRegime}:
\rev{
\begin{theorem}
\label{ThmWANnew}
For any matrix $\probmxstar \in \classPerm$ and any binary vector
$\ansstar \in \{-1,1\}^{\numques}$, suppose that the WAN estimator is
provided with the permutation $\permwork$ of workers. Consider the
subset of the questions given by
  \begin{subequations}
    \begin{align}
      \label{EqnDefnSetquesWAN}
      \setsignalques & \defn \Big\{  j \! \in \! [\numques] \mid \exists  \winsize_j
      \geq \frac{\log^{1.5} (\numques \numwork)}{\pp} \mbox{~~s.t.~~}
      \sum_{i=1}^{\winsize_j} (\probmxstar_{\permworkinv(i) j} -  \half)
      \geq  \frac{3}{4} \sqrt{\frac{\winsize_j}{\pp} \log^{1.5} (\numques
	\numwork)} \! \Big\}.
    \end{align}
Then the WAN estimator correctly estimates the labels of all
questions in set $\setsignalques$ with high probability:
\begin{align}
  \mprob \Big( [\ansWAN]_{j} = \ansstar_{j} ~~~ \mbox{for all $j
    \in \setsignalques$} \Big) & \geq 1 - e^{-\UHP \log^{1.5}(\numques
    \numwork)}.
\end{align}
  \end{subequations}
\end{theorem}
}
We provide the proof of Theorem~\ref{ThmWANnew}  in Section~\ref{SecProofThmWANnew}. \rev{At a high level, the theorem says that all questions that have some reasonable signal are estimated correctly by the WAN estimator. To gain intuition behind the notion of signal in~\eqref{EqnDefnSetquesWAN}, let us consider $\pp=1$ and consider the majority voting algorithm (that is, taking a majority vote over all $\numwork$ workers).  A straightforward application of Hoeffding's inequality yields that for any question $j \in [\numques]$, the condition $\sum_{i=1}^{\numwork} (\probmxstar_{ij} - \frac{1}{2}) = \widetilde{\Omega}(\sqrt{n})$ is sufficient for the majority voting estimator to estimate $\ansstar_j$ correctly (with high probability). Furthermore, in the appendix, we also show that there exist matrices $\probmxstar$ where this condition is also necessary. Theorem~\ref{ThmWANnew} says that the WAN algorithm can estimate a question correctly if there exists some subset of ``top'' workers (according to $\permwork$), such that this condition for majority voting applies when restricted to only the answers from these workers. 
}

\rev{
	While Theorem~\ref{ThmWANnew} (as well as other results in the sequel) focuses on exact recovery with high probability, we note that alternatively directly bounding the expected Hamming error may yield guarantees that go beyond what is captured in this result. We leave this interesting problem for future work.
}

\rev{
Theorem~\ref{ThmWANnew} has an immediate corollary, one which provides
guarantees on the WAN estimator in terms of certain norms of the
matrix $\probmxstar$ which may be more interpretable, and also
provides a guarantee on the $\probmxstar$-loss incurred by the WAN
estimator. Again, these results hold in the regime
\eqref{EqnMoreHalf}$\cap$\eqref{EqnRegime}.

\begin{corollary}
\label{CorWAN}
\begin{subequations}
For any matrix $\probmxstar \in \classPerm$ and any binary vector
$\ansstar \in \{-1,1\}^{\numques}$, suppose that the WAN estimator is
provided with the permutation $\permwork$ of workers. Then for every
question $j \in [\numques]$ such that
\begin{align}
\label{EqnWANConditions}
\Lnorm{\probmxstar_j - \half}{2}^2 \geq \frac{5 \log^{2.5} (\numques
  \numwork)}{\pp}, \quad \mbox{and} \quad \Lnorm{\probmx^\permwork_j -
  \probmx^{\permworkstar}_j}{2} \leq \frac{\Lnorm{\probmxstar_j -
    \half}{2}}{\sqrt{9 \log (\numques \numwork)}},
\end{align}
we have
\begin{align}
\label{EqnWanCorWorks}
\mprob( [\ansWAN]_j = \ansstar_j ) \geq 1 - e^{- \UHP \log^{1.5}
  (\numques \numwork)}.
\end{align}
Consequently, if $\permwork$ is the correct permutation of the
workers, then with probability at least \mbox{$1 - e^{- \UHP'
    \log^{1.5} (\numques \numwork)}$,} we have
\begin{align}
\label{EqnCalibrated}
\Qloss{\probmxstar}{\ansWAN}{\ansstar} \leq \UUP
\frac{1}{\numwork \pp} \log^{2.5} \numques.
\end{align} 
\end{subequations}
\end{corollary}
}
\noindent Please see Section~\ref{SecProofCorWAN} for the proof of
Corollary~\ref{CorWAN}.

The conditions~\eqref{EqnWANConditions} required for
the result of Corollary~\ref{ThmWANnew} are sharp up to logarithmic
factors \rev{in the following sense}. The required approximation guarantee
$\Lnorm{\probmx^\permwork_j - \probmx^{\permworkstar}_j}{2} \leq
\frac{\Lnorm{\probmxstar_j - \half}{2}}{\sqrt{9 \log (\numques
    \numwork)}}$, if weakened to $\Lnorm{\probmx^\permwork_j -
  \probmx^{\permworkstar}_j}{2} \leq 2\Lnorm{\probmxstar_j -
  \half}{2}$, would allow for any arbitrary permutation
$\permwork$. This is because every permutation $\permwork$ satisfies
$\Lnorm{\probmx^\permwork_j - \probmx^{\permworkstar}_j}{2} \leq
\Lnorm{\probmx^\permwork_j - \half}{2} + \Lnorm{
  \probmx^{\permworkstar}_j - \half}{2} = 2\Lnorm{\probmxstar_j -
  \half}{2}$. Secondly, there exist constants $\plaincon_0 > 0$ and
$\ULOW > 0$ such that if one were guaranteed a lower bound of only
$\frac{\plaincon_0}{\pp}$ on $\Lnorm{\probmxstar_j - \half}{2}^2$
instead of the stated condition of $\frac{5 \log^{2.5} (\numques
  \numwork)}{\pp}$, then there exists a $\probmxstar \in \classDS$
satisfying this weaker condition such that any estimator $\anshat$
incurs an error at least $\mprob(\anshat_j \neq \ansstar_j) \geq
\ULOW$. Furthermore, this lower bound holds not only when the ordering
of workers is exactly known, but even when the entire matrix
$\probmxstar$ is known. The proof for this claim follows from the
construction in the proof of Theorem~\ref{ThmMinimax}(b).

At this point, we recall from Theorem~\ref{ThmMinimax}(b) the lower
bound on the estimation error in the $\probmxstar$-loss for any
estimator.  This lower bound applies to estimators that know not only
the ordering of the workers, but also the entire matrix
$\probmxstar$. This lower bound matches the upper
bound~\eqref{EqnCalibrated} of Corollary~\ref{CorWAN}, and the two
results in conjunction imply that the bound~\eqref{EqnCalibrated} is
sharp up to logarithmic factors.

\rev{ We conclude this section with a key insight
  obtained from our analysis of the WAN estimator.
\begin{remark}[Insight for unknown worker ordering problem]
	\label{RemInsightWAN}
The aforementioned results for the WAN algorithm have the following
useful implication for the setting when the ordering of workers is
{unknown}, under either of the models $\classDS$ or $\classPerm$.
For any matrix $\probmxstar \in \classPerm$, there exists a set of
workers $\setexpertworkers_{\probmxstar} \subseteq [\numwork]$ such
that the majority vote of the answers of the workers in
$\setexpertworkers_{\probmxstar}$ incurs a small risk. Consequently,
it suffices to design an estimator that identifies a set of good
workers and computes a majority vote of their answers. The estimator
need not attempt to infer the values of the entries of $\probmxstar$,
as is otherwise required, for instance, to compute maximum likelihood
estimates.
\end{remark}
\noindent The estimator proposed in the next section is based on the
observation in Remark~\ref{RemInsightWAN}.  }


\subsection{The \obiwan estimator}
\label{ObiWan}

In this section, we return to the setting where the ordering of the
workers is \emph{unknown}. We begin by presenting a computationally efficient estimator. 

Our proposed estimator operates in two steps. The first step performs
an Ordering Based on Inner-products (OBI), that is, 
computes an
ordering of the workers based on an inner product with the data. The
second step calls upon the WAN estimator from
Section~\ref{SecCalibrated} with this ordering. We thus term our
proposed estimator as the \obiwan estimator, $\ansOBIWAN$. In order to
make its description precise, we augment the notation of the WAN
estimator $\ansWAN$ to let $\ansWAN[\permwork,\obs]$ to denote the
estimate given by $\ansWAN$ operating on $\obs$ when given the
permutation $\permwork$ of workers.

An important technical issue is that re-using the observe data $Y$ to both determine an appropriate ordering of workers as well as to
estimate the desired answers, results in a violation of important
independence assumptions.  We resolve this difficulty by partitioning
the set of questions into two sets, and using the ordering estimated
from one set to estimate the desired answers for the other set and
vice versa. We provide a careful error analysis for this
partitioning-based estimator in the sequel. In more precise terms, the
\obiwan estimator $\ansOBIWAN$ is defined by the following three
steps:
\begin{itemize}[leftmargin=*]
\item Step $0$ (preliminary): Split the set of $\numques$ questions
  into two sets, $\setques_0$ and $\setques_1$, with every question
  assigned to one of the two sets uniformly at random. Let $\obs_0$
  and $\obs_1$ denote the corresponding submatrices of $\obs$,
  containing the columns of $\obs$ associated to questions in
  $\setques_0$ and $\setques_1$ respectively.
\item Step $1$ (OBI): For $\ell \in \{0,1\}$, let 
\begin{align*}
\eigvecobs_\ell \in
  \argmax_{\Lnorm{\eigvecobs}{2}=1} \Lnorm{\obs_\ell^T \eigvecobs}{2}
  \end{align*}
  denote the top eigenvector of $\obs_\ell \obs_\ell^T$; in order to
  resolve the global sign ambiguity of eigenvectors, we choose the
  global sign so that $\sum_{i \in [\numwork]} [\eigvecobs_\ell]_i^2
  \indicator{[\eigvecobs_\ell]_i > 0} \geq \sum_{i \in [\numwork]}
            [\eigvecobs_\ell]_i^2 \indicator{[\eigvecobs_\ell]_i <
              0}$.  Let $\permwork_\ell$ be the permutation of the
            $\numwork$ workers in order of the respective entries of
            $\eigvecobs_\ell$ (with ties broken arbitrarily).
\item Step $2$ (WAN): Compute the quantities
\begin{align*}
\ansOBIWAN(\setques_0) \defn \ansWAN[\obs_0,\permwork_1], \quad
\mbox{and} \quad \ansOBIWAN(\setques_1) \defn
\ansWAN[\obs_1,\permwork_0],
\end{align*}
corresponding to estimates of the answers for questions in
the sets $\setques_0$ and $\setques_1$,
respectively.
\end{itemize}

This completes the description of the OBI-WAN algorithm. 

We note that with regard to the use of the singular vectors of the
observed data in the OBI step, previous
works~\cite{karger2011iterative, ghosh2011moderates,
  parisi2014ranking,zhang2014spectral,jaffe2015estimating} also use
singular vectors to estimate properties of the underlying parameters
in crowdsourcing.  In these previous works, this step is motivated by
the fact that the spectrum of the population matrix $\mathbb{E}[\obs
  \obs^T]$ (or its mean-centered counterpart), can be related to the
parameters that underlie the model.

\rev{In the next three subsections, we provide guarantees for our
  \obiwan estimator under three model classes. Importantly, the
  guarantees for \obiwan hold \emph{simultaneously} for all model
  classes, and the estimator does not know the
  true class to which the data actually belongs.}


\subsubsection{Guarantees for \obiwan under an intermediate model}

In addition to the Dawid-Skene and the permutation-based models
introduced earlier, we study the estimation problem in an intermediate
model that lies between these two models. This intermediate model
introduces a parameter $\hardstar_j \in [0,1]$ that captures the
difficulty of each question $j \in [\numques]$, along with parameters
$\probInt \in \reals^{\numwork}$ associated with the workers as in the
Dawid-Skene model. Under this intermediate model, the probability that
worker $i \in [\numwork]$ correctly answers question $j \in
[\numques]$ (when the worker is asked the question) is given by
\begin{align}
\label{EqnDefnInter}
\mprob (\obs_{ij} = \ansstar_{j}) = \probInt_i (1 - \hardstar_j ) +
\half \; \hardstar_j,~~~\forall~(i,j)~\mathrm{such~that}~\obs_{ij}
\neq 0.
\end{align}
Intuitively, the parameter $\hardstar_j$ corresponds to the difficulty
of question $j$. When $\hardstar_j = 1$, the worker is purely
stochastic and provides random guesses, while for smaller values of
$\hardstar_j$ the worker is more likely to provide a correct answer.

This modeling assumption leads to the class
\begin{align*}
\classInter \defn \left \{ \probmx = \probInt (1-\hard)^T + \half \;
\ones \hard^T \mid \mbox{for some } \probInt \in [0, \:
  1]^\numwork,~ \hard \in [0,1]^\numques \right \}.
\end{align*}
{Note that we have the nested relation $\classDS \subset \classInter$; the Dawid-Skene model is a special case of
$\classInter$ corresponding to $\hard = 0$. In the regime~\eqref{EqnMoreHalf}, we further have $\classDS \subset \classInter
\subset \classPerm$.}

Up to a bijective transformation of the parameters, the
  model~\eqref{EqnDefnInter} is identical to a recent model proposed
  independently by Khetan and Oh~\cite{khetan2016reliable}, where the
  probability of a correct answer is assumed to be \mbox{$\probInt_i
    (1 - \hardstar_j ) + (1 - \probInt_i) \hardstar_j$}. The two
  models however arise from different conceptual motivations: Khetan
  and Oh consider the probability of correctness as a convex
  combination of the worker's behavior $\probInt_i$ and the opposite
  behavior $(1-\probInt_i)$, whereas our consideration of rarity of
  adversarial behavior leads to the probability of correctness set as
  a convex combination of the worker's behavior $\probInt_i$ and
  random responses $\half$.

We now provide exact-recovery guarantees for the \obiwan estimator
under this intermediate model. As with our other results, the
following theorem applies to the
regime~\eqref{EqnMoreHalf}$\cap$\eqref{EqnRegime}: 
 \rev{
  \begin{theorem}
    \label{ThmOBIWAN}
Consider any binary vector $\ansstar \in \{-1,1\}^{\numques}$ and any
matrix $\probmxstar \in \classInter$ associated with vectors $(
\probInt,\hard)$ satisfying $\Lnorm{ \probInt - \half}{2}^2 \Lnorm{1 -
  \hardstar}{2}^2 \geq { \frac{\capconst \numques \log^{2.5} (\numques
    \numwork)}{\pp}}$ for a large enough constant $\capconst$.  Then
for every question $j \in [\numques]$ such that
\begin{subequations}
  \begin{align} 
    (1 - \hardstar_j)^2	\Lnorm{\probInt - \half
    }{2}^2 \geq \frac{5 \log^{2.5} (\numques \numwork)}{\pp},
  \end{align}
  we have 
  \begin{align}
    \mprob( [\ansOBIWAN]_j = \ansstar_j ) \geq 1 - e^{- \UHP \log^{1.5}
      (\numques \numwork)}.
  \end{align}
\end{subequations}
  \end{theorem}		
\noindent Please see Section~\ref{SecProofThmOBIWANa} for the proof of this
theorem.  See also Theorem~\ref{ThmOBIHamDS} in the sequel, which
provides a matching lower bound (up to logarithmic factors) for the
special case of $\hardstar = 0$.
		
We now provide some intuition about the OBI part of the \obiwan estimator, and we do so in the context of Theorem~\ref{ThmOBIWAN}. For simplicity in this explanation, let us ignore the sample splitting step (Step 0) of the \obiwan algorithm and assume the OBI step (Step 1) is applied to the entire observed data $\obs$. Then under the model $\classInter$, we can rewrite the observation matrix as $\obs = \pp (2\probInt- \ones) (1 -\hard)^T \diag{\ansstar} + \noise$, where $\noise$ is a ``noise'' matrix and $\pp (2\probInt- \ones) (1 -\hard)^T \diag{\ansstar}$ is the ``signal'' in the observed data. In this representation, the signal is a matrix of rank one,  its top left singular vector equals $2\probInt- \ones$ (up to a scaling), and the ``magnitude of the signal'' is \mbox{$\opnorm{\pp (2\probInt- \ones) (1 -\hard)^T \diag{\ansstar} } = \pp \Lnorm{2\probInt- \ones}{2} \Lnorm{1 -\hard}{2}$}. Furthermore, we show in the proof of Theorem~\ref{ThmOBIWAN} that the ``magnitude of the noise''   is bounded as $\opnorm{\noise} \leq \plaincon \sqrt{ \numques \pp\ \polylog{\numques\numwork}}$ with high probability. Consequently when the magnitude of signal exceeds the noise (condition stated in the beginning of Theorem~\ref{ThmOBIWAN}), the top left singular vector of $\obs$ approximately captures the ordering of entries in $(2\probInt- \ones)$ that represent the worker abilities.
		
		The following corollary now upper bounds the $\probmxstar$-loss for the \obiwan estimator under the intermediate model in the regime \eqref{EqnMoreHalf}$\cap$\eqref{EqnRegime}:
\begin{corollary}
	\label{CorCInt}
	For any
	$\probmxstar \in \classInter$ and any vector \mbox{$\ansstar \in
		\{-1,1\}^{\numques}$,} the estimate $\ansOBIWAN$ has error at most
			\begin{align}
			\label{EqnOBIWANCIntQstar} 
			\Qloss{\probmxstar}{\ansOBIWAN}{\ansstar} \leq \UUP\frac{ 1}{\numwork
				\pp}\log^{2.5} \numques,
			\end{align}
			with probability at least $1 - e^{-\UHP \log^{1.5} (\numques
				\numwork)}$.  
		\end{corollary}
}
\noindent Please see Section~\ref{SecProofCorCInt} for the proof of this
corollary.  A comparison with the lower bound of
Theorem~\ref{ThmMinimax}(b) reveals that the
bound~\eqref{EqnOBIWANCIntQstar} is tight up to logarithmic factors.


\subsubsection{Guarantees for \obiwan under the Dawid-Skene model}

In this section, we present results relating the performance of the
\obiwan estimator under the Dawid-Skene model. Unlike the rest of the paper, in this section the simplicity of the model allows us to generalize in another direction: handling adversarial
workers, that is,  \emph{not} being restricted to regime~\eqref{EqnMoreHalf} and allowing $\probDS_i <
\half$ for some workers $i \in [\numwork]$. 

We introduce some additional notation. For the vector $\probDS \in [0,1]^\numwork$, we
define two associated vectors $\probDSplus, \probDSminus \in
[0,1]^\numwork$ as $\probDSplus_i = \max \{\probDS_i, \half\}$ and
$\probDSminus_i = \min \{\probDS_i, \half\}$ for every $i \in
[\numwork]$. Then we have $(\probDS - \half) = (\probDSplus- \half) +
(\probDSminus - \half)$, with $\probDSplus$ representing normal
workers and $\probDSminus$ representing adversarial workers who are more
inclined to provide incorrect answers.  The
following result holds in the regime~\eqref{EqnRegime}:
\begin{theorem}
\label{ThmOBIHamDS}
Consider any Dawid-Skene matrix of the form $\probmxstar = \probDS
\ones^T$ for some $\probDS \in [0,1]^\numwork$. Then:
\begin{enumerate}[leftmargin=*]
\item[(a)] If $\Lnorm{\probDSplus - \half}{2} \geq \Lnorm{\probDSminus
  - \half}{2} + \sqrt{\frac{4 \log^{2.5} (\numques \numwork)}{\pp}}$
  and $ (\probDS - \half)^T \ones \geq 0$, then for any $\ansstar \in
  \{-1,1\}^{\numques}$, the \obiwan estimator satisfies
\begin{subequations}
\begin{align}
\mprob( \ansOBIWAN \; = \; \ansstar) \geq 1- e^{- \UHP \log^{1.5}
  (\numques \numwork)}.
\end{align}
\item[(b)] Conversely, there exists a positive universal constant
  $\plaincon$ such that for any $\probDS \in
  [\frac{1}{10},\frac{9}{10}]^\numwork$ with $\Lnorm{\probDS - \half}{2} \leq
  \sqrt{\frac{\plaincon}{\pp}}$, any estimator $\anshat$ has (normalized) Hamming
  error at least
\begin{align}
\sup_{\ansstar \in \{-1,1\}^\numques} \Exs\big[ \sum_{i=1}^{\numques}  \frac{1}{\numques}  \indicator{ \anshat_i \neq \ansstar_i} \big] \geq \frac{1}{10}.
\end{align}
\end{subequations}
\end{enumerate}
\end{theorem}

\noindent The proofs of  the two parts of Theorem~\ref{ThmOBIHamDS} are provided in Sections~\ref{SecProofThmOBIHamDSa} and~\ref{SecProofThmOBIHamDSb}. \\

{ A couple of remarks are in order. For the following
  discussion, consider the two mild conditions $\Lnorm{\probDSplus -
    \half}{2} \geq 1.01 \Lnorm{\probDSminus - \half}{2}$ and $(\probDS
  - \half)^T \ones > 0$.  We claim that under these mild
  conditions, the OBI-WAN estimator is optimal up to logarithmic
  factors. To see this, first observe that the lower bound in
  Theorem~\ref{ThmOBIHamDS}(b) implies that for any non-trivial
  recovery guarantee to hold, it must be the case that $\Lnorm{\probDS
    - \half}{2} > \sqrt{\frac{c}{\pp}}$ for some positive universal
  constant $c$.  Now suppose that $\Lnorm{\probDS - \half}{2} >
  \sqrt{\frac{c' \log^{2.5} (\numques \numwork)}{\pp}}$ for a large
  enough positive constant $c'$; observe that this condition is only a
  logarithmic factor away from the necessary condition. Then under the
  mild aforementioned conditions, we have $\Lnorm{\probDSplus -
    \half}{2} \geq \Lnorm{\probDSminus - \half}{2} + \sqrt{\frac{4
      \log^{2.5} (\numques \numwork)}{\pp}}$. Part (a) of
  Theorem~\ref{ThmOBIHamDS} then guarantees that the OBI-WAN estimator
  recovers the true answers $\ansstar$ with high probability.
}


Secondly, an application of Theorem~\ref{ThmOBIHamDS} is to the setting that has been
the focus of our paper, where we have no adversarial workers. In this
case, we have \mbox{$\probDSminus = 0$} and \mbox{$\probDSplus =
  \probDS$,} and the upper and lower bounds match upto a logarithmic
factor.  The upper bound shows that when \mbox{$\Lnorm{\probDS-
    \half}{2} \geq \sqrt{\frac{4 \log^{2.5} (\numques
      \numwork)}{\pp}}$,} the Hamming error is vanishingly small,
whereas the lower bound shows that there is a universal constant $c$
such that the Hamming error is essentially as large as possible when
$\Lnorm{\probDS- \half}{2} \leq \sqrt{\frac{c}{\pp}}$.


\subsubsection{Guarantees for \obiwan under the permutation-based model }
\rev{
The previous two subsections provided strong guarantees for \obiwan for exact recovery and the $\probmxstar$-loss under the Dawid-Skene and intermediate models. A natural question that arises then is how robust is \obiwan to mismatches with respect to the Dawid-Skene and intermediate models. We analyze \obiwan under the considerably richer permutation-based model class in this section. 
\begin{proposition}
  \label{PropOBIWANCperm}
  Consider any matrix $\probmxstar \in \classPerm$ and any binary
  vector $\ansstar \in \{-1,1\}^{\numques}$. For every question $j \in
  [\numques]$ that such that
  \begin{subequations}
    \begin{align}
      \label{EqnPropOBIWANcondition} 
      \sum_{i=1}^{\numwork} (\probmxstar_{i j} -  \half)
      \geq  \frac{3}{4} \sqrt{\frac{\numwork}{\pp} \log^{1.5} (\numques
	\numwork)},
    \end{align}
    the \obiwan estimator satisfies
    \begin{align}
      \label{EqnPropOBIWANexact} 
      \mprob ( [\ansOBIWAN]_{j} = \ansstar_{j} ) & \geq 1 - e^{-\UHP \log^{1.5}(\numques
	\numwork)}.
    \end{align}
  \end{subequations}
Consequently for any $\probmxstar \in \classPerm$ and any
\mbox{$\ansstar \in \{-1,1\}^{\numques}$,} with probability at least
$1 - e^{-\UHP \log^{1.5} (\numques \numwork)}$, the estimator incurs a
$\probmxstar$-loss of at most
\begin{align}
  \label{EqOBIWANQstarCPerm}
  \Qloss{\probmxstar}{\ansOBIWAN}{\ansstar} \leq \UUP\frac{
    1}{\sqrt{\numwork \pp}}\log \numques.
\end{align}
\end{proposition}
\noindent See Section~\ref{SecProofPropOBIWANPerm} for the proof of this
result.


It is well known that the majority voting estimator is highly robust
to model specification (see, for instance, the discussion
in~\cite[Section 4.2]{gao2013minimax}); this robustness perhaps
underlies its popularity in practice.  In the appendix, we show that
the majority voting estimator achieves a rate $\Omega(\frac{
  1}{\sqrt{\numwork \pp}})$ in terms of the $\probmxstar$-loss in the
worst case over the permutation-based model. Thus the
guarantee~\eqref{EqOBIWANQstarCPerm} for \obiwan matches the lower bound for 
majority voting. However, importantly, in addition to this guarantee over
$\classPerm$, the \obiwan estimator \emph{simultaneously} also
achieves the strong exact recovery and $\probmxstar$-loss guarantees of Theorem~\ref{ThmOBIWAN},
Corollary~\ref{CorCInt}, and Theorem~\ref{ThmOBIHamDS} over the
simpler models $\classInter$ and $\classDS$.
}


\section{Experiments}
\label{SecExperiments}

In this section, we report the results of a suite of experiments, on
both synthetic and real-world data, so as to evaluate the \obiwan
estimator which was introduced in Section~\ref{ObiWan}. We compare
\obiwan to the Spectral-EM estimator due to Zhang et
al.~\cite{zhang2014spectral}, which to the best of our knowledge, has
the strongest established guarantees in the literature.  For the
Spectral-EM estimator, we used an implementation provided by the
authors of the paper~\cite{zhang2014spectral}.  The code for the
\obiwan estimator as well as the constituent WAN estimator is freely
available on the first author's website.


\subsection{Simulations} 
\label{SecSimulations}

\begin{figure}[ht!]
\centering
\begin{subfigure}{.03\textwidth}
\includegraphics[width=\textwidth]{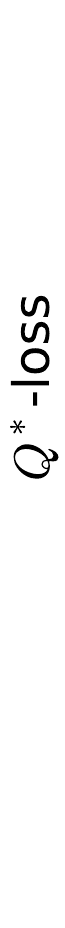}
\end{subfigure}
\begin{subfigure}{0.3\textwidth}
\includegraphics[width = \textwidth]{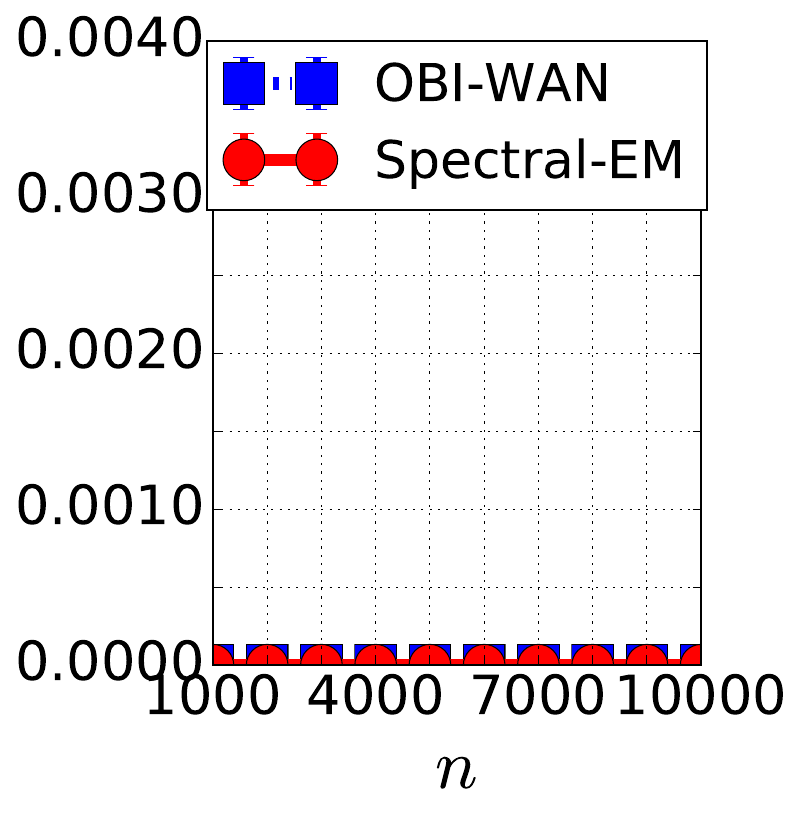}\label{FigSimulationsEasy}
\vspace{-.6cm}
\caption{Easy}
\end{subfigure}~~
\begin{subfigure}{0.3\textwidth}
\includegraphics[width = \textwidth]{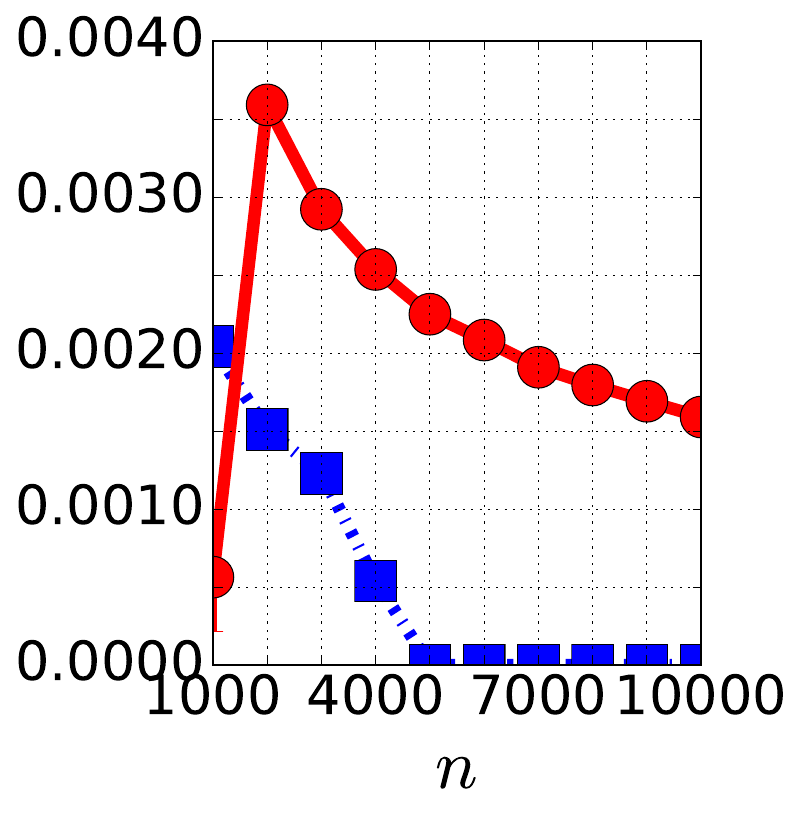}\label{FigSimulationsBreakMajorityVoting}
\vspace{-.6cm}
\caption{Few smart}
\end{subfigure}~~
\begin{subfigure}{0.29\textwidth}
\includegraphics[width = \textwidth]{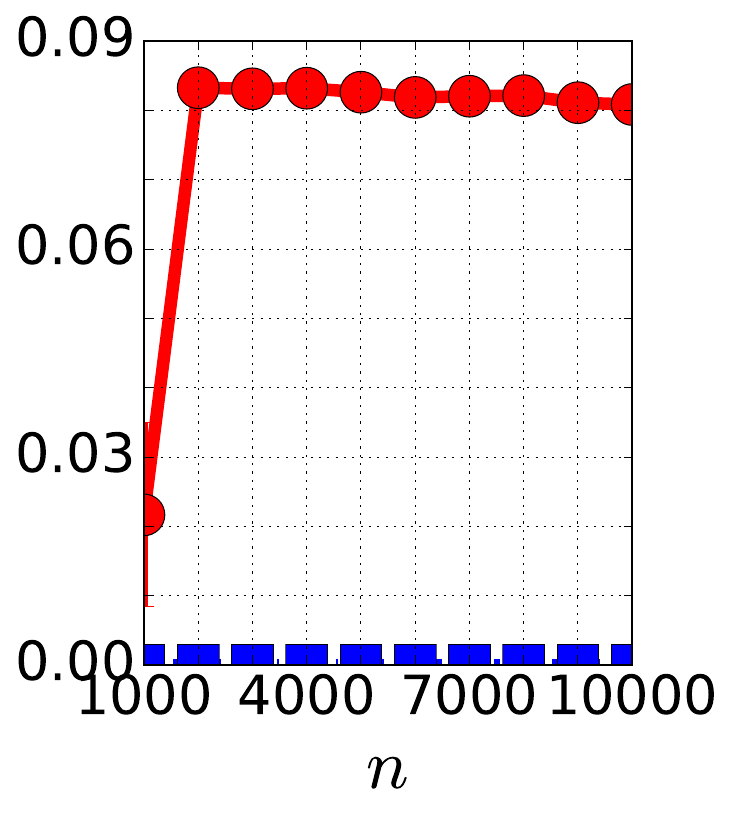}\label{FigSimulationsAdversarial}
\vspace{-.6cm}
\caption{Adversarial}
\end{subfigure}\\
\vspace{-1cm}
\begin{subfigure}{.03\textwidth}
\includegraphics[width=\textwidth]{fig_ylabel}
\end{subfigure}
\begin{subfigure}{0.3\textwidth}
\includegraphics[width =
  \textwidth]{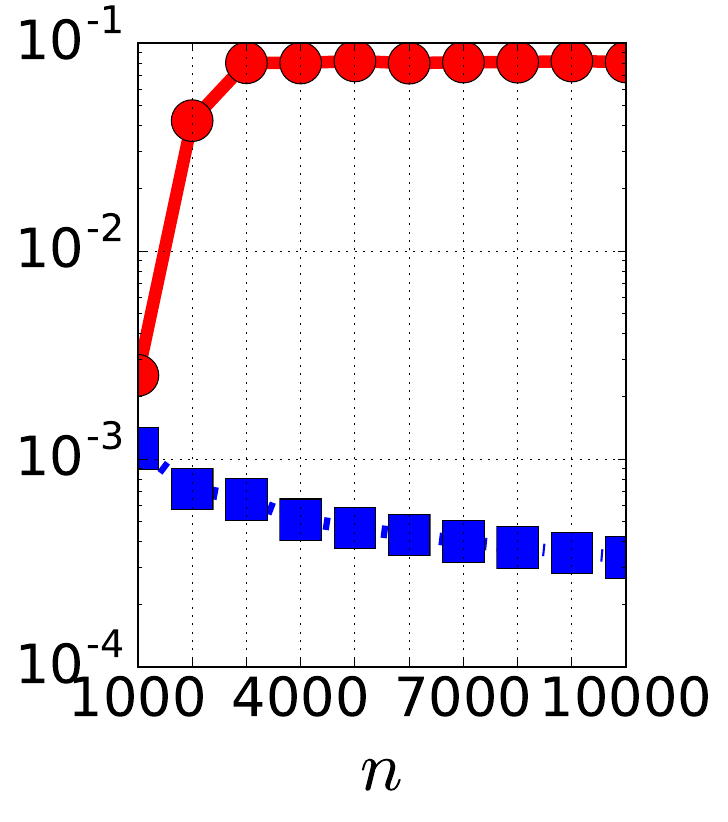}\label{FigSimulationsBreakDS}
\vspace{-.6cm}
\caption{In $\classPerm \backslash \classInter$}
\end{subfigure}~~
\begin{subfigure}{0.3\textwidth}
\includegraphics[width =
  \textwidth]{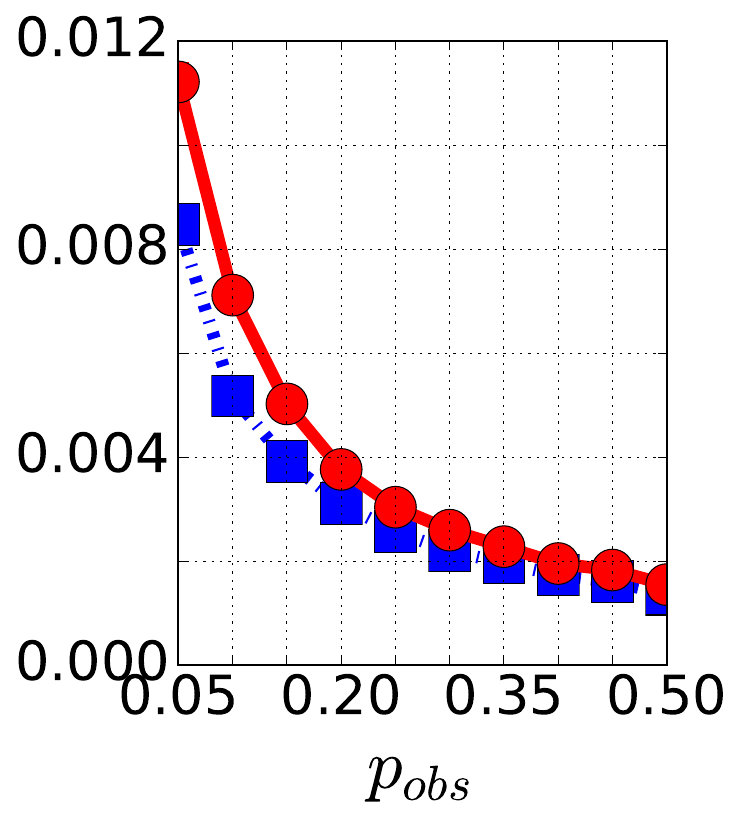}\label{FigSimulationsLowerBound}
\vspace{-.6cm}
\caption{Minimax lower bound}
\end{subfigure}~~
\begin{subfigure}{0.3\textwidth}
\includegraphics[width =
  \textwidth]{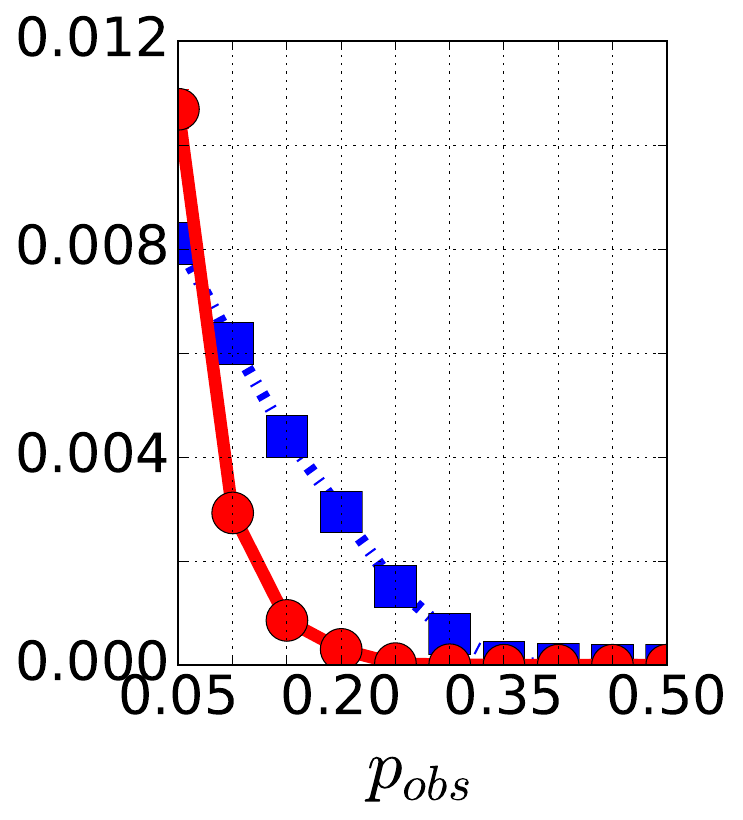}\label{FigSimulationsSuperSparse}
\vspace{-.6cm}
\caption{Super sparse}
\end{subfigure}
\caption{Results from numerical simulations comparing the Spectral-EM algorithm (circular markers) with \obiwan (square markers). The plots in panels
  (a)-(d) measure the $\probmxstar$-loss as a function of $\numwork$,
  and the plots in panels (e)-(f) measure the $\probmxstar$-loss as a
  function of $\pp$. Each point is an average of over $20$
  trials. Recall that when $\probmxstar$ follows the Dawid-Skene
  model, as in panels (a)-(c), (e)-(f), the Hamming error is
  proportional to the $\probmxstar$-loss. Also note that the Y-axis of
  panel (d) is plotted on a logarithmic scale.}
\label{FigSimulations}
\end{figure}

We first conduct synthetic simulations to evaluate various aspects of the algorithms. We conduct six sets of simulations as detailed below. The results from our simulations are plotted in
Figure~\ref{FigSimulations}. The plots in the six panels (a) through
(f) of the figure are discussed below.

\begin{enumerate}[itemindent=16pt,label=\!\!(\alph*),leftmargin=3pt]
\item \underline{Easy}: $\probmxstar = \probDS \ones^T \in \classDS$
  where $\probDS_i = \frac{9}{10}$ if $i < \frac{ \numwork}{2}$, and
  $\probDS_i = \half$ otherwise. The parameter $\numwork$ is varied,
  and the regime of operation is $(\numques = \numwork,\; \pp=1)$. In
  this setting, both estimators correctly recover
  $\ansstar$.
\item \underline{Few smart}: $\probmxstar = \probDS \ones^T \in
  \classDS$ where $\probDS_i = \frac{9}{10}$ if $i < \sqrt{\numwork}$,
  and $\probDS_i = \frac{1}{2}$ otherwise. The parameter $\numwork$ is
  varied, and the regime of operation $(\numques = \numwork,\;
  \pp=1)$. Even though the data is drawn from the Dawid-Skene model,
  the error of Spectral-EM is much higher than that of the \obiwan
  estimator. Recall that the \obiwan estimator has guarantees
  of recovery over the entire Dawid-Skene class, unlike the estimators
  in prior literature.
\item \underline{Adversarial}: $\probmxstar = \probDS \ones^T \in
  \classDS$ where $\probDS_i = \frac{9}{10}$ if $i < \frac{\numwork}{4} + \sqrt{\numwork}$, $\probDS_i = \frac{1}{10}$ if $i >
  \frac{3\numwork}{4}$, and $\probDS_i = \frac{1}{2}$ otherwise. The
  parameter $\numwork$ is varied, and the regime of operation is
  $(\numques = \numwork,\; \pp=1)$. This set of simulations moves
  beyond the assumption that the entries of $\probmxstar$ are lower
  bounded by $\frac{1}{2}$, and allows for adversarial workers. The
  \obiwan estimator is successful in such a setting as well.
\item \underline{In $\classPerm$ but outside $\classInter$}:
  $\probmxstar_{ij} = \frac{9}{10}$ if $(i < \sqrt{\numwork}$ or $j <
  \frac{\numques}{2})$, and $\probmxstar_{ij} = \frac{1}{2}$
  otherwise. The parameter $\numwork$ is varied, and the regime of
  operation is $(\numques = \numwork,\; \pp=1)$. Here we have
  $\probmxstar \in \classPerm \backslash \classInter$. The
  $\probmxstar$-loss incurred by the \obiwan
  estimator decays as $\frac{1}{\sqrt{\numwork}}$, whereas the $\probmxstar$-loss of Spectral-EM  remains a constant.
\item \underline{Minimax lower bound}: $\probmxstar = \probDS \ones^T
  \in \classDS$ where $\probDS_i = \frac{9}{10}$ if $i \leq
  \frac{5}{\pp}$ and $\probDS_i = \frac{1}{2}$ otherwise. The
  parameter $\pp$ is varied, and the regime of operation is $(\numques
  = 1000,\; \numwork=1000)$. This setting is the cause of the minimax
  lower bound of Theorem~\ref{ThmMinimax}(b). The error of both estimators, in this case, behaves in an almost identical manner with a
  scaling of $\frac{1}{\pp}$.
\item \underline{Super sparse}: $\probmxstar = \probDS \ones^T \in
  \classDS$ where $\probDS_i = \frac{9}{10}$ if $i \leq
  \frac{\numwork}{10}$ and $\probDS_i = \half$ otherwise. The parameter
  $\pp$ is varied, and the regime of operation is $(\numques = 1000,\;
  \numwork= 1000)$. We see that the \obiwan estimator incurs a relatively higher error
  when data is very sparse --- more generally, we have observed a
  higher error when $\pp = o( \frac{\log^2 (\numques
    \numwork)}{\numwork})$, and this gap is also reflected in our
  upper bounds for the \obiwan estimator in Theorem~\ref{ThmOBIWAN}(a)
  and Theorem~\ref{ThmOBIHamDS}(a) that are loose by precisely a
  polylogarithmic factor as compared to the associated lower bounds.
\end{enumerate}

The relative benefits and disadvantages of of the proposed \obiwan
estimator, as observed from the simulations, may be summarized as
follows. In terms of limitations, the error of \obiwan is higher than
prior works when $\pp$ is small (as observed in the super-sparse case)
or when $n$ and $d$ are small (for instance, less than $200$).  On the
positive side, the simulations reveal that the \obiwan estimator leads
to accurate estimates in a variety of settings, providing 
guarantees over the $\classDS$ and $\classInter$ classes, and
demonstrating significant robustness in more general settings in
comparison to the best known estimator in the literature.


\subsection{Real-world crowdsourcing data}

\begin{figure}[!t]
	\centering
	\includegraphics[width=.5\textwidth]{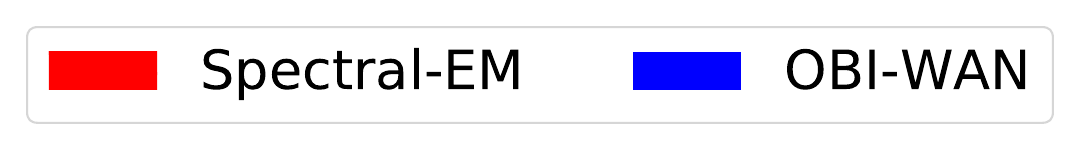}\\
	\rotatebox{90}{\hspace{-1.6cm}(Normalized) Hamming error}
	\begin{subfigure}{.2\textwidth}	
		\includegraphics[width =\textwidth]{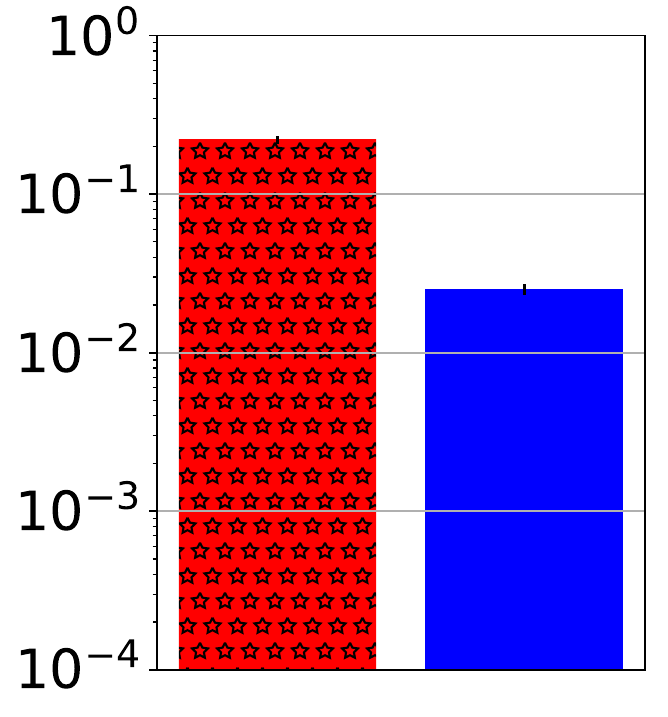}
		\caption{Dysplasia\label{FigExpCancer}}
	\end{subfigure}\quad
	\begin{subfigure}{.2\textwidth}	
		\includegraphics[width =\textwidth]{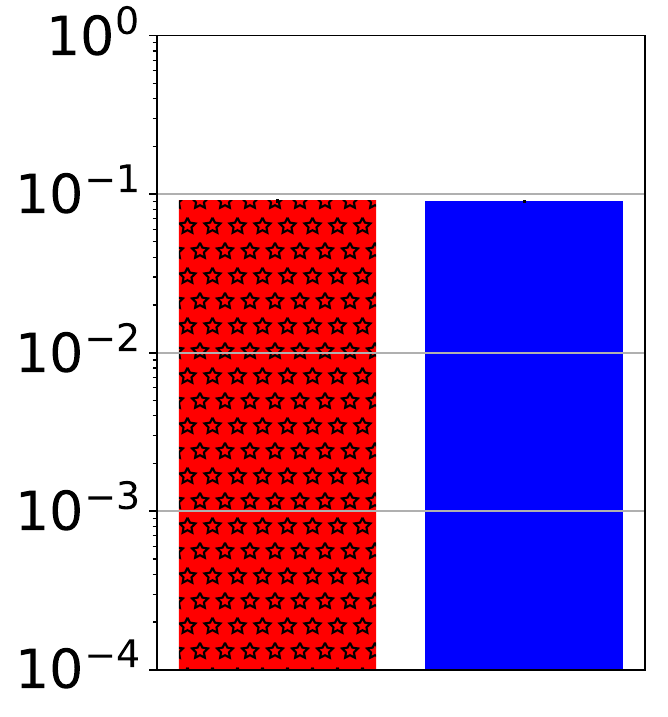}
		\caption{Bridges\label{FigExpBridges}}
	\end{subfigure}\quad
	\begin{subfigure}{.2\textwidth}	
		\includegraphics[width =\textwidth]{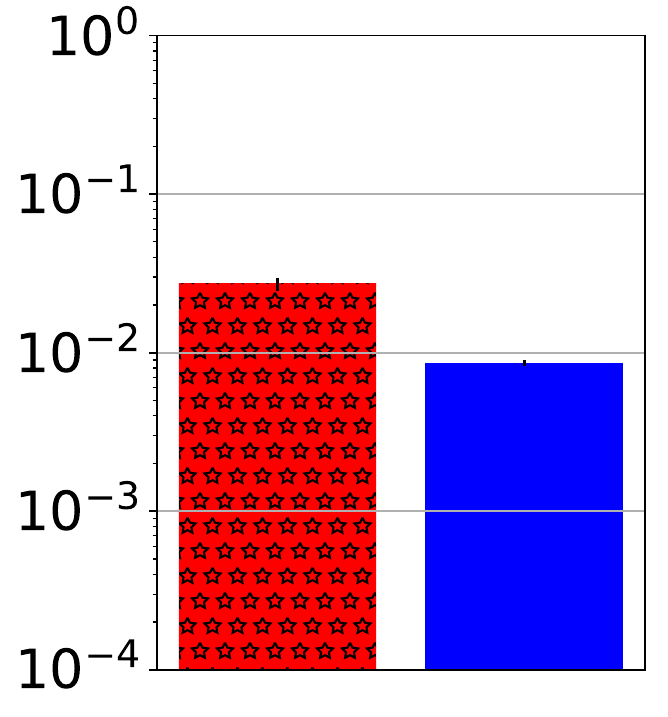}
		\caption{Dogs\label{FigExpDogs}}
	\end{subfigure}
	\\
	\rotatebox{90}{\hspace{-1.6cm}(Normalized) Hamming error}
	\begin{subfigure}{.2\textwidth}	
		\includegraphics[width =\textwidth]{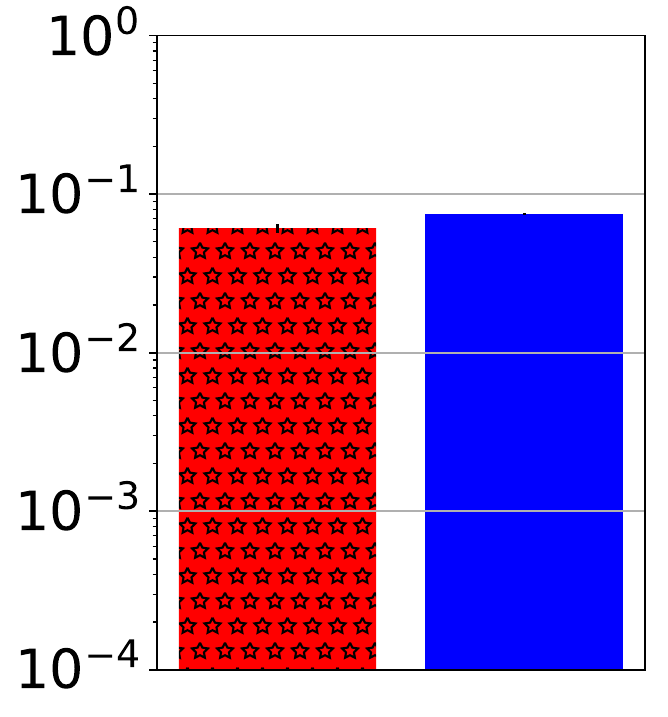}
		\caption{Flags\label{FigExpFlags}}
	\end{subfigure}\quad
	\begin{subfigure}{.2\textwidth}	
		\includegraphics[width =\textwidth]{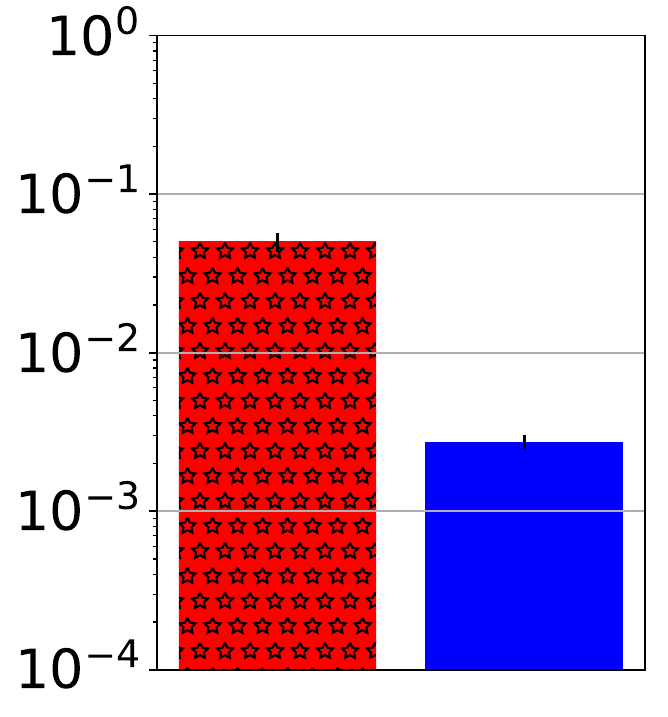}
		\caption{People\label{FigExpPeople}}
	\end{subfigure}\quad
	\begin{subfigure}{.2\textwidth}	
		\includegraphics[width =\textwidth]{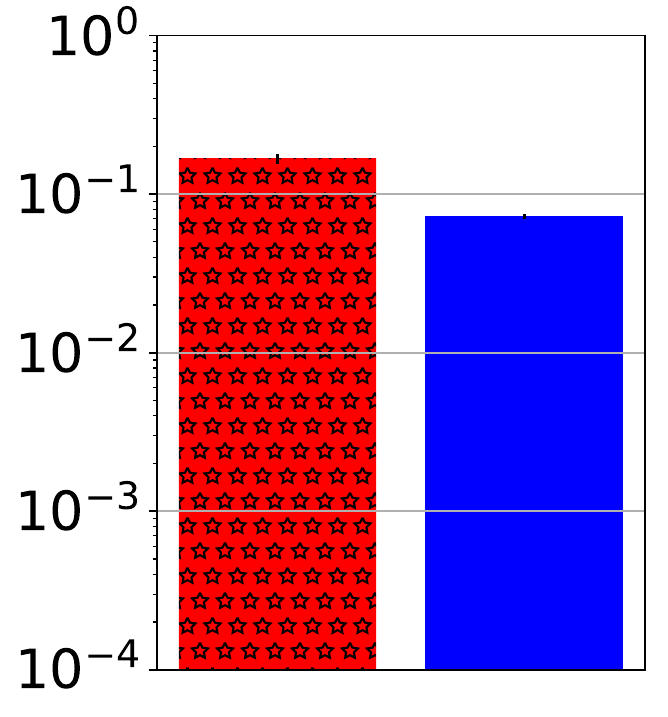}
		\caption{Textures\label{FigExpTextures}}
	\end{subfigure}
	\caption{Results from experiments on the Amazon Mechanical Turk crowdsourcing platform comparing the Spectral-EM algorithm (left bars) with OBI-WAN (right bars). The y-axes of the plots are on a logarithmic scale and represent the Hamming error normalized by the number of questions (lower is better).\label{FigExperiments}}
\end{figure}

In this section we describe a set of six experiments conducted using real-world data from the Amazon Mechanical Turk crowdsourcing platform,  ranging from visual recognition to knowledge elicitation. The experiments involved more than 200 workers in total. In each experiment, workers are asked to answer a number of questions, an we then employ statistical aggregation algorithms to estimate the ground truth answers. The results of these experiments are shown in Figure~\ref{FigExperiments}. As before, we compare the OBI-WAN estimator with the Spectral-EM estimator~\cite{zhang2014spectral}. 

We now describe more details regarding the experiments. The error bars in each figure represent the standard error of the mean. The plots and error bars shown in each figure are obtained via 300 iterations per experiment of subsampling the worker's answers with $\pp = 1/3$ and executing the two algorithms on the subsampled data.

\paragraph{(\subref{FigExpCancer}) Dysplasia}

The experiment was based on 48 pictures of (biological) cells to
workers. They had to classify each image as either ``Mild dysplasia''
where the ratio of the nucleus' area to cytoplasm's area is less than
33\%, or as ``severe dysplasia'' where the ratio of the nucleus' area
to cytoplasm's area is more than 33\%. The images and the ground truth
were obtained from the DTU/Herlev Pap Smear Database
2005~\cite{jantzen2005pap}. We collected responses from a total of 41
workers from Amazon Mechanical Turk, and Figure~\ref{FigExpCancer}
depicts the results of this experiment. The data is freely available
for download on the first author's website.

\paragraph{(\subref{FigExpBridges}) Bridges}
The experiment was based on 21 images of bridges, and the task for any
worker was to classify each image as either the golden bridge or
not. The data for this experiment was collected in past
work~\cite{shah2016doubleJMLR} from a total of 35 workers from Amazon
Mechanical Turk. Figure~\ref{FigExpBridges} depicts the results from
this data.

\paragraph{(\subref{FigExpDogs}) Dogs} The experiment comprised
85 images of dogs (from~\cite{stanfordDogsDataset,deng2009imagenet}),
and each worker was asked to identify the breed of each dog from ten
provided options. The data was collected in past
work~\cite{shah2016doubleJMLR} from a total of 35 workers from Amazon
Mechanical Turk. The data is converted to binary choice form by
choosing one breed uniformly at random in each iteration and
considering a binary-choice task of identifying whether or not the
dogs belong to this breed. Figure~\ref{FigExpDogs} plots the results
of the experiments.

\paragraph{(\subref{FigExpFlags}) Flags} In this experiment, each worker was shown a series of 126 flags. Each question required the worker to identify if a displayed flag belonged to a place in Africa, Asia/Oceania,  Europe, or neither of these. We use the data collected in the past work~\cite{shah2016doubleJMLR} which contains responses of 35 workers to all of the questions. We convert this data into binary choice format in the same manner as the dogs experiment above. Finally, we plot the results from this experiment in Figure~\ref{FigExpFlags}.

\paragraph{(\subref{FigExpPeople}) People} In this experiment, the names of 20 personalities were provided and the worker were asked to classify whether they were ever the President of the USA, President of India, Prime Minister of Canada, or  neither of these. Responses from 35 workers were collected in past work~\cite{shah2016doubleJMLR}, and we convert this data to binary choice as per the aforementioned procedure. Figure~\ref{FigExpPeople} plots the results of this experiment. 

\paragraph{(\subref{FigExpTextures}) Textures} As a final experiment, we evaluated the performance of the algorithms on workers' classification of textures. Specifically, workers were asked to classify 24 images from the dataset~\cite[Dataset 1: Textured surfaces]{lazebnik2005sparse} into one of eight possible textures. We use the responses of the workers collected in~\cite{shah2016doubleJMLR}, with a conversion to binary-choice as described above. The aggregate results from executing the two algorithms are depicted in Figure~\ref{FigExpTextures}.

All in all, the experiments reveal that OBI-WAN compares favorably to
Spectral-EM.


\section{Proofs}
\label{SecProofs}

In this section, we present the proofs of main theoretical
results. In the proofs, we use $\plaincon$, $\plaincon_1$,
$\plaincon'$ etc. to denote positive universal constants, and
ignore floors and ceilings unless critical to the proof. We assume
that $\numwork$ and $\numques$ are greater than some universal
constants; the case of smaller values of these parameters are then
directly implied by only changing the constant prefactors.


\subsection{Proof of Theorem~\ref{ThmMinimax}(a): Minimax upper bound}
\label{SecProofThmMinimaxa}

We begin by proving the minimax upper bound stated in part (a) of
Theorem~\ref{ThmMinimax}. The proof is divided into two parts, where
in the first part, we obtain an upper bound on the error term
$\frobnorm{(2 \probmxstar - \ones \ones^T) \diag{\ansstar} - (2
  \probmxlse - \ones \ones^T) \diag{\anslse} }^2$, following which we
convert this bound to one on
$\Qloss{\probmxstar}{\ansstar}{\anslse}$. 
follows along the lines of the proof of our previous
work~\cite[Theorem 1(a)]{shah2015stochastically}. In relation of our
present problem, one can think of the setting
of~\cite{shah2015stochastically} as that of estimating the matrix
$(2\probmxstar - \ones \ones^T)$ when the value of $\ansstar$ is
\emph{known}. The primary additional challenge in the first part of
the present proof is to accommodate the additional uncertainty about
$\ansstar$.

We begin with the first part of the proof, where we bound the error in estimating the product term $(2 \probmxstar - \ones \ones^T) \diag{\ansstar}$. Let us rewrite our observation
model in a ``linearized'' fashion that is convenient for subsequent
analysis.  In particular, let us define a random matrix $\noise \in
\reals^{\numwork \times \numques}$ with entries independently drawn
from the distribution
\begin{align}
\label{EqnDefnNoise}
\noise_{ij} =
\begin{cases}
1 - \pp (2 \probmxstar_{ij} - 1) \ansstar_{j} & ~~~ \mbox{w.p.~~~} \pp \Big( \probmxstar_{ij} \big( \frac{1 +
	\ansstar_{j}}{2} \big) + (1 - \probmxstar_{ij}) \big( \frac{1 -
	\ansstar_{j} }{2} \big) \Big) \\
-1 - \pp (2 \probmxstar_{ij} - 1) \ansstar_{j} & ~~~ \mbox{w.p.~~~}\pp \Big( \probmxstar_{ij} \big( \frac{1 -
	\ansstar_{j}}{2} \big) + (1 - \probmxstar_{ij}) \big( \frac{1 +
	\ansstar_{j} }{2} \big) \Big) \\
- \pp (2 \probmxstar_{ij} - 1) \ansstar_{j} & ~~~ \mbox{w.p.~~~} 1 - \pp,
\end{cases}
\end{align}
where ``w.p.'' is a shorthand for ``with probability''.
One can verify that $\Exs [ \noise ] = 0$, every entry of $\noise$ is
bounded by $2$ in absolute value, and moreover that our observed
matrix $\obs$ can be written in the form
\begin{align}
\label{EqnLinearized}
\frac{1}{\pp} \obs = (2\probmxstar - \ones \ones^T) \; \diag{\ansstar}
+ \frac{1}{\pp} \noise.
\end{align}

Let $\permallwork$ denote the set of all permutations of the
$\numwork$ workers, and let $\permallques$ denote the set of all
permutations of the $\numques$ questions. For any pair of permutations
$(\permwork, \permques) \in \permallwork \times \permallques$, define
the set
\begin{align*}
\classPerm(\permwork, \permques) \defn \Big \{ \probmx \in
[0,1]^{\numwork \times \numques} \mid & \probmx_{ij} \geq
\probmx_{i'j'} \mbox{ whenever } \\
& \permwork(i) \leq
\permwork(i') \mbox{ and } \permques(j) \leq \permques(j')
\Big \},
\end{align*}
corresponding to the subset of $\classPerm$ consisting of matrices
that are faithful to the permutations $\permwork$ and $\permques$.
For any fixed $\ans \in \{-1,1\}^{\numques}$, $\permwork \in \permallwork$ and $\permques \in \permallques$, define the matrix
\begin{align*}
\probmxtilde(\permwork, \permques, \ans) & \in \argmin \limits_{\probmx
	\in \classPerm(\permwork, \permques)} \Cost(\probmx, \ans), \\
\mbox{ where} \quad \Cost(Q, \ans) &\defn \frobnorm{\frac{1}{\pp} \obs
	- (2 \probmx - \ones \ones^T) \diag{\ans} }^2.
\end{align*}
Using this notation, we can rewrite the least squares
estimator~\eqref{EqnDefnLSE} in the compact form
\begin{align*}
(\anslse, \permworklse, \permqueslse) \in \argmin
\limits_{\substack{ (\permwork, \permques) \in \permallwork \times
		\permallques\\ \ans \in \{-1,1\}^\numques}}
\Cost(\probmxtilde(\permwork, \permques, \ans), \ans), \quad
\mbox{and} \quad \probmxlse = \probmxtilde(\permworklse,
\permqueslse, \anslse).
\end{align*}
For the purposes of analysis, let us define the set
\begin{align}
\label{EqnDefnPermSet}
\permset \defn \Big\{ (\permwork, \permques, \ans) \in \permallwork
\times \permallques \times \{-1,1\}^{\numques} \mid
\Cost(\probmxtilde(\permwork, \permques, \ans), \ans) \leq
\Cost(\probmxstar, \ansstar) \Big \}.
\end{align}

With this set-up, we claim that it is sufficient to show the following:
fix a triplet $(\permwork, \permques, \ans) \in \permset$,
for this fixed triplet there is a universal constant $\plaincon_1$ such that
\begin{align}
\label{EqnPartialPermReq}
\mprob \big( \frobnorm{(2 \probmxtilde(\permwork, \permques, \ans) -
	\ones \ones^T) \diag{\ans - \ansstar} }^2 \leq \plaincon_1
\frac{\numques}{\pp} \log^2 \numques \big) \geq 1 - e^{- 4 \numques
	\log (\numques \numwork)}.
\end{align}
Given
this bound, since the cardinality of the set $\permset$ is upper
bounded by $e^{3 \numques \log \numques}$ (since $\numques \geq
\numwork$), a union bound over all these permutations applied
to~\eqref{EqnPartialPermReq} yields
\begin{align*}
\mprob \big( \max_{(\permwork, \permques, \ans) \in \permset}
\frobnorm{(2 \probmxtilde(\permwork, \permques, \ans) - \ones \ones^T)
	\diag{\ans - \ansstar} }^2 \leq \plaincon_1 \frac{\numques \log^2
	\numques}{\pp}  \big) \geq 1 - e^{-  \numques \log (\numques \numwork)}.
\end{align*}
The set $\permset$ is guaranteed to be non-empty since the true
permutations $\permworkstar$ and $\permquesstar$ corresponding to
$\probmxstar$ and the true answer $\ansstar$ always lie in $\permset$,
and consequently, the above tail bound yields the claimed result.

The remainder of our analysis is devoted to proving the
bound~\eqref{EqnPartialPermReq}.  Given any triplet \mbox{$(\permwork,
	\permques, \ans) \in \permset$,} we define the matrices
\begin{align*}
\lseinterstar \defn (2\probmxstar - \ones \ones^T) \; \diag{\ansstar},
~~ \mbox{and} ~~ \lseintertilde(\permwork, \permques, \ans)
\defn (2\probmxtilde(\permwork, \permques, \ans) - \ones \ones^T) \;
\diag{\ans}.
\end{align*}
Henceforth, for brevity, we refer to the matrix
$\lseintertilde(\permwork, \permques, \ans)$ simply as
$\lseintertilde$ and the matrix $\probmxtilde(\permwork, \permques,
\ans)$ simply as $\probmxtilde$, since the values of the associated
quantities $(\permwork, \permques, \ans)$ are fixed and clear from
context.

Since $(\pi,\sigma,x) \in \mathcal{P}$, the definition of set
$\mathcal{P}$ in~\eqref{EqnDefnPermSet} yields the inequality
\begin{align*}
\frobnorm{\frac{1}{\pp} Y - (2 \widetilde{Q}(\pi,\sigma,x) - \ones
	\ones^T ) \diag{x}}^2 \leq \frobnorm{\frac{1}{\pp} Y - (2Q^* - \ones
	\ones^T ) \diag{x^*}}^2.
\end{align*}
Substituting the expression $\frac{1}{\pp} Y = (2 \probmxstar - \ones
\ones^T) \diag{\ansstar}+ \frac{1}{\pp} \noise$
from~\eqref{EqnLinearized}, we obtain the relations
\begin{align*}
\frobnorm{ & (2 \probmxstar - \ones \ones^T) \diag{\ansstar} +
	\frac{1}{\pp} W - (2 \widetilde{\probmx}(\pi,\sigma, \ans) - \ones
	\ones^T ) \diag{\ans}}^2 \\
& \leq \frobnorm{(2 \probmxstar - \ones \ones^T) \diag{x^*}+ \frac{1}{\pp}  W - (2Q^* - \ones \ones^T ) \diag{\ansstar}}^2 
= \frobnorm{ \frac{1}{\pp}  \noise }^2.
\end{align*}
Using the expansion $(a+b)^2 = a^2 + b^2 + 2ab$, and substituting the
expressions for $\lseinterstar$ and $\lseintertilde$, we obtain
following basic inequality
\begin{align}
\label{EqnLSEThmBasicIneq}
\half \frobnorm{\lseinterstar - \lseintertilde}^2 \leq \frac{1}{\pp}
\tracer{\lseintertilde- \lseinterstar}{\noise}.
\end{align}
The following lemma uses this inequality to obtain an upper bound on
the quantity $\half \frobnorm{\lseinterstar - \lseintertilde}^2$.
\begin{lemma}
	\label{LemBoundLSEinter}
	There exists a universal constant $\plaincon_1 > 0$ such that
	\begin{align}
	\label{EqnBoundLSEinter}
	\mprob \Big( \frobnorm{\lseinterstar - \lseintertilde}^2 \leq
	\plaincon_1 \frac{\numques \log^2 \numques}{\pp} \Big) \geq 1 - e^{- 4
		\numques \log (\numques \numwork )}.
	\end{align}
\end{lemma}
\noindent See Section~\ref{SecProofLemBoundLSEinter} for the proof of
this lemma. This completes the first part of the proof.

In the second part of the proof, we now convert our
bound~\eqref{EqnBoundLSEinter} on the Frobenius norm
$\frobnorm{\lseinterstar - \lseintertilde}$ into one on the error in
estimating $\ansstar$ under the $\probmxstar$-loss. The following
lemma is useful for this conversion:
\begin{lemma}
	\label{LemFrobToLoss}
	For any pair of matrices $A_1, A_2 \in \reals_+^{\numwork \times
		\numques}$ and any pair of vectors $v_1, v_2 \in
	\{-1,1\}^{\numques}$, we have
	\begin{align}
	\label{EqnFrobToLoss}
	\frobnorm{ A_1 \; \diag{v_1 - v_2} }^2 \leq 4 \frobnorm{ A_1 \;
		\diag{v_1} - A_2 \; \diag{v_2} }^2.
	\end{align}
\end{lemma}
\noindent See Section~\ref{SecProofLemFrobToLoss} for the proof of
this claim.

Recall our assumption that every entry of the matrices $\probmxstar$
and $\probmxtilde$ is at least $\half$. Consequently, we can apply
Lemma~\ref{LemFrobToLoss} with $A_1 = (\probmxstar - \half \ones
\ones^T)$, $A_2 = (\probmxtilde - \half \ones \ones^T)$, $v_1 =
\ansstar$ and $v_2 = \ans$ to obtain the inequality
\begin{align}
\frobnorm{ (\probmxstar - \half \ones \ones^T) \diag{\ansstar -
		\ans} }^2 & \leq 4 \frobnorm{ (\probmxstar- \half \ones \ones^T)  \diag{\ansstar} \!- \! (\probmxtilde - \half \ones \ones^T) 
	\diag{\anshat} }^2 \nonumber \\
& = 4 \frobnorm{ \lseinterstar - \lseintertilde}^2.
\end{align}
Coupled with Lemma~\ref{LemBoundLSEinter}, this bound yields the desired
result~\eqref{EqnPartialPermReq}.


\subsubsection{Proof of Lemma~\ref{LemBoundLSEinter}}
\label{SecProofLemBoundLSEinter}

Our proof of this lemma closely follows along the lines of the proof
of a related result in our past work~\cite{shah2015stochastically}.
Denote the error in the estimate as $\DelHat \defn \lseintertilde -
\lseinterstar$. Then from the inequality~\eqref{EqnLSEThmBasicIneq}, have
\begin{align}
\label{EqnBasicPartial}
\frac{1}{2} \frobnorm{\DelHat}^2 & \leq \frac{1}{\pp}
\tracer{\noise}{\DelHat}.
\end{align}
For the quadruplet $(\permwork, \permques, \ans, \lseinterstar)$ under
consideration, define the set
\begin{align*}
\lseinterdiff(\permwork, \permques, \ans, \lseinterstar) \defn \Big \{
\alpha( \lseinter - \lseinterstar) \mid & \lseinter = (2 \probmx - \ones
\ones^T) \diag{\ans},\\
& \probmx \in \classPerm(\permwork, \permques),
\ \alpha \in[0,1] \Big \}.
\end{align*}
Since the terms $\permwork$, $\permques$, $\ans$ and $\lseinterstar$
are fixed for the purposes of this proof, we will use the abbreviated
notation $\lseinterdiff$ for $\lseinterdiff(\permwork, \permques,
\ans, \lseinterstar)$.

For each choice of radius $t > 0$, define the random variable
\begin{subequations}
	\begin{align}
	\label{EqnDefnZPartial}
	Z(t) & \defn \sup_{ \substack{\chattDiffmx \in
			\lseinterdiff,\\ \frobnorm{\chattDiffmx} \leq t}} ~~
	\frac{1}{\pp} \tracer{\chattDiffmx}{\noise}.
	\end{align}
	Using the basic inequality~\eqref{EqnBasicPartial}, the Frobenius norm
	error $\frobnorm{\DelHat}$ then satisfies the bound
	\begin{align}
	\label{EqnDefnDelcritPartial}
	\frac{1}{2} \frobnorm{\DelHat}^2 & \leq \frac{1}{\pp}
	\tracer{\noise}{\DelHat} \; \leq \; Z \big( \frobnorm{\DelHat} \big).
	\end{align}
	Thus, in order to obtain a high probability bound, we need to
	understand the behavior of the random quantity $Z(t)$.
	
	One can verify that the set $\lseinterdiff$ is star-shaped, meaning
	that $\alpha \chattDiffmx \in \lseinterdiff$ for every $\alpha \in
	[0,1]$ and every $\chattDiffmx \in \lseinterdiff$. Using this
	star-shaped property, we are guaranteed that there is a non-empty set
	of scalars $\delcrit > 0$ satisfying the critical inequality
	\begin{align}
	\label{EqnCriticalPartial}
	\Exs[Z(\delcrit)] & \leq \frac{\delcrit^2}{2}.
	\end{align}
\end{subequations}
Our interest is in an upper bound to the smallest (strictly) positive
solution $\delcrit$ to the critical
inequality~\eqref{EqnCriticalPartial}, and moreover, our goal is to
show that for every $t \geq \delcrit$, we have $\frobnorm{\DelHat}
\leq c \sqrt{t \delcrit}$ with high probability.

Define a ``bad'' event
\begin{align}
\label{EqnDefnBadeventPartial}
\AuxEvent & \defn \Big \{ \exists \Delta \in \lseinterdiff \mid
\frobnorm{\Delta} \geq \sqrt{t \delcrit} \quad \mbox{and} \quad
\frac{1}{\pp} \tracer{\Delta}{\noise} \geq 2 \frobnorm{\Delta} \sqrt{t
	\delcrit} \Big \}.
\end{align}

Now suppose the event $\AuxEvent$ is true for some $t \geq \delcrit$, and let $\Delta_0 \in \lseinterdiff$ be a matrix that satisfies the two conditions required for $\AuxEvent$ to occur. Furthermore, since $\lseinterdiff$ is star-shaped, the function $Z(t)$ grows at most linearly in $t$. Consequently whenever event $\AuxEvent$ is true, we have $\frobnorm{\Delta_0} \geq \delcrit$ and hence
\begin{align*}
Z(\delcrit) \geq \frac{\delcrit}{\frobnorm{\Delta_0}} Z(\frobnorm{\Delta_0})  \stackrel{(i)}{ \geq }  \frac{\delcrit}{\frobnorm{\Delta_0}} \frac{1}{\pp}  \tracer{\Delta_0}{\Wmat} \stackrel{(ii)}{ \geq } 2 \delcrit
\sqrt{t \delcrit},
\end{align*}
where inequality $(i)$ follows from the definition of function $Z$ and inequality $(ii)$ uses the second condition in the definition of event $\AuxEvent$. As a consequence, we obtain the following bound on the probabilities of the associated events
\begin{align*}
\mprob[\AuxEvent] \leq \mprob[Z(\delcrit)\geq 2
\delcrit \sqrt{t \delcrit}] \qquad \mbox{for all $t \geq \delcrit$.}
\end{align*}
The following lemma helps control the behavior of the random variable
$Z(\delcrit)$.
\begin{lemma}
	\label{LemZpermPartial}
	For any $\delta >0$, the mean of $Z(\delta)$ is upper bounded as
	\begin{subequations}
		\begin{align}
		\label{EqnZpermPartialMean}
		\Exs [ Z(\delta) ] \leq \plaincon_1 \frac{\numwork + \numques}{\pp} \log^2
		(\numwork \numques),
		\end{align}
		and for every $u>0$, its tail probability is bounded as
		\begin{align}
		\label{EqnZTailBound}
		\mprob \Big(Z(\delta) > \Exs[Z(\delta)] + u \Big) \leq \exp\Big(
		\frac{-\plaincon_2 u^2 \pp}{\delta^2 + \Exs[Z(\delta)] + u} \Big),
		\end{align}
	\end{subequations}
	where $\plaincon_1$ and $\plaincon_2$ are positive universal constants.
\end{lemma}
\noindent See Section~\ref{SecProofLemZpermPartial} for the proof of
this lemma.

Setting $u = \delcrit \sqrt{t \delcrit}$ in the tail
bound~\eqref{EqnZTailBound}, we find that
\begin{align*}
\mprob \big(Z(\delcrit) > \Exs[Z(\delcrit)] \! + \! \delcrit \sqrt{t
	\delcrit} \big) \leq \exp\!\Big( \frac{-\plaincon_2 (\delcrit \sqrt{t
		\delcrit})^2 \pp }{\delcrit^2 + \Exs[Z(\delcrit)] + \delcrit
	\sqrt{t \delcrit}} \Big), ~ \mbox{for all $t\! > \! 0$.}
\end{align*}
By the definition of $\delcrit$ in~\eqref{EqnCriticalPartial}, we have
$\Exs[Z(\delcrit)] \leq \delcrit^2 \leq \delcrit \sqrt{t \delcrit}$
for any $t \geq \delcrit$, and with these relations we obtain the
bound
\begin{align*}
\mprob[\AuxEvent] & \leq \mprob[Z(\delcrit) \geq 2 \delcrit \sqrt{t
	\delcrit} \big] \; \leq \; \exp\big( -\frac{\plaincon_2}{3} \delcrit
\sqrt{t \delcrit} \pp \big), \quad \mbox{for all $t \geq \delcrit$.}
\end{align*}
Consequently, either $\frobnorm{\DelHat} \leq \sqrt{t \delcrit}$, or
we have $\frobnorm{\DelHat} > \sqrt{t \delcrit}$. In the latter case,
conditioning on the complement $\AuxEvent^c$, our basic inequality
implies that $\frac{1}{2} \frobnorm{\DelHat}^2 \leq 2
\frobnorm{\DelHat} \sqrt{t \delcrit}$ and hence $\frobnorm{\DelHat}
\leq 4 \sqrt{t \delcrit}$.  Putting together the pieces yields that
\begin{align}
\label{EqWithDelCritHPPartial}
\mprob \big( \frobnorm{\DelHat} \leq 4 \sqrt{t \delcrit} \big) \geq 1
- \exp\big( -\frac{\plaincon_2}{3} \delcrit \sqrt{t \delcrit} \pp \big),
\qquad \mbox{valid for all $t \geq \delcrit$.}
\end{align}

Finally, from the bound on the expected value of $Z(t)$ in
Lemma~\ref{LemZpermPartial}, we see that the critical
inequality~\eqref{EqnCriticalPartial} is satisfied for $\delcrit =
\sqrt{\frac{2\plaincon_1 (\numwork + \numques)}{\pp}}\log (\numwork
\numques)$. Setting $t = \delcrit = \sqrt{\frac{2\plaincon_1 (\numwork +
		\numques)}{\pp}}\log (\numwork \numques)$ in
equation~\eqref{EqWithDelCritHPPartial} yields the claimed result.


\subsubsection{Proof of Lemma~\ref{LemFrobToLoss}}
\label{SecProofLemFrobToLoss}

Consider any four scalars $a_1\geq 0, a_2 \geq 0, b_1 \in \{-1,1\}$
and $b_2 \in \{-1,1\}$. If $b_1 = b_2$ then
\begin{align*}
( a_1 b_1 - a_1 b_2 )^2 = 0 \leq ( a_1 b_1 - a_2 b_2 )^2.
\end{align*}
Otherwise we have $b_1 = - b_2$. In this case, since $a_1$ and $a_2$
have the same sign,
\begin{align*}
( a_1 b_1 - a_2 b_2 )^2 \geq ( a_1 b_1 )^2 = \frac{1}{4} (a_1 b_1 -
a_1 b_2)^2.
\end{align*}
The two results above in conjunction yield the inequality $( a_1 (b_1
- b_2) )^2 \leq 4 ( a_1 b_1 - a_2 b_2 )^2$.  Applying the above
argument to each entry of the matrices $A_1 \diag{v_1 - v_2}$ and
$(A_1 \diag{v_1} - A_2 \diag{v_2})$ yields the claim.


\subsubsection{Proof of Lemma~\ref{LemZpermPartial}}
\label{SecProofLemZpermPartial}

We need to prove the upper bound~\eqref{EqnZpermPartialMean} on the
mean, as well as the tail bound~\eqref{EqnZTailBound}.

\paragraph{Upper bounding the mean:} We 
upper bound the mean by using Dudley's entropy integral, as well as
some auxiliary results on metric entropy. Given a set $\genericclass$
equipped with a metric $\rho$ and a tolerance parameter $\epsilon \geq
0$, we let $\metent(\epsilon,\genericclass,\rho)$ denote the
$\epsilon$-metric entropy of the class $\genericclass$ in the metric
$\rho$.

With this notation, the truncated form of Dudley's entropy integral
inequality\footnote{Here we use $(\Delta \epsilon)$ to denote the
	differential of $\epsilon$, so as to avoid confusion with the number
	of questions $\numques$.} yields
\begin{align}
\label{EqnDudleyPartial} 
\Exs[ Z(\delta)] & \leq \frac{\plaincon}{\pp} \; \Big \{
\numques^{-8} + \int_{\frac{1}{2} \numques^{-9}}^{2 \sqrt{\numwork
		\numques}} \sqrt{ \metent(\epsilon,\lseinterdiff,\frobnorm{.})}
(\Delta \epsilon) \Big \}.
\end{align}
The upper limit of $2\sqrt{\numwork \numques}$ in the integration is
due to the fact $\frobnorm{\chattDiffmx} \leq 2 \sqrt{\numwork
	\numques}$ for every $\chattDiffmx \in \lseinterdiff$.

It is known~\cite{shah2015stochastically} that the metric entropy of
the set $\lseinterdiff$ is upper bounded as
\begin{align*}
\metent(\epsilon, \lseinterdiff, \frobnorm{.}) & \leq
8\frac{\max\{\numwork,\numques\}^2}{ \epsilon^2} \big(\log
\frac{\max\{\numwork,\numques\}}{ \epsilon} \big)^2 \qquad \mbox{for
	each $\epsilon > 0$.}
\end{align*}
Combining this upper bound with the Dudley entropy
integral~\eqref{EqnDudleyPartial}, and observing that the integration
has $\epsilon \geq \half \numques^{-9}$, the claimed upper
bound~\eqref{EqnZpermPartialMean} follows.


\paragraph{Bounding the tail probability of $Z(\delta)$:} 

In order to establish the claimed tail bound~\eqref{EqnZTailBound}, we
use a Bernstein-type bound on the supremum of empirical processes due
to Klein and Rio~\cite[Theorem 1.1c]{klein2005concentration}.  In
particular, this result applies to a random variable of the form
$X^\dagger = \sup_{v \in \mathcal{V} } \inprod{X}{v}$, where $X =
(X_1, \ldots, X_m)$ is a vector of independent random variables taking
values in $[-1, 1]$, and $\mathcal{V}$ is some subset of $[-1, 1]^m$.
Their theorem guarantees that for any $t>0$,
\begin{align}
\label{EqnKleinRio}
\mprob \big( X^\dagger > \Exs[ X^\dagger] + t \big) \leq \exp \Biggr(
\frac{ - t^2 }{2 \sup \limits_{v \in \mathcal{V} }
	\Exs[\inprod{v}{X}^2] + 4\Exs[ X^\dagger] + 3t} \Biggr).
\end{align}

In our setting, we apply this tail bound with the choices 
\begin{align*}
X =  \half \noise, \quad \mbox{and} \quad
X^\dagger = \half \sup_{ \substack{\chattDiffmx \in
		\lseinterdiff,\\ \frobnorm{\chattDiffmx} \leq \delta}} ~
\tracer{\chattDiffmx}{\noise} = \half \pp Z(\delta). 
\end{align*}
The entries of the matrix $ \noise$ are 
independently distributed with a mean of zero and a variance of at
most $4\pp$, and are bounded in absolute value by $2$. As a result, we have
$\Exs[\tracer{\chattDiffmx}{\noise}^2] \leq 4 \pp
\frobnorm{\chattDiffmx}^2 \leq 4 \pp \delta^2$ for every
$\chattDiffmx \in \lseinterdiff$. With these assignments, 
inequality~\eqref{EqnKleinRio} guarantees that
\begin{align*}
\mprob \big( \pp Z(\delta) > \pp \Exs[Z(\delta)] + \pp u \big) \leq \exp \Big( &
\frac{ - (u \pp)^2 }{2 \pp \delta^2 + 2 \pp \Exs[Z(\delta)] + 3 u \pp} \Big),
\end{align*}
for all $u > 0$, and some algebraic simplifications yield the claimed result.


\subsection{Proof of Theorem~\ref{ThmMinimax}(b): Minimax lower bound}
\label{SecProofThmMinimaxb}
We now turn to the proof of the minimax lower bound.  For a numerical
constant $\delta \in (0, \frac{1}{4})$ whose precise value is
determined later, define the vector $\probDS \in
[0,1]^{\numwork}$ with entries
\begin{align}
\label{EqnMinimaxLowerQstar}
\probDS_{i} = 
\begin{cases}
\half + \delta & \qquad \mbox{if $i \leq \frac{1}{\pp}$}\\
\half & \qquad \mbox{otherwise}.
\end{cases}
\end{align}
Set the probability matrix $\probmxstar \in [0,1]^{\numwork \times \numques}$ as $\probmxstar = \probDS \ones^T$. Observe that we then have $\probmxstar \in \classDS$. One may assume that the matrix $\probmxstar$ is known to the estimator under consideration. 

The Gilbert-Varshamov
bound~\cite{gilbert1952comparison,varshamov1957estimate} guarantees
that for a universal constant $\plaincon > 0$, there is a collection
$\packnum = \exp( \plaincon \numques)$ binary vectors---that is, a
collection of vectors $\{\ans^1,\ldots, \ans^\packnum\}$ all belonging
to the Boolean hypercube $\{-1,1\}^{\numques}$---such that the normalized Hamming
distance~\eqref{EqnDefnHamming} between any pair of vectors in this set is lower bounded as
\begin{align*}
\hamming ( \ans^\ell, \ans^{\ell'}) \geq \frac{1}{10} , \qquad
\mbox{for every $\ell, \ell' \in [\packnum]$.}
\end{align*}
For each $\ell \in [\packnum]$, let $\mprob^\ell$ denote the
probability distribution of $\obs$ induced by setting $\ansstar =
\ans^\ell$. For the choice of $\probmxstar$ specified
in~\eqref{EqnMinimaxLowerQstar}, following some algebra, we obtain a
upper bound on the Kullback-Leibler divergence between any pair of
distributions from this collection as
\begin{align*}
\kl{ \mprob^\ell }{ \mprob^{\ell'} } \leq \plaincon'  \numques  \delta^2
\qquad \mbox{for every $\ell \neq \ell' \in [\packnum]$},
\end{align*}
for another constant $\plaincon' > 0$.
Combining the above observations with Fano's
inequality~\cite{cover2012elements} yields that any estimator
$\anshat$ has 
expected normalized Hamming error lower bounded as
\begin{align*}
\Exs[ \hamming(\anshat, \ansstar ) ] \geq \frac{1}{20}  \Big(
1 - \frac{\plaincon' \numques \delta^2 + \log 2}{ \log \packnum }
\Big).
\end{align*}
Consequently, for the choice of $\probmxstar$ given by~\eqref{EqnMinimaxLowerQstar}, the $\probmxstar$-loss is lower bounded as
\begin{align*}
\Exs[\Qloss{\probmxstar}{\anshat}{\ansstar}] = \frac{4\delta^2}{\pp} \frac{
	\Exs[\hamming(\anshat, \ansstar)] }{\numwork} \geq \frac{4\delta^2}{20 \numwork \pp}
\Big( 1 - \frac{\plaincon' \numques
	\delta^2 + \log 2}{ \plaincon \numques } \Big) \stackrel{(i)}{\geq}
\frac{\plaincon''}{\numwork \pp},
\end{align*}
for some constant $\plaincon''>0$ as claimed. Here inequality (i)
follows by setting $\delta$ to be a sufficiently small positive
constant (depending on the values of $\plaincon'$ and $\plaincon''$).


\subsection{Proof of Corollary~\ref{CorEstimateQstar}(a)}
\label{SecProofCorEstimateQstara}

In the proof of Theorem~\ref{ThmMinimax}(a), we showed that there is a
constant $\plaincon_1 > 0$ such that
\begin{align*}
\frobnorm{(2 \probmxstar - \ones \ones^T) \ansstar - (2 \probmxlse -
	\ones \ones^T) \anslse}^2 \leq \plaincon_1 \frac{ \numques}{\pp} \log^2 \numques,
\end{align*}
with probability at least $1 - e^{- \numques \log (\numques
	\numwork)}$.  Since all entries of the matrices $2 \probmxstar -
\ones \ones^T$ and $2 \probmxlse - \ones \ones^T$ are non-negative,
and since every entry of the vectors $\ansstar$ and $\anslse$ lies in
$\{-1,1\}$, some algebra yields the bound
\begin{align*}
\big( (2 \probmxstar_{ij} - 1) - (2 [\probmxlse]_{ij} - 1) \big)^2
\leq \big( (2 \probmxstar_{ij} - 1)\ansstar_j - (2 [\probmxlse]_{ij} -
1) [\anslse]_j \big)^2,
\end{align*}
for every $i \in [\numwork],\; j
\in [\numques]$. Combining these inequalities yields the  claimed bound.


\subsection{Proof of Corollary~\ref{CorEstimateQstar}(b)}
\label{SecProofCorEstimateQstarb}

We begin by constructing a set, of cardinality $\packnum$, of possible
matrices $\probmxstar$, for some integer $\packnum > 1$, and
subsequently we show that it is hard to identify the true matrix if
drawn from this set. We begin by defining a $\packnum$-sized
collection of vectors $\{\hard^1,\ldots,\hard^\packnum\}$, all
contained in the set $[\half,1]^\numques$, as follows. The
Gilbert-Varshamov
bound~\cite{gilbert1952comparison,varshamov1957estimate} guarantees a
constant $\plaincon \in (0,1)$ such that there exists set of $\packnum
= \exp( \plaincon \numques)$ vectors, $\genericvec^1,\ldots,
\genericvec^\packnum \in \{-1,1\}^{\numques}$ with the property that the normalized Hamming distance~\eqref{EqnDefnHamming} between any pair of these vectors is lower bounded as
\begin{align*}
\hamming ( \genericvec^\ell, \genericvec^{\ell'}) \geq \frac{1}{10}, \qquad \mbox{for every $\ell, \ell' \in [\packnum]$}.
\end{align*}

Fixing some $\delta \in (0, \frac{1}{4})$, let us define, for each
$\ell \in [\packnum]$, the vector $\hard^\ell \in \real^\numques$ with
entries
\begin{align*}
[\hard^\ell]_j & \defn
\begin{cases}
\half + \delta & \quad \mbox{if } [\genericvec^\ell]_j = 1 \\ \half &
\quad \mbox{otherwise}.
\end{cases}
\end{align*}
For each $\ell \in [\packnum]$, define the matrix $\probmx^\ell =
\ones (\hard^\ell)^T$, and let $\mprob^\ell$ denote the probability
distribution of the observed data $\obs$ induced by setting
$\probmxstar = \probmx^\ell$ and $\ansstar = \ones$. Since the entries
of $\obs$ are all independent, some algebra leads to the following
upper bound on the Kullback-Leibler divergence between any pair of
distributions from this collection:
\begin{align*}
\kl{ \mprob^\ell }{ \mprob^{\ell'} } \leq  4 \pp \numwork \numques \delta^2 \qquad \mbox{for every $\ell \neq \ell' \in [\packnum]$}.
\end{align*}
Moreover, some simple calculation shows that the squared Frobenius
norm distance between any two matrices in this collection is lower
bounded as
\begin{align*}
\frobnorm{ \probmx^\ell - \probmx^{\ell'} }^2 \geq \frac{1}{10}
\numques \numwork \delta^2 \qquad \mbox{for every $\ell \neq \ell' \in
	[\packnum]$}.
\end{align*}
Combining the above observations with Fano's
inequality~\cite{cover2012elements} yields that any estimator
$\probmxhat$ for $\probmxstar$ has mean squared error lower bounded as
\begin{align*}
\Exs[ \frobnorm{\probmxstar - \probmxhat}^2 ] \geq \frac{1}{20}
\numques \numwork \delta^2 \Big( 1 - \frac{4 \pp \numques \numwork
	\delta^2 + \log 2}{ \log \packnum } \Big) \geq \plaincon'
\frac{\numques}{\pp},
\end{align*}
where we have set $\delta^2 = \frac{\plaincon''}{\pp \numwork}$ for a small enough positive constant $\plaincon''$, where
$\plaincon'$ is another positive constant whose value
may depend only on $\plaincon$ and $\plaincon''$.


\subsection{Proof of Theorem~\ref{ThmWANnew}: WAN under the permutation-based model}
\label{SecProofThmWANnew}

Observe that the windowing step of the WAN estimator identifies a
group of $\winsizeWAN$ workers such that their aggregate responses
towards questions are biased (towards either answer $\{-1,1\}$) by at
least $\sqrt{ \winsizeWAN \pp \log^{1.5}(\numques \numwork)}$. We
first derive three properties associated with having such a bias.
These properties involve function $\actualbias_\permwork:
[\numwork] \times [\numques] \times \{-1,1\} \rightarrow \reals$,
where $\actualbias_\permwork(\winsize,j, \ans)$ represents the amount
of bias in the responses of the top $\winsize \in [\numwork]$ workers
for question $j \in [\numques]$ towards the answer $\ans \in
\{-1,1\}$:
\begin{align*}
\actualbias_\permwork(\winsize,j, \ans) \defn \sum_{i = 1}^{\winsize}
( \indicator{ \obs_{ \permworkinv(i) j} = \ans } - \indicator{ \obs_{
		\permworkinv(i) j} = -\ans } ) = \ans \sum_{i=1}^{\winsize} \obs_{\permworkinv(i)j} .
\end{align*}
A straightforward application of the Bernstein
inequality~\cite{bernstein1924modification}, using the fact that the
entries of the observed matrix $\obs$ are all independent, with
moments bounded as
\begin{align*}
\Exs [ \obs_{ij} ] = 2 \pp (\probmxstar_{ij} - \half) \ansstar_j,
\quad \mbox{and} \quad \Exs [ \obs_{ij}^2 ] = \pp,
\end{align*}
ensures that all three properties stated below are satisfied with
probability at least $1 - e^{\plaincon \log^{1.5} (\numques \numwork)}$ for
every question $j \in [\numques]$ and every $\winsize \in \{\pp^{-1}
\log^{1.5}(\numques \numwork),\ldots,\numwork\}$. For the remainder of
the proof we work conditioned on the event where the following
properties hold:
\begin{enumerate}[label=(P{\arabic*})]
	\item Sufficient condition for bias towards correct answer: If
	$\sum_{i = 1}^{\winsize} ( \probmxstar_{ \permworkinv(i) j} - \half)
	\geq \frac{3}{4} \sqrt{ \frac{\winsize \log^{1.5}(\numques
			\numwork)}{ \pp} }$, then $\actualbias_\permwork(\winsize,j,
	\ansstar_j) \geq \sqrt{\winsize \pp \log^{1.5} (\numques
		\numwork)}$.
	\label{PropSufBias}
	\item Necessary condition for bias towards any answer $\ans \in
	\{-1,1\}$: $\actualbias_\permwork(\winsize,j, \ans) \geq
	\sqrt{\winsize \pp \log^{1.5} (\numques \numwork)}$ only if 
	$\ans = \ansstar_j$ and $\sum_{i = 1}^{\winsize} ( \probmxstar_{
		\permworkinv(i) j} - \half) \geq \frac{1}{4} \sqrt{ \frac{\winsize
			\log^{1.5}(\numques \numwork)}{\pp}}$.
	\label{PropNecBias}
	\item Sufficient condition for aggregate to be correct: If $\sum_{i =
		1}^{\winsize} ( \probmxstar_{ \permworkinv(i) j} - \half) \geq
	\frac{1}{4} \sqrt{ \frac{\winsize \log^{1.5}(\numques
			\numwork)}{\pp}}$, then $\actualbias_\permwork(\winsize,j,
	\ansstar_j) > 0$.
	\label{PropSufCorrect}
\end{enumerate}
%
%
%
We now show that when these three properties hold, for any question
$j_0 \in \setsignalques$, we must have that $[\ansWAN]_{j_0} =
\ansstar_{j_0}$. In particular, we do so by exihibiting a question
that is at least as hard as $j_0$ on which the WAN estimator is
definitely correct, and use the above properties to conclude that it
therefore must also be correct on the question $j_0$.

Recall that by the definition~\eqref{EqnDefnSetquesWAN} of
$\setsignalques$, for any question $j_0 \in \setsignalques$, it must be
the case that there exists a $\winsize_{j_0} \geq \pp^{-1} \log^{1.5}(\numques \numwork)$ such that
\begin{align}
\label{eqnCross}
\sum_{i=1}^{\winsize_{j_0}} (\probmxstar_{\permworkinv(i) j} - \half)
\geq\frac{3}{4} \sqrt{\frac{\winsize_{j_0}}{\pp} \log^{1.5} (\numques
	\numwork)}.
\end{align}
We define an associated set $\setsignalques_0$ as the set of questions
that are at least as easy as question $j_0$ according to the
underlying permutation $\permquesstar$, that is,
\begin{align*}
\setsignalques_0 \defn \{j \in [\numques] \mid \permquesstar(j) \leq
\permquesstar(j_0)\}.
\end{align*}
By the monotonicity of the columns of $\probmxstar$, every question in
$\setsignalques_0$ also satisfies condition~\eqref{eqnCross}.  For each
positive integer $\winsize$, define the set
\begin{align*}
\setquesnew{\winsize} & \defn \Big\{ j \in [\numques] \Big| \actualbias_\permwork(\winsize,j,\ans)  \geq \sqrt{\winsize
	\pp \log^{1.5} (\numques \numwork) } \mbox{~~~for some~~} \ans \in \{-1,1\} \Big\}.
\end{align*}
Property~\ref{PropSufBias} ensures that every question in the set
$\setsignalques_0$ is also in the set $\setquesnew{\winsize_{j_0}}$.  We
then have
\begin{align*}
|\setquesnew{\winsizeWAN}| \stackrel{(i)}{\geq}
|\setquesnew{\winsize_{j_0}}| \geq |\setsignalques_0|,
\end{align*}
where step (i) uses the optimality of $\winsizeWAN$ for the
optimization problem in equation~\eqref{EqnWindow}.  Given this, there
are two possibilities: either (1) we have the equality
$\setquesnew{\winsizeWAN} = \setsignalques_0$, or (2) the set
$\setquesnew{\winsizeWAN}$ contains some question not in the set
$\setsignalques_0$.  We address each of these possibilities in turn.

\noindent \underline{Case 1:} It suffices to observe by
Properties~\ref{PropSufBias}--\ref{PropSufCorrect}, that the
aggregate of the top $\winsizeWAN$ workers is correct on every
question in the set $\setquesnew{\winsizeWAN}$ and this implies that
it must be the case that $[\ansWAN]_{j_0} = \ansstar_{j_0}$ as
desired.

\noindent \underline{Case 2:} In this case, there is some question $j'
\notin \setsignalques_0$ such that $\actualbias_\permwork(\winsizeWAN,j,\ans) \geq \sqrt{\winsizeWAN \pp \log^{1.5}
	(\numques \numwork)}$ for some $\ans \in \{-1,1\}$.  Property~\ref{PropNecBias} guarantees that
$\sum_{i = 1}^{\winsizeWAN} ( \probmxstar_{ \permworkinv(i) j'} - \half)
\geq \frac{1}{4} \sqrt{ \frac{\winsizeWAN \log^{1.5}(\numques
		\numwork)}{\pp}}$ and that $\ans = \ansstar_j$.  Now, since every question easier than $j_0$ is
in the set $\setsignalques_0$, question $j'$ must be more difficult than
$j_0$, which implies that
\begin{align*}
\sum_{i = 1}^{\winsizeWAN} ( \probmxstar_{ \permworkinv(i) j_0} - \half)
\geq \frac{1}{4} \sqrt{ \frac{\winsizeWAN \log^{1.5}(\numques
		\numwork)}{\pp}}.
\end{align*}
Applying Property~\ref{PropSufCorrect}, we can then conclude that
$[\ansWAN]_{j_0} = \ansstar_{j_0}$ as desired.


\subsection{Proof of Corollary~\ref{CorWAN}}
\label{SecProofCorWAN}

Theorem~\ref{ThmWANnew} guarantees that the WAN estimator correctly answers
all questions that have some reasonable signal.  Note that the
set~\eqref{EqnDefnSetquesWAN} is defined in terms of the $\ell_1$-norm
of subvectors of columns of $\probmxstar - \half$, whereas the
conditions 
\begin{align}
\label{EqnEspresso}
\Lnorm{\probmxstar_j - \half}{2}^2 \geq \frac{5 \log^{2.5} (\numques
  \numwork)}{\pp} \quad \mbox{and} \quad \Lnorm{\probmx^\permwork_j -
  \probmx^{\permworkstar}_j}{2} \leq \frac{\Lnorm{\probmxstar_j -
    \half}{2}}{\sqrt{9 \log (\numques \numwork)}}
\end{align}
in the theorem claim are in terms of the $\ell_2$-norm of the columns
of $\probmxstar$. The following lemma allows us to connect the
$\ell_1$ and $\ell_2$-norm constraints for any vector in a general
class.

\begin{lemma}
\label{LemVecBig}
For any vector $\genericvec \in [0,1]^{\numwork}$ such that
$\genericvec_1 \geq \ldots \geq \genericvec_{\numwork}$, there must be
some $\numtopwork \geq \lceil \half \Lnorm{\genericvec}{2}^2 \rceil$ such that
\begin{align}
\label{EqnVecBig}
\sum_{i=1}^{\numtopwork} \genericvec_i \geq \sqrt{\frac{\numtopwork
    \Lnorm{\genericvec}{2}^2}{ 2 \log \numwork}}.
\end{align}
\end{lemma}
\noindent See Section~\ref{SecProofLemVecBig} for the proof of this
lemma.

We now complete the proof of the theorem.
We may assume without loss of generality that the rows of
$\probmxstar$ are ordered to be non-decreasing downwards along any
column, that is, that $\permworkstar$ is the identity permutation. Consider any question $j \in [\numques]$ for which the
permutation $\permwork$ satisfies the bounds~\eqref{EqnEspresso}.  For
any $\ell \in [\numwork]$, let $\onesvec{\ell} \in \reals^\numwork$
denote a vector with ones in its first $\ell$ positions and zeros
elsewhere. The Cauchy-Schwarz inequality implies that $(\probmx^\perm_j - \half)^T \onesvec{\ell} \geq (\probmxstar_j-
  \half)^T \onesvec{\ell} - \sqrt{\ell} \Lnorm{\probmx^\perm_j -
    \probmxstar_j}{2}$. 
By applying Lemma~\ref{LemVecBig} to the vector $\probmxstar_j -
\half$, we are guaranteed the existence of some value $\winsize \geq
\frac{5 \log^{2.5} (\numques \numwork)}{2\pp}$ such that
$(\probmxstar_j- \half)^T \onesvec{\winsize} \geq
\Lnorm{\probmxstar_j- \half}{2} \sqrt{\frac{\winsize}{2 \log
    \numwork}}$.  Consequently, we have the lower bound
\begin{align*}
(\probmx^\perm_j - \half)^T \onesvec{\winsize}
& \; \geq \; \Lnorm{\probmxstar_j- \half}{2} \sqrt{\frac{\winsize}{2
    \log \numwork}} - \sqrt{\winsize} \Lnorm{\probmx^\perm_j -
  \probmxstar_j}{2} \\
& \; \stackrel{(i)}{\geq} \; .37 \Lnorm{\probmxstar_j- \half}{2}
\sqrt{\frac{\winsize}{\log (\numques \numwork)}} 
 \; \stackrel{(ii)}{\geq} \; \frac{3}{4} \sqrt{\frac{\winsize}{\pp}
  \log^{1.5} (\numques \numwork)} ,
\end{align*}
where inequalities (i) and (ii) follow from
conditions~\eqref{EqnEspresso}.  Consequently, we can apply
Theorem~\ref{ThmWANnew} for every such question $j$, thereby yielding the
 result~\eqref{EqnWanCorWorks}.
 
 Finally, the claimed result~\eqref{EqnCalibrated} on the $\probmxstar$-loss under the correct permutation is obtained by considering a zero error (with high probability) for all questions $j \in [\numques]$ for which $\Lnorm{\probmxstar_j - \half}{2}^2 \geq \frac{5 \log^{2.5} (\numques
 	\numwork)}{\pp} $ and where each of the remaining (at most $\numques$) questions contribute a $\probmxstar$-loss of at most $\frac{5 \log^{2.5} (\numques
 	\numwork)}{\numwork \numques \pp} $.


\subsubsection{Proof of Lemma~\ref{LemVecBig}}
\label{SecProofLemVecBig}

We partition the proof into two cases depending on the value of $\Lnorm{\genericvec}{2}^2$.

\noindent {\bf Case 1:} First, suppose that $\half
\Lnorm{\genericvec}{2}^2 \geq e$.  In this case, we proceed via proof
by contradiction. If the claim were false, then we would have
\begin{align*}
\sqrt{\frac{\numtopwork \Lnorm{\genericvec}{2}^2}{ 2\log \numwork}} >
\sum_{i=1}^{\numtopwork} \genericvec_i \geq \numtopwork
\genericvec_\numtopwork \qquad \mbox{for every $\numtopwork \geq
	\lceil \half \Lnorm{\genericvec}{2}^2 \rceil$.}
\end{align*}
It would then follow that
\begin{align*}
\sum_{i=1}^{\numwork} \genericvec_i^2 \; = \; \sum_{i=1}^{\lceil \half
	\Lnorm{\genericvec}{2}^2 \rceil - 1} v_i^2 + \sum_{i=\lceil \half
	\Lnorm{\genericvec}{2}^2 \rceil}^n v_i^2 \;
& \stackrel{\mathrm{(i)}}{\leq} \; \lceil \half \Lnorm{\genericvec}{2}^2
\rceil - 1 + \sum_{i=\lceil \half \Lnorm{\genericvec}{2}^2 \rceil}^n
v_i^2 \\
& < \; \half \Lnorm{\genericvec}{2}^2 \; \, + \sum_{i=\lceil
	\half \Lnorm{\genericvec}{2}^2 \rceil }^{\numwork} \frac{
	\Lnorm{\genericvec}{2}^2}{2 i \log \numwork},
\end{align*}
where step (i) uses the fact that $v_i \in [0,1]$. Using the standard
bound $\sum_{i=a}^{b} \frac{1}{i} \leq \log(\frac{e b}{a})$ and the
assumption $\lceil \half \Lnorm{\genericvec}{2}^2 \rceil \geq e$, we
find that
\begin{align*}
\half \Lnorm{\genericvec}{2}^2 \; \, + \sum_{i=\lceil \half
	\Lnorm{\genericvec}{2}^2 \rceil }^{\numwork} \frac{
	\Lnorm{\genericvec}{2}^2}{2 i \log \numwork} \leq
\Lnorm{\genericvec}{2}^2.
\end{align*}
The resulting chain of inequalities contradicts the definition of
$\Lnorm{\genericvec}{2}^2$.\\

\noindent {\bf Case 2:} Otherwise, we may assume that $\half
\Lnorm{\genericvec}{2}^2 < e$. Observe that the case
$\genericvec = 0$ trivially satisfies the claim with
$\numtopwork=1$, and hence we restrict attention to non-zero
vectors. Define a vector $\genericvec' \in [0,1]^{\numwork}$ as
\begin{align*}
\genericvec' = \frac{1}{\genericvec_1} \genericvec.
\end{align*}
We first prove the claim of the lemma for the vector $\genericvec'$,
that is, we prove that there exists some value $\numtopwork \geq
\lceil \half \Lnorm{\genericvec'}{2}^2 \rceil$ such that
\begin{align}
\sum_{i=1}^{\numtopwork} \genericvec'_i \geq \sqrt{\frac{\numtopwork
		\Lnorm{\genericvec'}{2}^2}{ 2\log \numwork}}.
\label{EqnBigVecPrime}
\end{align}
Observe that $1 = \genericvec'_1 \geq \cdots \geq
\genericvec'_\numwork \geq 0$. If $\half \Lnorm{\genericvec'}{2}^2
\geq e$, then our claim~\eqref{EqnBigVecPrime} is proved via the
analysis of Case 1 above. Otherwise, we have that $\half
\Lnorm{\genericvec'}{2}^2 \leq e$ and $\genericvec'_1 = 1$. Setting
$\numtopwork = 1$, we obtain the inequalities
\begin{align*}
\sum_{i=1}^{\numtopwork} \genericvec'_i = 1 \mbox{\quad and \quad} \sqrt{\frac{\numtopwork \Lnorm{\genericvec'}{2}^2}{ 2\log \numwork}} \leq 1,
\end{align*}
where we have used the assumption that $\numwork$ is large enough
(concretely, $\numwork \geq 16$). We have thus proved the
bound~\eqref{EqnBigVecPrime}, and it remains to translate this bound
on $\genericvec'$ to an analogous bound on the vector $\genericvec$.
Observe that since $\genericvec_1 \leq 1$, we have the relation
$\Lnorm{\genericvec'}{2} \geq \Lnorm{\genericvec}{2}$. Using the same
value of $\numtopwork$ as that derived for vector $\genericvec'$, we
then obtain from~\eqref{EqnBigVecPrime} that this value $\numtopwork
\geq \lceil \half \Lnorm{\genericvec'}{2}^2 \rceil \geq \lceil \half
\Lnorm{\genericvec}{2}^2 \rceil$ satisfies
\begin{align*}
\genericvec_1 \sum_{i=1}^{\numtopwork} \genericvec'_i \geq
\genericvec_1 \sqrt{\frac{\numtopwork \Lnorm{\genericvec'}{2}^2}{
		2\log \numwork}},
\end{align*}
which establishes the claim.


\subsection{Proof of Theorem~\ref{ThmOBIWAN}: OBI-WAN under the intermediate model}
\label{SecProofThmOBIWANa}
To simplify notation, let us define the vector $\probshift \defn \probInt - \half$.  
Note that the value of the constant $\capconst$ in the statement of the theorem is
specified later in the proof via 
equation~\eqref{EqnApproxSValmostDone} in Lemma~\ref{LemOBIPerturb}.

Our proof of this case is divided into
three parts, each corresponding to one of the three steps in the
\obiwan algorithm. The first step is to derive certain properties of
the split of the questions. The second step is to derive
approximation-guarantees on the outcome of the OBI step. The third and
final step is to show that this approximation guarantee ensures that
the output of the WAN estimator meets the claimed error guarantee.

\paragraph{Step 1:  Analyzing the split}
Our first step is to exhibit a useful property of the split of the
questions---namely, that with high probability, the questions in the
two sets $\setques_0$ and $\setques_1$ have a similar total
difficulty.

The random sets $(\setques_0, \setques_1)$ chosen in the first step can be
obtained as follows: first generate an i.i.d. sequence
$\{\epsilon_j\}_{j=1}^\numques$ of equiprobable $\{0,1\}$ variables,
and then set $\setques_\ell \defn \{ j \in [\numques] \, \mid \epsilon_j =
\ell \}$ for $\ell \in \{0,1\}$.  Note that we have $\Exs[\sum_{j \in [\numques]} (1 - \hardstar_j)^2 \epsilon_j ] =
\frac{1}{2} \Lnorm{1 - \hardstar}{2}^2$, and $\Exs[\sum_{j \in [\numques]} ((1 - \hardstar_j)^2 \epsilon_j)^2 ] =
\frac{1}{2} \sum_{j \in [\numques]} (1 - \hardstar_j)^4 \leq
\frac{1}{2} \Lnorm{1 - \hardstar}{2}^2$. 
Applying Bernstein's inequality then guarantees that
\begin{align*}
\mprob \big( \sum_{j \in \setques_\ell} (1 - \hardstar_j)^2 >
\frac{2}{3} \Lnorm{1 - \hardstar}{2}^2 \big) \leq \exp \big( - \plaincon
\Lnorm{1 - \hardstar}{2}^2 \big) \qquad \mbox{for each $\ell \in
  \{0,1\}$},
\end{align*}
where $\plaincon$ is a positive universal constant.  We are thus
guaranteed that
\begin{align}
\label{EqnHardnessSplit}
\frac{1}{3} \Lnorm{1 - \hardstar}{2}^2 \leq \sum_{j \in \setques_\ell}
(1 - \hardstar_j)^2 \leq \frac{2}{3} \Lnorm{1 - \hardstar}{2}^2 \qquad
\mbox{for both $\ell \in \{1,2\}$},
\end{align}
with probability at least $1- e^{-\plaincon \capconst \frac{\log^{2.5}
    \numques}{\pp}}$, where we have used the fact that $\Lnorm{1 -
  \hardstar}{2}^2 \geq \frac{\capconst \numques \log^{2.5}
  \numques}{\pp \Lnorm{\probshift}{2}^2} \geq \frac{\capconst
  \log^{2.5} \numques}{\pp}$.  Now define the error event
\begin{align*}
\errevent & \defn \Big \{
\Qloss{\probmxstar}{\ansOBIWAN}{\ansstar} > \frac{6 \capconst
   \log^{2.5} \numques}{\numwork \pp} \Big \}.
\end{align*}
Combining the sandwich relation~\eqref{EqnHardnessSplit} with the
union bound, we find that
\begin{align*}
\mprob(\errevent) & = \!\sum \limits_{\mbox{\small partitions }
  \setquesval_0,\setquesval_1} \mprob(\errevent \mid \setques_0 =
\setquesval_0,\setques_1 = \setquesval_1) \mprob(\setques_0 =
\setquesval_0,\setques_1 = \setquesval_1) \\
& \leq\! \sum \limits_{\substack{\mbox{\small partitions
    }\setquesval_0,\setquesval_1\\ \mbox{\small
      satisfying}~\eqref{EqnHardnessSplit}}} \! \mprob(\errevent \mid
\setques_0 = \setquesval_0,\setques_1 = \setquesval_1)
\mprob(\setques_0 = \setquesval_0,\setques_1 = \setquesval_1) +
e^{-\plaincon \capconst \frac{\log^{2.5} \numques}{\pp}}.
\end{align*}
Consequently, in the rest of the proof we consider any partition
$(\setquesval_0,\setquesval_1)$ that satisfies the sandwich
bound~\eqref{EqnHardnessSplit} and derive an upper bound on the error
conditioned on this partition. In other words, it suffices to prove
the following bound for any partition $(\setquesval_0,\setquesval_1)$
satisfying~\eqref{EqnHardnessSplit}:
\begin{align}
\label{EqnOBIWANToProve}
\mprob(\errevent \mid \setques_0 = \setquesval_0,\setques_1 =
\setquesval_1) \leq e^{- \plaincon' \log^{1.5} (\numques \numwork)},
\end{align}
for some positive universal constant $\plaincon'$
whose value may depend only on $\capconst$.  We note that conditioned
on the partition $(\setquesval_0,\setquesval_1)$, and for any fixed
values of $\probmxstar$ and $\ansstar$, the responses of the workers
to the questions in one set are statistically independent of the
responses in the other set. Consequently, we describe the proof for
any one of the two partitions, and the overall result is implied by a
union bound of the error guarantees for the two partitions. We use the
notation $\ell$ to denote either one of the two partitions in the
sequel, that is, $\ell \in \{0,1\}$.


\paragraph{Step 2: Guarantees for the OBI step} 

Assume without loss of generality that the rows of the matrix
$\probmxstar$ are ordered according to the abilities of the
corresponding workers, that is, the entries of $\probInt$ are arranged
in a non-increasing order. Recall that $\permwork_\ell$ denotes the
permutation of the workers in order of their respective values in
$\eigvecobs_\ell$. Let $\probshiftapprox_\ell \in \reals^\numwork$
denote the vector obtained by permuting the entries of $\probshift$ in
the order given by $\permwork_\ell$. Thus the entries of
$\probshiftapprox_\ell$ are identical to those of $\probshift$ up to a
permutation; the ordering of the entries of $\probshiftapprox_\ell$ is
identical to the ordering of the entries of $\eigvecobs_\ell$. The
following lemma---central for the proof of this theorem---establishes a relation between these
vectors. The proof of this lemma combines matrix perturbation theory with some
careful algebraic arguments.

\begin{lemma}
\label{LemOBIPerturb}
Suppose that condition~\eqref{EqnNiharCondition} holds for a
sufficiently large constant $\capconst > 0$.  Then for any split
$(\setques_0,\setques_1)$ satisfying the
relation~\eqref{EqnHardnessSplit}, we have
\begin{align}
\label{EqnOBIPerturb}
\mprob \Big( \Lnorm{\probshiftapprox_\ell - \probshift }{2}^2 >
  \frac{ \Lnorm{\probshift}{2}^2}{ 9 \log (\numques \numwork)} \Big)
& \leq e^{-\plaincon \log^{1.5} \numques}.
\end{align}
\end{lemma}
\noindent See Section~\ref{SecProofLemOBIPerturb} for the proof of
this lemma. \\

\noindent At this point, we are now ready to apply the bound for the
WAN estimator from Corollary~\ref{CorWAN}.


\paragraph{Step 3: Guarantees for the WAN step} 

Recall that for any choice of index $\ell \in \{0,1\}$, the OBI step
operates on the set $\setques_\ell$ of questions, and the WAN step
operates on the alternate set $\setques_{1-\ell}$. Consequently,
conditioned on the partition $(\setquesval_0,\setquesval_1)$, the
outcomes $\obs_{1 - \ell}$ of the comparisons in set $(1 - \ell)$ are
statistically independent of the permutation $\permwork_\ell$ obtained
from set $\ell$ in the OBI step.  

Consider any question $j \in \setques_{1 - \ell}$ that satisfies the
inequality $\Lnorm{(1 - \hardstar_j) \probshift}{2}^2 \geq \frac{5
  \log^{2.5} (\numques \numwork)}{\pp}$. We now claim that this
question $j$ satisfies the pair of conditions~\eqref{EqnWANConditions}
required by the statement of Corollary~\ref{CorWAN}. First
observe that $(1 - \hardstar_j) \probshift$ is simply the $j^{th}$
column of the matrix $(\probmxstar - \half)$, we have
$\Lnorm{\probmxstar_j - \half}{2}^2 \geq \frac{5 \log^{2.5} (\numques
  \numwork)}{\pp}$. The first condition in~\eqref{EqnWANConditions} is
thus satisfied.

In order to establish the second condition, observe that a rescaling
of the inequality~\eqref{EqnOBIPerturb} by the non-negative scalar $(1
- \hardstar_j)$ yields the bound
\begin{align}
\label{EqnOneMinusEll}
\Lnorm{(1 - \hardstar_j) \probshiftapprox_\ell - (1 -
  \hardstar_j)\probshift }{2}^2 \leq \frac{ \Lnorm{(1 - \hardstar_j)
    \probshift}{2}^2}{9 \log ( \numques \numwork) } \quad \mbox{ for
  every $j \in \setques_{1 - \ell}$}.
\end{align}
Recall our notational assumption that the entries of $\probInt$ (and
hence the rows of $\probmxstar$) are arranged in order of the workers'
abilities, and that $\probmx^\permwork$ is a matrix obtained by
permuting the rows of $\probmxstar$ according to a given permutation
$\permwork$. Also observe that the vector $(1 - \hardstar_j)
\probshiftapprox_\ell$ equals the $j^{th}$ column of
$(\probmx^{\permwork_\ell} - \half)$, where $\permwork_\ell$ is the
permutation of the workers obtained from the OBI step. Consequently,
the approximation guarantee~\eqref{EqnOneMinusEll} implies that
$\Lnorm{\probmx^{\permwork_\ell}_j - \probmxstar_j}{2} \leq
\frac{\Lnorm{\probmxstar_j}{2}}{\sqrt{9 \log (\numques
    \numwork)}}$. Thus the second condition in
equation~\eqref{EqnWANConditions} is also satisfied for the question
$j$ under consideration.

This allows us to apply the result of Corollary~\ref{CorWAN} for the WAN step, which yields that this question $j$ is decoded correctly with a probability
at least $1 - e^{- \plaincon \log^{1.5} (\numques \numwork)}$, for
some positive constant $\plaincon$. This argument holds for
every question $j$ satisfying $\Lnorm{(1 - \hardstar_j)
  \probshift}{2}^2 \geq \frac{5 \log^{2.5} (\numques \numwork)}{\pp}$, and applying the union bound shows that all these questions are decoded correctly with high probability.


\subsubsection{Proof of Lemma~\ref{LemOBIPerturb}}
\label{SecProofLemOBIPerturb}

\noindent The proof of this lemma consists of three main steps:
\begin{enumerate}[leftmargin=*]
	\item[(i)] First, we show that $\eigvecobs_\ell$ is a good
	approximation for the vector of worker abilities $\probshift$ up to
	a global sign.
	\item[(ii)]  We then show that the global sign is correctly identified
	with high probability.
	\item[(iii)] The final step in the proof is to convert this guarantee
	to one on the permutation induced by $\eigvecobs_\ell$.
\end{enumerate}


\paragraph{Step 1}

We first show that the vector $\eigvecobs_\ell$ approximates
$\probshift$ up to a global sign.  Assume without loss of generality
that $\ansstar_j = 1$ for every question $j \in [\numques]$. As in the
proof of Theorem~\ref{ThmMinimax}(a), we begin by rewriting the model
in a ``linearized'' fashion which is convenient for our analysis. Let
$\probmxstar_0$ and $\probmxstar_1$ denote the submatrices of
$\probmxstar$ obtained by splitting its columns according to the sets
$\setques_0$ and $\setques_1$. Then we have for $\ell \in \{0,1\}$,
\begin{align}
\label{EqnLinearized3}
\frac{1}{\pp} \obs_\ell = (2\probmxstar_\ell - \ones \ones^T) \;
\diag{\ansstar} + \frac{1}{\pp} \noise_\ell,
\end{align}
where conditioned on $\setques_0$ and $\setques_1$, the noise matrices
$\noise_0,\noise_1 \in \reals^{\numwork \times \numques}$ have entries
independently drawn from the distribution~\eqref{EqnDefnNoise}. One
can verify that the entries of $\noise_0$ and $\noise_1$ have a mean
of zero, second moment upper bounded by $4\pp$, and their absolute
values are upper bounded by $2$.

We now require a standard result on the perturbation of eigenvectors
of symmetric matrices~\cite{stewart1990matrix}.  Consider a symmetric
and positive semidefinite matrix $\mxorig \in \reals^{\numques \times
	\numques}$, a second symmetric matrix $\mxdelta \in \reals^{\numques
	\times \numques}$, and let $\mxnew = \mxorig + \mxdelta$.  Let
$\vectoprightSV \in \reals^{\numques}$ be an eigenvector associated to
the largest eigenvalue of $\mxorig$. Likewise define
$\vecnewtoprightSV \in \reals^{\numques}$ as an eigenvector associated
to the largest eigenvalue of $\mxnew$. Then we are
guaranteed~\cite{stewart1990matrix} that
\begin{align}
\label{EqnStewart}
\min\{ \Lnorm{\vecnewtoprightSV - \vectoprightSV}{2},
\Lnorm{\vecnewtoprightSV + \vectoprightSV}{2} \} \leq \frac{
	2\opnorm{\mxdelta} }{\max\{ \eigenvalue{1}{\mxorig} -
	\eigenvalue{2}{\mxorig} - 2\opnorm{\mxdelta} , 0\} },
\end{align}
where $\eigenvalue{1}{\mxorig}$ and $\eigenvalue{2}{\mxorig}$ denote
the largest and second largest eigenvalues of $\mxorig$, respectively.

In order to apply the bound~\eqref{EqnStewart}, we define the matrix
$\probmxshift_\ell \defn \probmxstar_\ell - \half \ones \ones^T$, as
well as the matrices
\begin{align*} \mxnew \defn 
\frac{1}{\pp^2} \obs_\ell \obs_\ell^T, \quad \mxorig =
4 \probmxshift_\ell (\probmxshift_\ell)^T, \quad \mbox{and} \quad \\
\mxdelta \defn \frac{2}{\pp} \noise_\ell (\probmxshift_\ell)^T +
\frac{2}{\pp} \probmxshift_\ell \noise_\ell^T + \frac{1}{\pp^2}
\noise_\ell \noise_\ell^T.
\end{align*}
Using our linearized observation model~\eqref{EqnLinearized3}, it is
straightforward to verify that these choices satisfy the condition
$\mxnew = \mxorig + \mxdelta$, so that the bound~\eqref{EqnStewart}
can be applied.

Recall that for any matrix $\probmxstar \in \classInter$, we have
$\probmxstar = \probInt (1 - \hardstar)^T + \half (\hardstar)^T$ for some vectors $\probInt
\in [\half,1]^{\numwork}$ and $ \hardstar \in [0,1]^\numques$. Also
recall our definition of the associated quantity $\probshift \in
[0,\half]^{\numwork}$ as $\probshift = \probInt - \half$. We denote
the magnitude of the vector $\probshift$ as $\probshiftmag \defn
\Lnorm{\probshift}{2}$.

With the notation introduced above, we are ready to apply the
bound~\eqref{EqnStewart}.  First observe that the matrix
$\probmxshift_\ell$ has a rank of one, and consequently $\opnorm{
	\probmxshift_\ell } = \probshiftmag \sqrt{\sum_{j \in \setques_\ell}
	(1 - \hardstar_j)^2}$. Conditioned on the
bound~\eqref{EqnHardnessSplit}, we obtain 
\begin{align*}
\sqrt{\frac{1}{3}} \probshiftmag \Lnorm{1 - \hardstar}{2} \leq \opnorm{
	\probmxshift_\ell } \leq \sqrt{\frac{2}{3}} \probshiftmag
\Lnorm{1 - \hardstar}{2}.
\end{align*} 
Moreover, the entries of the matrix $\noise_\ell$ are independent,
zero-mean, and have a second moment upper bounded by $4
\pp$. Consequently, known results on random matrices~\cite[Remark
3.13]{bandeira2014sharp} guarantee that
\begin{align*}
\opnorm{ \noise_\ell } \leq \plaincon \sqrt{\max\{\numques,\numwork\}
	\pp \log^{1.5} \numques} \leq \plaincon \sqrt{\numques \pp
	\log^{1.5} \numques},
\end{align*}
with probability at least $1 - e^{- \plaincon \log^{1.5} \numques}$, where
we have used the fact that $\numques \geq \numwork$ and $\pp \geq
\frac{1}{\numwork}$.  These inequalities, in turn, imply that the top
eigenvalue of $\mxorig$ is lower bounded as $\eigenvalue{1}{\mxorig} =
\opnorm{\probmxshift}^2 \geq \frac{1}{3} \probshiftmag^2 \Lnorm{1 -
	\hardstar}{2}^2$, the second eigenvalue vanishes (that is,
$\eigenvalue{2}{\mxorig} = 0$), and moreover that
\begin{align*}
\opnorm{\mxdelta} & \leq \frac{2}{\pp} \opnorm{\probmxshift}
\opnorm{\noise} + \frac{1}{\pp^2} \opnorm{\noise}^2 \\
& \leq \plaincon'
\frac{\sqrt{\numques \log^{1.5} \numques} }{\pp} ( \probshiftmag
\Lnorm{1 - \hardstar}{2} \sqrt{\pp} + \sqrt{\numques \log^{1.5} \numques}
).
\end{align*}
Recall the lower bound $\probshiftmag \Lnorm{1 - \hardstar}{2} \geq
\sqrt{ \frac{\capconst \numques \log^{2.5} \numques}{\pp} }$, assumed in the statement of
the lemma.  Using these facts and doing some algebra, we find that
with probability at least $1 - e^{- \plaincon \log^{1.5} \numques}$, for
any pair of sets $\setques_0$ and $\setques_1$
satisfying~\eqref{EqnHardnessSplit}, we have the bound
\begin{align}
\label{EqnApproxSValmostDone}
\min\{ \Lnorm{\eigvecobs_\ell - \frac{1}{\probshiftmag} \probshift
}{2}^2, \Lnorm{\eigvecobs_\ell + \frac{1}{\probshiftmag} \probshift
}{2}^2 \} \leq \frac{1}{36} \frac{1}{\probshiftmag^2 \Lnorm{
		1 - \hardstar_j}{2}^2} \frac{\numques \log^{1.5} \numques}{\pp},
\end{align}
where the prefactor $\frac{1}{36}$ is obtained by setting the constant $\capconst > 20$ to a large enough value.


\paragraph{Step 2}
We now verify that the global sign is correctly identified.
Recall our selection
\begin{align*}
\sum_{j=1}^{\numwork} [\eigvecobs_\ell]_j^2
\indicator{[\eigvecobs_\ell]_j > 0} \geq \sum_{j=1}^{\numwork}
[\eigvecobs_\ell]_j^2 \indicator{[\eigvecobs_\ell]_j < 0}.
\end{align*}
Since every entry of the vector $\probshift$ is non-negative, we have
the inequality
\begin{align*}
\Lnorm{\eigvecobs_\ell + \frac{1}{\rho}\probshift }{2}^2 \geq
\sum_{j=1}^{\numwork} [\eigvecobs_\ell]_j^2
\indicator{[\eigvecobs_\ell]_j > 0} \geq \sum_{j=1}^{\numwork}
[\eigvecobs_\ell]_j^2 \indicator{[\eigvecobs_\ell]_j < 0},
\end{align*}
and consequently,
\begin{subequations}
	\begin{align}
	\label{EqnEVSign1}
	\Lnorm{\eigvecobs_\ell + \frac{1}{\rho} \probshift }{2}^2 \geq \half
	\Lnorm{\eigvecobs_\ell}{2}^2.
	\end{align}
	On the other hand, a version of the triangle inequality yields
	\begin{align}
	\label{EqnEVSign2}
	2\Lnorm{ \eigvecobs_\ell}{2}^2 + 2 \Lnorm{ \eigvecobs_\ell +
		\frac{1}{\rho} \probshift }{2}^2 \geq \Lnorm{ \frac{1}{\rho}
		\probshift }{2}^2 = 1
	\end{align}
	Now suppose that $\Lnorm{\eigvecobs_\ell - \frac{1}{\rho}\probshift
	}{2}^2 \geq \Lnorm{\eigvecobs_\ell + \frac{1}{\rho}\probshift
	}{2}^2$. Then from our earlier result~\eqref{EqnApproxSValmostDone},
	we have the bound
	\begin{align}
	\label{EqnEVSign3}
	\Lnorm{ \eigvecobs_\ell + \frac{1}{\rho} \probshift}{2}^2 \leq
	\frac{\numques \log^{1.5} \numques}{36 \rho^2
		\Lnorm{1 - \hardstar}{2}^2 \pp},
	\end{align}
\end{subequations}
with probability at least $1 - e^{-\plaincon \log^{1.5} (\numques
	\numwork)}$. Putting together the
inequalities~\eqref{EqnEVSign1},~\eqref{EqnEVSign2}
and~\eqref{EqnEVSign3} and rearranging some terms yields the
inequality
\begin{align*}
\probshiftmag^2 \Lnorm{1 - \hardstar}{2}^2  \leq 
\frac{\numques \log^{1.5} \numques}{9 \pp}.
\end{align*}
This requirement contradicts our initial assumption $\probshiftmag^2
\Lnorm{1 - \hardstar}{2}^2 \geq \frac{\capconst \numques \log^{2.5}
	\numques}{\pp}$, with $\capconst > 20$, 
thereby proving that $\Lnorm{\eigvecobs_\ell -
	\frac{1}{\rho}\probshift }{2}^2 < \Lnorm{\eigvecobs_\ell +
	\frac{1}{\rho}\probshift }{2}^2$. Substituting this inequality into
equation~\eqref{EqnApproxSValmostDone} yields the bound
\begin{align}
\label{EqnApproxSValmostDone2}
\Lnorm{\eigvecobs_\ell - \frac{1}{\probshiftmag} \probshift }{2}^2
\leq  \frac{1}{36 \probshiftmag^2 \Lnorm{ 1 - \hardstar_j}{2}^2}
\frac{\numques \log^{1.5} \numques}{\pp}.
\end{align}


\paragraph{Step 3}

The final step of this proof is to convert the approximation
guarantee~\eqref{EqnApproxSValmostDone2} on $\eigvecobs_\ell$ to an
approximation guarantee on the vector $\probshiftapprox_\ell$ (which,
recall, is a permutation of $\probshift$ according to the permutation
induced by $\eigvecobs_\ell$). An additional lemma is useful for this
step:
\begin{lemma}
	\label{LemPermVec}
	For any $\ell \in \{0,1\}$, we have $\Lnorm{ \probshiftapprox_\ell -
		\probshift}{2} \leq 2 \Lnorm{ \probshiftmag \eigvecobs_\ell -
		\probshift }{2}$.
\end{lemma}
\noindent See Section~\ref{SecProofLemPermVec} for the proof of this
claim. \\

Combining Lemma~\ref{LemPermVec} with the
inequality~\eqref{EqnApproxSValmostDone2} yields that for any choice
of the set $\setques_0$ and $\setques_1$ satisfying the
condition~\eqref{EqnHardnessSplit}, with probability at least $1 -
e^{- \plaincon \log^{1.5} \numques}$, we have
\begin{align*}
\Lnorm{\probshiftapprox_\ell - \probshift }{2}^2 \; \leq \; 
\frac{1}{18 \Lnorm{1 - \hardstar}{2}^2} \frac{\numques
	\log^{1.5} \numques}{\pp} \; \stackrel{(i)}{\leq} \;  \frac{ \Lnorm{\probshift}{2}^2}{18 \log (\numques \numwork)}.
\end{align*}
Here, inequality (i) follows from our earlier assumption that
$\Lnorm{\probshift}{2} \Lnorm{1 - \hardstar}{2} \geq \sqrt{
	\frac{\capconst \numques \log^{2.5} \numques}{\pp}}$ with $\capconst > 20$.


\subsubsection{Proof of Lemma~\ref{LemPermVec}}
\label{SecProofLemPermVec}

Recall that the two vectors $\probshiftapprox_\ell$ and $\probshift$
are identical up to a permutation. Now suppose $\probshiftapprox_\ell
\neq \probshift$. Then there must exist some position $i \in
[\numwork-1]$ such that $[\probshift]_{i} < [\probshift]_{i+1}$
and $[\probshiftapprox_\ell]_{i} \geq
[\probshiftapprox_\ell]_{i+1}$.  Define the vector
$\probshiftapprox'$ obtained by interchanging the entries in
positions $i$ and $(i+1)$ in $\probshift$. The difference $\Delta
\defn \Lnorm{ \probshiftapprox' - \probshiftmag \eigvecobs_\ell
}{2}^2 - \Lnorm{ \probshift - \probshiftmag \eigvecobs_\ell
}{2}^2$ then can be bounded as
\begin{align*}
\Delta & = ([\probshiftapprox']_i \!-\! \probshiftmag
[\eigvecobs_\ell]_i)^2 + ([\probshiftapprox']_{i+1} \!-\!
\probshiftmag [\eigvecobs_\ell]_{i+1})^2 - ([\probshift]_i \! -\!
\probshiftmag [\eigvecobs_\ell]_i)^2 - ([\probshift]_{i+1} \! - \!
\probshiftmag [\eigvecobs_\ell]_{i+1})^2 \\
& = ([\probshift]_{i+1} \!-\! \probshiftmag [\eigvecobs_\ell]_i)^2 +
([\probshift]_{i} \!-\! \probshiftmag [\eigvecobs_\ell]_{i+1})^2 -
([\probshift]_i \!-\! \probshiftmag [\eigvecobs_\ell]_i)^2 -
([\probshift]_{i+1} \!-\! \probshiftmag [\eigvecobs_\ell]_{i+1})^2
\\
& = 2 \probshiftmag ([\probshift]_{i+1} -
[\probshift]_{i})([\eigvecobs_\ell]_{i+1} -
[\eigvecobs_\ell]_i)\\ & \leq 0,
\end{align*}
where the final inequality uses the fact that the ordering of the
entries in the two vectors $\probshiftapprox_\ell$ and
$\eigvecobs_\ell$ are identical, which in turn implies that
$[\eigvecobs_\ell]_{i} \geq [\eigvecobs_\ell]_{i+1}$. We have thus
shown an interchange of the entries $i$ and $(i+1)$ in $\probshift$,
which brings it closer to the permutation of $\probshiftapprox_\ell$,
cannot increase the distance to the vector $\probshiftmag
\eigvecobs_\ell$. A recursive application of this argument leads to
the inequality $\Lnorm{ \probshiftapprox_\ell - \probshiftmag
	\eigvecobs_\ell }{2} \leq \Lnorm{ \probshift - \probshiftmag
	\eigvecobs_\ell }{2}$.  Applying the triangle inequality then yields
\begin{align*}
\Lnorm{ \probshiftapprox_\ell - \probshift}{2} \leq \Lnorm{
	\probshiftapprox_\ell - \probshiftmag \eigvecobs_\ell }{2} +
\Lnorm{\probshiftmag \eigvecobs_\ell - \probshift }{2} \leq 2
\Lnorm{\probshiftmag \eigvecobs_\ell - \probshift }{2},
\end{align*}
as claimed.

\subsection{Proof of Corollary~\ref{CorCInt}}\label{SecProofCorCInt}

First suppose the matrix $\probmxstar$ satisfies the condition
\begin{align}
\label{EqnNiharCondition}
\Lnorm{\probInt - \half}{2} \Lnorm{1 - \hardstar}{2} \geq \sqrt{
	\frac{\capconst \numques \log^{2.5} (\numques \numwork)}{\pp}}
\end{align}
for a large enough constant $\capconst$ whose value is determined by the result of Theorem~\ref{ThmOBIWAN}.  
Applying the result of Theorem~\ref{ThmOBIWAN}, we
obtain that every question $j$ satisfying $(1 - \hardstar_j)^2 \Lnorm{
	\probInt - \half}{2}^2 \geq \frac{5 \log^{2.5} (\numques \numwork)}{\pp}$ is decoded correctly with a probability
at least $1 - e^{- \plaincon \log^{1.5} (\numques \numwork)}$. The total contribution from the remaining questions to the
$\probmxstar$-loss is at most $\frac{5  \log^{2.5} (\numques
	\numwork)}{\pp \numwork}$. A union bound over all questions and both
values of $\ell \in \{0,1\}$ then yields the claim that the aggregate
$\probmxstar$-loss is at most $\frac{5 \log^{2.5} (\numques
	\numwork)}{\pp \numwork}$ with probability at least $1 - e^{-
	\plaincon' \log^{1.5} (\numques \numwork)}$, for some positive constant
$\plaincon'$, as claimed in~\eqref{EqnOBIWANToProve}.

Otherwise, suppose that condition~\eqref{EqnNiharCondition} is
{violated}.  Then for any arbitrary $\anshat \in \{-1,1\}^{\numques}$, we 
have
\begin{align*}
\Qloss{\probmxstar}{\anshat}{\ansstar} &\leq \frac{1}{\numques \numwork}
\Lnorm{\probInt - \half}{2}^2 \Lnorm{1 - \hardstar}{2}^2 \leq \frac{6 \capconst
	\log^{2.5} \numques }{\numwork \pp},
\end{align*}
as claimed, where we have made use of the fact that $\numques \geq \numwork$.


\subsection{Proof of Theorem~\ref{ThmOBIHamDS}(a): OBI-WAN under the Dawid-Skene model}
\label{SecProofThmOBIHamDSa}
Throughout the proof, we make use the notation previously introduced
in the proof of Theorem~\ref{ThmOBIWAN}(a). As in this same proof, we
condition on some choice of $\setques_0$ and $\setques_1$ that
satisfies~\eqref{EqnHardnessSplit}. The proof of this theorem
follows the same structure as the proof of Theorem~\ref{ThmOBIWAN}(a) and
the lemmas within it. However, we must make additional arguments in
order to account for adversarial workers.  In the remainder of the
proof, we consider any $\ell \in \{0,1\}$, and then apply the union
bound across both values of $\ell$.

\noindent Our proof consists of the three steps:
\begin{enumerate}
	\item[(1)] We first show that the vector $\eigvecobs_\ell$ is a good
	approximation to $(\probDS - \half)$ up to a global sign.
	\item[(2)] Second, we show that the global sign of $\probshift$ is
	indeed recovered correctly.
	\item[(3)] Third, we establish guarantees on the performance of the
	WAN estimator for our setting.
\end{enumerate}
We work through each of these steps in turn.


\paragraph{Step 1}

We first show that the vector $\eigvecobs_\ell$ is a good
approximation to $\probDS - \half$ up to a global sign.  When
$\probmxstar = \probDS \ones^T$, we can set the vector $\hardstar = 0$
in the proof of Theorem~\ref{ThmOBIWAN}(a). We
also have $\probshift = \probDS - \half$. With these assignments, the
the arguments up to equation~\eqref{EqnApproxSValmostDone} in
Lemma~\ref{LemOBIPerturb} continue to apply even for the present
setting where $\probDS \in [0,1]^{\numwork}$. From these arguments, we
obtain the following approximation
guarantee~\eqref{EqnApproxSValmostDone} on recovering $\probshift$ up
to a global sign:
\begin{align}
\label{EqnCorOBIWAN0}
\min\{ \Lnorm{ \eigvecobs_\ell - \frac{1}{\probshiftmag} \probshift
}{2}^2, \Lnorm{\eigvecobs_\ell + \frac{1}{\probshiftmag} \probshift
}{2}^2 \} \leq \frac{1}{36} \frac{1}{\probshiftmag^2} \frac{\log^{1.5}
	\numques}{\pp},
\end{align}
with probability at least $1 - e^{- \plaincon \log^{1.5} \numques}$. 


\paragraph{Step 2}

The next step of the proof is to show that the global sign of
$\probshift$ is indeed recovered correctly.  Define two pairs of
vectors $\{\eigvecobsplus_\ell,\; \eigvecobsminus_\ell\}$ and
$\{\probshiftplus_\ell,\; \probshiftminus_\ell\}$, all lying in the
unit cube $[0,1]^\numwork$, with entries
\begin{align*}
[\eigvecobsplus_\ell]_i \defn \max \{ [\eigvecobsplus_\ell]_i, 0 \}
\quad & \mbox{and} \quad [\eigvecobsminus_\ell]_i \defn \min \{
[\eigvecobsminus_\ell]_i, 0 \} \quad \mbox{for every $i \in
	[\numwork]$; }\\
[\probshiftplus_\ell]_i \defn \max \{ [\probshiftplus_\ell]_i, 0 \},
\quad & \mbox{and} \quad [\probshiftminus_\ell]_i \defn \min \{
[\probshiftminus_\ell]_i, 0 \} \quad \mbox{for every $i \in
	[\numwork]$.}
\end{align*}
From the conditions assumed in the statement of the theorem, we have
$\Lnorm{\probshiftplus}{2} \geq \Lnorm{\probshiftminus}{2} +
\sqrt{\frac{4 \log^{2.5} (\numques \numwork)}{\pp}}$, whereas from the
choice of $\eigvecobs$ in the \obiwan estimator, we have
$\Lnorm{\eigvecobsplus}{2} \geq \Lnorm{\eigvecobsminus}{2}$. One can
also verify that
\begin{subequations}
	\begin{align}
	\label{EqnCorOBIWAN1}
	\Lnorm{\eigvecobs_\ell + \frac{1}{\probshiftmag} \probshift }{2}^2
	\geq \Lnorm{\eigvecobsplus_\ell + \frac{1}{\probshiftmag}
		\probshiftminus }{2}^2 + \Lnorm{\eigvecobsminus_\ell +
		\frac{1}{\probshiftmag} \probshiftplus }{2}^2.
	\end{align}
	Now suppose that $\Lnorm{\frac{1}{\probshiftmag} \probshiftplus}{2}
	\geq \Lnorm{\eigvecobsminus_\ell}{2} +  \sqrt{\frac{\log^{2.5}
			(\numques \numwork)}{\probshiftmag^2 \pp}}$. Then from the
	triangle inequality, we obtain the bound
	\begin{align}
	\label{EqnCorOBIWAN2}
	\Lnorm{\eigvecobsminus_\ell + \frac{1}{\probshiftmag} \probshiftplus
	}{2} \; \geq \; \Lnorm{\frac{1}{\probshiftmag} \probshiftplus }{2} -
	\Lnorm{\eigvecobsminus_\ell}{2} \; \geq \; \sqrt{\frac{\log^{2.5}
			(\numques \numwork)}{\probshiftmag^2 \pp}}.
	\end{align}
	Otherwise we have that $\Lnorm{\frac{1}{\probshiftmag}
		\probshiftplus}{2} < \Lnorm{\eigvecobsminus_\ell}{2} + 
	\sqrt{\frac{\log^{2.5} (\numques \numwork)}{\probshiftmag^2 \pp}}$. In
	this case, we have
	\begin{align}
	\label{EqnCorOBIWAN3}
	\Lnorm{\eigvecobsplus_\ell + \frac{1}{\probshiftmag} \probshiftminus
	}{2} 
	\geq  \Lnorm{\eigvecobsplus_\ell}{2} -
	\Lnorm{\frac{1}{\probshiftmag} \probshiftminus }{2} 
	& \geq 
	\Lnorm{\eigvecobsminus_\ell}{2} - \Lnorm{\frac{1}{\probshiftmag}
		\probshiftplus }{2} + 2\sqrt{\frac{\log^{2.5} (\numques
			\numwork)}{\probshiftmag^2 \pp}} \nonumber \\
	&    \geq  \sqrt{\frac{\log^{2.5} (\numques \numwork)}{\probshiftmag^2 \pp}}.
	\end{align}
\end{subequations}
Putting together the
conditions~\eqref{EqnCorOBIWAN1},~\eqref{EqnCorOBIWAN2}
and~\eqref{EqnCorOBIWAN3}, we obtain the bound $\Lnorm{ \eigvecobs_\ell + \frac{1}{\probshiftmag} \probshift }{2}^2
\geq  \frac{\log^{2.5} (\numques \numwork)}{\probshiftmag^2
	\pp}$. 
In conjunction with the result of equation~\eqref{EqnCorOBIWAN0}, this
bound guarantees the correct detection of the global sign, that is,
$\Lnorm{ \eigvecobs_\ell - \frac{1}{\probshiftmag} \probshift }{2}^2
\leq \frac{1}{36} \frac{1}{\probshiftmag^2} \frac{\log^{1.5}
	\numques}{\pp}$. 
The deterministic inequality afforded by Lemma~\ref{LemPermVec} then
guarantees that 
\begin{align}
\label{EqnCorOBIWAN4}
\Lnorm{ \probshiftapprox_\ell - \probshift }{2}^2
\leq \frac{1}{18} \frac{\log^{1.5} \numques}{\pp},
\end{align}
and this completes
the analysis of the OBI part of the estimator.


\paragraph{Step 3}

In the third step, we establish guarantees on the performance of the
WAN estimator for our setting.  Recall that since the WAN estimator
uses the permutation given by $\probshiftapprox_\ell$ and with this
permutation, acts on the observation $\obs_{1-\ell}$ of the other set
of questions, the noise $\noise_{1-\ell}$ is statistically independent
of the choice of $\probshiftapprox_\ell$, when conditioned on the
split $(\setques_0,\setques_1)$. Assume without loss of generality that
$\ansstar = 1$ and that the rows of $\probmxstar$ are arranged
according to the worker abilities, meaning that $\probDS_i \geq
\probDS_{i'}$ for every $i < i'$, or in other words, $\probshift_i \geq \probshift_{i'}$ for every $i < i'$. Recall our earlier notation of
$\onesvec{\winsize} \in \{0,1\}^{\numwork}$ denoting a vector with
ones in its first $\winsize$ positions and zeros elsewhere.

Now from the proof of Theorem~\ref{ThmWANnew} the following two properties
ensure that the WAN estimator decodes every question correctly with
probability at least $1 - e^{-\plaincon \log^{1.5} (\numques \numwork)}$:
(i) There exists some value $\winsize \geq \pp^{-1} \log^{1.5}(\numques \numwork)$ such that
$\inprod{\probshiftapprox_\ell}{\onesvec{\winsize}} \geq \frac{3}{4}
\sqrt{\frac{\winsize \log^{1.5} (\numques \numwork)}{\pp}}$, and (ii)
for every
$\winsize  \in [\numwork]$, it must be that $\inprod{\probshiftapprox_\ell}{\onesvec{\winsize}} > - \frac{1}{4}
\sqrt{\frac{\winsize \log^{1.5} (\numques \numwork)}{\pp}}$.  Let us first address property (i).
Lemma~\ref{LemVecBig} guarantees the existence of some value $\winsize \geq \lceil \half \Lnorm{\probshift}{2}^2 \rceil$ such that
\begin{align*}
\inprod{\probshiftplus}{\onesvec{\winsize}} \geq \frac{\sqrt{\winsize}
	\Lnorm{\probshiftplus}{2} }{\sqrt{\log(\numques \numwork)}}.
\end{align*}
If there exist multiple such values of $\winsize$, then choose the
smallest such value. Since the vector $\probshift$ has its entries
arranged in order, and since $\Lnorm{\probshiftplus}{2} \geq
\Lnorm{\probshiftminus}{2}$, we obtain the following relations for
this chosen value of $\winsize$:
\begin{align*}
\inprod{\probshift}{\onesvec{\winsize}} 
= (\probshiftplus)^T
\onesvec{\winsize} 
\geq \frac{\sqrt{\winsize}
	\Lnorm{\probshiftplus}{2} }{\sqrt{\log(\numques \numwork)}} 
\geq
\frac{\Lnorm{ \probshift }{2}}{2} \sqrt{\frac{\winsize}{\log (\numques
		\numwork)}} 
\geq \sqrt{\frac{\log^{2.5} (\numques
		\numwork)}{\pp} \frac{\winsize}{\log (\numques
		\numwork)}}.
\end{align*}
The Cauchy-Schwarz inequality then implies
\begin{align*}
\inprod{\probshiftapprox_\ell}{\onesvec{\winsize}} \geq
\inprod{\probshift}{\onesvec{\winsize}} - \sqrt{\winsize}
\Lnorm{\probshiftapprox_\ell - \probshift}{2} \stackrel{(i)}{\geq}
\frac{3}{4} \sqrt{\frac{\winsize \log^{1.5} (\numques
		\numwork)}{\pp}},
\end{align*}
where the inequality (i) also uses our earlier
bound~\eqref{EqnCorOBIWAN4}, thereby proving the first property. Now
towards the second property, we use the condition
$\inprod{\probshift}{\ones} \geq 0$. Since the entries of $\probshift$
are arranged in order, we have $\inprod{\probshift}{\onesvec{\winsize}}
\geq 0$ for every $\winsize \in [\numwork]$.  Applying the Cauchy-Schwarz
inequality yields
\begin{align*}
\inprod{\probshiftapprox_\ell}{\onesvec{\winsize}} \geq
\inprod{\probshift}{\onesvec{\winsize}} - \sqrt{\winsize}
\Lnorm{\probshiftapprox_\ell - \probshift}{2} \stackrel{(ii)}{>}
-\frac{1}{4} \sqrt{\frac{\winsize \log^{1.5} (\numques \numwork)}{\pp}},
\end{align*}
where the inequality (ii) also uses our earlier
bound~\eqref{EqnCorOBIWAN4}, thereby proving the second property. This
argument completes the proof of part (a).


\subsection{Proof of Theorem~\ref{ThmOBIHamDS}(b): Converse result under the Dawid-Skene model}
\label{SecProofThmOBIHamDSb}
The Gilbert-Varshamov
bound~\cite{gilbert1952comparison,varshamov1957estimate} guarantees
existence of a set of $\packnum$ vectors, $\ans^1,\ldots,
\ans^\packnum \in \{-1,1\}^{\numques}$ such that the normalized Hamming distance~\eqref{EqnDefnHamming} between any pair of vectors in this set is lower bounded as
$\hamming ( \ans^\ell, \ans^{\ell'}) \geq 0.25 , \qquad
\mbox{for every $\ell, \ell' \in [\packnum]$}$, 
where $\packnum = \exp( \plaincon_1 \numques)$ for some
constant $\plaincon_1 > 0$. For each $\ell \in [\packnum]$, let
$\mprob^\ell$ denote the probability distribution of $\obs$ induced by
setting $\ansstar = \ans^\ell$. When $\probmxstar = \probDS \ones^T$
for some $\probDS \in [\frac{1}{10}, \frac{9}{10}]^\numwork$, we have an upper bound on the Kullback-Leibler divergence between any
pair of distributions $\ell \neq \ell' \in [\packnum]$ as $
\kl{ \mprob^\ell }{ \mprob^{\ell'} } \; \leq 25 \;\pp \numques \;
\Lnorm{\probDS - \half}{2}^2 \; \leq \; 25 \plaincon \numques$,
where we have used the assumption $\Lnorm{\probDS - \half}{2}^2 \leq \frac{\plaincon}{\pp}$. 
Putting the above observations together into Fano's
inequality~\cite{cover2012elements} yields a lower bound on the
expected value of the normalized Hamming error~\eqref{EqnDefnHamming} for any estimator $\anshat$ as:
\begin{align*}
\Exs[ \hamming(\anshat, \ansstar ) ] \geq \frac{1}{8} \Big( 1
- \frac{25 \plaincon \numques + \log 2}{ \plaincon_1 \numques } \Big)
\stackrel{(i)}{\geq} \frac{1}{10},
\end{align*}
as claimed, where inequality (i) results from setting the value of
$\plaincon$ as a small enough positive constant.


\subsection{Proof of Proposition~\ref{PropOBIWANCperm}: OBI-WAN under the permutation-based model}
\label{SecProofPropOBIWANPerm}
First, suppose that $\pp < \frac{\log^{1.5} (\numques
	\numwork)}{\numwork}$. Then the condition~\eqref{EqnPropOBIWANcondition} is \emph{not} satisfied for any question, and hence the first part of the claim is trivially (vacuously) true. In this case, we also have
\begin{align*}
\Qloss{\probmxstar}{\ansOBIWAN}{\ansstar} \leq 1 \leq
\frac{1}{\sqrt{\numwork \pp}} \log (\numques \numwork),
\end{align*}
due to which the second claim also follows immediately. \\

Otherwise, we may assume that $\pp \geq \frac{\log^{1.5} (\numques
	\numwork)}{\numwork}$.  For any index $\ell \in \{0,1\}$, consider
an arbitrary permutation $\permwork_\ell$. Observe that conditioned on
the split $(\setques_0,\setques_1)$, the data $\obs_{1 - \ell}$ is
independent of the choice of the permutation $\permwork_\ell$. Now
consider any question $j \in \setques_{1 - \ell}$ that satisfies~\eqref{EqnPropOBIWANcondition}. 	We then apply Theorem~\ref{ThmWANnew} with the parameter  $\winsize_j = \numwork$ in~\eqref{EqnDefnSetquesWAN}, and note that the permutation $\permwork$ specified in the statement of Theorem~\ref{ThmWANnew} does not matter when $\winsize_j = \numwork$ . This result guarantees that our estimator satisfies $
\mprob( [\ansOBIWAN]_{j} = \ansstar_{j} ) \geq 1- e^{-\plaincon
	\log^{1.5}(\numques \numwork)}$. A union bound over all questions $j \in [\numques]$ satisfying condition~\eqref{EqnPropOBIWANcondition} implies that all of these questions are decoded correctly with probability at least $1- e^{-\plaincon'
	\log^{1.5}(\numques \numwork)}$.  
Furthermore, all remaining questions can contribute a total of at most $\frac{3}{2}
\frac{1}{\sqrt{\numwork \pp}} \log (\numques \numwork)$ to the
$\probmxstar$-loss. This yields the second part of the claim.


\section{Discussion}
\label{SecConclusions}

We propose a new \emph{permutation-based model} for crowdsourced labeling which is considerably more general than the popular Dawid-Skene model, provide a computationally-efficient algorithm ``OBI-WAN'', and associated statistical guarantees and empirical evaluations. We hope that the desirable
features of the permutation-based model will encourage researchers and
practitioners to further build on the permutation-based core of this
model.

This work gives rise to several open problems that are
theoretically challenging and of interest to practitioners.
\begin{itemize}
	\item The problem of establishing optimal minimax risk under the
	permutation-based model for computationally-efficient estimators
	remains open, and is related to several
	problems~\cite{shah2015stochastically, flammarion2016optimal, shah2018low}
	involving permutations that have an unresolved difference in the
	computationally efficient and inefficient rates.\footnote{That said, there are related problems~\cite{shah15simple,heckel2019active} involving permutation-based models where statistically optimal techniques are computationally efficient, and also adapt optimally to much more restrictive parameter-based models.} It is of interest to reduce this gap in the future, possibly building on recent work~\cite{mao2018breaking,liu2020better} on rates of computationally-efficient algorithms for permutation-based models. 
	
	\item \rev{In addition to the global  minimax error, and it is of interest to obtain sharp bounds on adaptivity to the underlying noise levels under various models. Such adaptive bounds  are obtained for the Dawid-Skene and intermediate models in  the papers~\cite{karger2011iterative, zhang2014spectral,    gao2016exact, khetan2016reliable}.}
	
	\item It will be useful to extend the proposed
	permutation-based model and associated algorithms to more general
	settings in crowdsourcing such as a fixed design setup (i.e., where each worker answers a fixed, given subset of questions), questions with more than two
	choices, and with asymmetric error probabilities of workers (two-coin Dawid-Skene model). 
	
	\item \rev{Our results are loose by logarithmic factors. In the future, it will be of interest to tighten this gap, possibly via new results on the law of iterative logarithm in  non-asymptotic regimes such as~\cite{brunel2019nonasymptotic}, and alongside understand its relation with such gaps in other problems involving shape-constrained estimation~\cite{han2019global}.}
\end{itemize}  
Finally, there are many other problem settings involving estimation from noisy (as well as biased and subjective) labelers, such as in peer review~\cite{stelmakh2019testing,wang2018your,noothigattu2018choosing,stelmakh2018forall,wang2020debiasing}, and it is of interest to see whether permutation-based models and associated techniques can play a useful role in these applications. 


\subsection*{Acknowledgements}
This work was partially supported by Office of Naval Research MURI
grant DOD-002888, Air Force Office of Scientific Research Grant
AFOSR-FA9550-14-1-001, Office of Naval Research grant ONR-N00014, National Science Foundation Grants CIF: 31712-23800, CRII: CIF: 1755656, and  CIF: 1763734.   The work of NBS was also supported in part by a
Microsoft Research PhD fellowship.

We thank the authors of the
paper~\cite{zhang2014spectral} for sharing their implementation of
their Spectral-EM algorithm. 


 \bibliographystyle{abbrv} 

\bibliography{bibtex}

\appendix

~\\~\\\noindent {\bf \Large Appendix: Analysis of the majority voting estimator}\\


In this section, we analyze the majority voting estimator, given by
\begin{align*}
[\ansmv]_{j} \in \argmax \limits_{b
  \in \{-1,1\}} \sum_{i=1}^{\numwork} \indicator{ \obs_{ij} = b}
  \mbox{\qquad for
every $j \in [\numques]$}.
\end{align*}  
Here we use $\indicator{\cdot}$ to denote
the indicator function.  

The following theorem provides bounds on the
risk of majority voting under the $\probmxstar$-semimetric loss in the regime of interest~\eqref{EqnRegime}.
\begin{proposition}
\label{PropMajorityLower}
For the majority vote estimator, the risk over the Dawid-Skene
class is lower bounded as
\begin{align}
\label{EqnLowerMajority}
\sup_{\ansstar \in \{-1,1\}^\numques} \sup_{\probmxstar \in \classDS}
\Exs [ \Qloss{\probmxstar}{\ansmv}{\ansstar} ] \geq \ULOW
\frac{1 }{ \sqrt{\numwork \pp}},
\end{align}
for some positive constant $\ULOW$.
\end{proposition}
A comparison of the bound~\eqref{EqnLowerMajority} with the results of
Theorem~\ref{ThmMinimax}, Theorem~\ref{ThmOBIWAN}(a) and Theorem~\ref{ThmOBIHamDS} shows that the majority voting estimator is
suboptimal in terms of the sample complexity.  Since this
suboptimality holds for the (smaller) Dawid-Skene model class, it also
holds for the (larger) intermediate model class, as well as the
permutation-based model class.

\noindent The remainder of this section is devoted to the proof of
this claim.

\subsubsection*{Proof of Proposition~\ref{PropMajorityLower}}

We begin with a lower bound due to
Feller~\cite{feller1943generalization} (see also~\cite[Theorem
  7.3.1]{matouvsek2001probabilistic}) on the tail probability of a sum
of independent random variables.
\begin{lemma}[Feller]
\label{LemLowerTail}
There exist positive universal constants $\plaincon_1$ and
$\plaincon_2$ such that for any set of independent random variables
$X_1,\ldots,X_\numwork$ satisfying $\Exs[X_i]=0$ and $|X_i| \leq M$
for every $i \in [\numwork]$, if $\sum_{i=1}^{\numwork} \Exs[(X_i)^2]
\geq \plaincon_1$ then \begin{align*} \mprob \big(
  \sum_{i=1}^{\numwork} X_i > t \big) \geq \plaincon_2 \exp
  \Big(\frac{-t^2}{12 \sum_{i=1}^{\numwork} \Exs[(X_i)^2]} \Big),
\end{align*}
for every $t \in [0, \frac{ \sum_{i=1}^{\numwork} \Exs[(X_i)^2]}{M^2
    \sqrt{\plaincon_1}}]$.
\end{lemma}

In what follows, we use Lemma~\ref{LemLowerTail} to derive the claimed
lower bound on the error incurred by the majority voting algorithm. To
this end, let $\setexpertworkers \subset [\numwork]$ denote the set of
some $\cardinality{\setexpertworkers} = \sqrt{\frac{\numwork}{2\pp}}$
workers. Consider the following value of matrix $\probmxstar$:
\begin{align*}
\probmxstar_{ij} = 
\begin{cases}
1 & \qquad \mbox{if $i \in \setexpertworkers$}\\ \half & \qquad
\mbox{otherwise}.
\end{cases}
\end{align*}
Then for any question $j \in [\numques]$, we have $\sum_{i =
  1}^{\numwork} ( 2 \probmxstar_{ij} - 1)^2 =
\sqrt{\frac{\numwork}{2\pp}}$.

Now suppose that $\ansstar_{j} = -1$ for every question $j \in
[\numques]$. Then for every $i \in \setexpertworkers$, the
observations are distributed as
\begin{align*}
\obs_{ij} =
\begin{cases}
0 & \qquad \mbox{with probability $1-\pp$}\\ -1 & \qquad \mbox{with
  probability $\pp$,}
\end{cases}
\end{align*}
and for every $i \notin \setexpertworkers$, as
\begin{align*}
\obs_{ij} =
\begin{cases}
0 & \qquad \mbox{with probability $1-\pp$}\\ -1 & \qquad \mbox{with
  probability $0.5 \pp$}\\ 1 & \qquad \mbox{with probability $0.5
  \pp$.}
\end{cases}
\end{align*}

Consider any question $j \in [\numques]$. Then in this setting, the
majority voting estimator incorrectly estimates the value of
$\ansstar_j$ when $\sum_{i =1}^{\numwork} \obs_{ij} > 0$. We now use
Lemma~\ref{LemLowerTail} to obtain a lower bound on the probability of
the occurrence of this event.  Some simple algebra yields
\begin{align*}
\sum_{i=1}^{\numwork} \Exs[ \obs_{ij} ] = -
\cardinality{\setexpertworkers} \pp \qquad \mbox{and} \qquad
\sum_{i=1}^{\numwork} \Exs[ (\obs_{ij})^2 ] = \numwork \pp.
\end{align*}
In order to satisfy the conditions required by the lemma, we assume
that $\numwork \pp > \plaincon_1$. Note that this condition makes the
problem strictly easier than the condition $\numwork \pp \geq 1$
assumed otherwise, and affects the lower bounds by at most a constant
factor $\plaincon_1$. An application of Lemma~\ref{LemLowerTail} with
$t = - \sum_{i=1}^{\numwork} \Exs[ \obs_{ij} ] =
\cardinality{\setexpertworkers} \pp$ now yields
\begin{align*}
\mprob( \sum_{i=1}^{\numwork} \obs_{ij} > 0) \geq \plaincon_2 \exp
\Big( \frac{- \cardinality{\setexpertworkers}^2 \pp^2}{12 \numwork
  \pp} \Big) \stackrel{(i)}{\geq} \plaincon',
\end{align*}
for some constant $\plaincon'>0$ that may depend only on $\plaincon_1$
and $\plaincon_2$, where inequality $(i)$ is a consequence of the
choice $\cardinality{\setexpertworkers} =
\sqrt{\frac{\numwork}{2\pp}}$.

Now that we have established a constant-valued lower bound on the
probability of error in the estimation of $\ansstar_j$ for every $j
\in [\numques]$, for the value of $\probmxstar$ under consideration,
we have
\begin{align*}
\mprob( [\ansmv]_j \neq \ansstar_j) \sum_{i=1}^{\numwork} (\probmxstar_{ij} -
\half)^2 \geq \sqrt{\frac{\numwork}{2\pp}} \plaincon',
\end{align*}
and consequently $\Exs[\Qloss{\probmxstar}{\ansmv}{\ansstar}] \geq
\frac{\plaincon'}{\sqrt{2 \numwork \pp}}$, as claimed.


\end{document}